\documentclass[default,iicol]{sn-jnl}
\usepackage{lmodern}

\usepackage{lmodern}

\usepackage{graphicx}%
\usepackage{multirow}%
\usepackage{amsmath,amssymb,amsfonts}%
\usepackage{amsthm}%
\usepackage{mathrsfs}%
\usepackage[title]{appendix}%
\usepackage{xcolor}%
\usepackage{textcomp}%
\usepackage{manyfoot}%
\usepackage{booktabs}%
\usepackage{algorithm}%
\usepackage{algorithmicx}%
\usepackage{algpseudocode}%
\usepackage{listings}%

\usepackage{makecell}

\usepackage[switch]{lineno}

\usepackage{url}            
\usepackage{nicefrac}       
\usepackage{microtype}      
\usepackage{siunitx}
\usepackage{colortbl}
\usepackage{array}
\usepackage{siunitx} 
\usepackage{tabularx}
\usepackage{colortbl}
\definecolor{BrickRed}{RGB}{156, 31, 46}     
\definecolor{OliveGreen}{RGB}{106, 133, 82}  
\usepackage{multirow} 
\definecolor{backcolor}{RGB}{232, 242, 255}
\usepackage{pifont} 
\newcommand{\gou}{{\color{OliveGreen}\ding{51}}}
\newcommand{\cha}{{\color{BrickRed}\ding{55}}}
\newcommand{\grayrow}{\rowcolor{gray!10}}

\theoremstyle{thmstyleone}%
\theoremstyle{thmstyletwo}%
\definecolor{upcolor}{RGB}{57,182,74}
\newcommand{\up}[1]{\textcolor{upcolor}{$\uparrow$ #1}}
\newcommand{\down}[1]{\textcolor{black}{$\downarrow$ #1}}

\theoremstyle{thmstylethree}%

\definecolor{color_obs}{RGB}{255, 253, 230} 
\definecolor{color_rel}{RGB}{255, 240, 230} 
\definecolor{color_sym}{RGB}{235, 245, 255} 
\definecolor{color_plan}{RGB}{235, 255, 235} 

\definecolor{lightpurple}{RGB}{240, 230, 255} 
\definecolor{lightpink}{RGB}{255, 235, 240}   

\begin{document}

\title[Article Title]{Evidence-Grounded Trustworthy Multimodal Reasoning and Evaluation Benchmark in Complex Urban Scenes}






\author[1]{\fnm{Zhaoyang} \sur{Wei}}\email{\{\small{weizhaoyang23, jiangbowen24, hanxumeng19, lijiashu24, yuxuehui17\}@mails.ucas.ac.cn}}
\author[1]{\fnm{Bowen} \sur{Jiang}}
\author[1]{\fnm{Xumeng} \sur{Han}}
\author[1]{\fnm{Jiashu} \sur{Li}}

\author[2]{\fnm{Xuehui} \sur{Yu}}
\author[1,3]{\fnm{Yuling} \sur{Liu}}\email{\small{liuyuling@iie.ac.cn}}
\author[1]{\fnm{Guorong} \sur{Li}}\email{\small{\{liguorong, jiaojb\}@ucas.ac.cn}}

\author*[1]{\fnm{Zhenjun} \sur{Han}}\email{\small{hanzhj@ucas.ac.cn}}

\author[1]{\fnm{Jianbin} \sur{Jiao}}

\affil[1]{\small{\orgname{University
of Chinese Academy of Sciences (UCAS)}, \orgaddress{\city{Beijing}, \postcode{101480}, \country{China}}}}

\affil[2]{\small{\orgname{Tencent CDG}, \orgaddress{\city{Shenzhen}, \postcode{518054}, \country{China}}}}

\affil[3]{\small{\orgname{Institute of Information Engineering, CAS}, \orgaddress{\city{Beijing}, \postcode{100085}, \country{China}}}}

\newcommand{\etal}{\textit{et al}.}
\newcommand{\ie}{\textit{i}.\textit{e}.}
\newcommand{\eg}{\textit{e}.\textit{g}.}

\abstract{
While Multimodal Large Language Models (MLLMs) demonstrate impressive proficiency in benign scenarios, their cognitive reliability significantly deteriorates in complex scenes characterized by adverse conditions. 
In such scenarios, models relying on implicit inference devoid of visual evidence succumb to disconnects between perception and logic; simultaneously, existing outcome-oriented benchmarks fail to diagnose the underlying reasoning process.
To bridge this gap, we present AD$^2$-Bench, which establishes a rigorous \emph{Hierarchical Visual Diagnosis} to decompose reasoning into a structured Chain of Evidence (CoE). 
This granular perspective reveals that robust cognition is predicated on precise evidence acquisition.
From this perspective, we formulate reasoning probabilistically, attributing the collapse to two bottlenecks: \textbf{(1) Spatial Ambiguity}, failing to distinguish targets from background clutter (i.e., localization error); and \textbf{(2) Semantic Uncertainty}, misinterpreting semantics due to feature degradation (i.e., understanding error).
To resolve these evidence gaps, we propose Evidence-Grounded Visual Reasoning (EGVOR). 
Departing from implicit inference, EGVOR reformulates reasoning as the explicit generation of Evidence Atoms—structured triplets that enforce strict spatial-semantic alignment. 
We orchestrate training via a hierarchical curriculum, progressing from reflective supervision establishment to reinforcement learning that explicitly rewards the reduction of reasoning variance. 
Extensive experiments demonstrate that EGVOR substantially mitigates instability, establishing a robust framework for trustworthy multimodal cognition.
}





\maketitle

\section{Introduction}
\label{sec:intro}
\begin{table*}[t]
\centering
\resizebox{\linewidth}{!}{%
  \begin{tabular}{@{}l|ccc|cccc@{}} 
    \toprule
    Dataset & Multimodal  & Adverse Condition & Real World & QA Pairs & Multi-prompt & Hierarchical assessment & Decision  \\
    \midrule
    \rowcolor{lightpurple}
    \multicolumn{8}{c}{\textit{Standard Vision Datasets}} \\
    \midrule
    COCO 2017 & \cha & \cha & \cha & \cha & \cha & \cha & \cha\\
    CODA & \cha & \cha & \gou & \cha & \cha & \cha & \cha\\
    StreetHazards & \cha & \cha & \gou & \cha & \cha & \cha & \cha\\
    ACDC & \cha & \gou & \gou & \cha & \cha & \cha & \cha\\
    \midrule
    \rowcolor{lightpink}
    \multicolumn{8}{c}{\textit{General MLLM Benchmarks}} \\
    \midrule
    RealWorldQA & \gou & \cha & \gou & 0.8k & \cha & \cha & \cha \\
    CV-Bench & \gou & \cha & \cha & 2.6k & \cha & \cha & \cha \\
    MME & \gou & \cha & \cha & 2.4k & \cha & \cha & \cha \\
    MMBench & \gou & \cha & \cha & 3.2k & \cha & \cha & \cha \\
    MME-RealWorld & \gou & \cha & \gou & 29k & \cha & \cha & \cha \\
    \midrule
    \midrule
    \rowcolor{backcolor}
    \textbf{AD$^2$-Bench (ours)}  & \gou & \gou & \gou & 70k & \gou & \gou & \gou \\
    \bottomrule
  \end{tabular}%
} 
\caption{Comparison with existing datasets. Adverse Condition and Real World indicate whether the dataset specifically targets in real complex scenes.}
\label{tab:dataset_comparison}
\vspace{-10pt}
\end{table*}
\begin{figure}[t]
    \centering
    \includegraphics[width=1\linewidth]{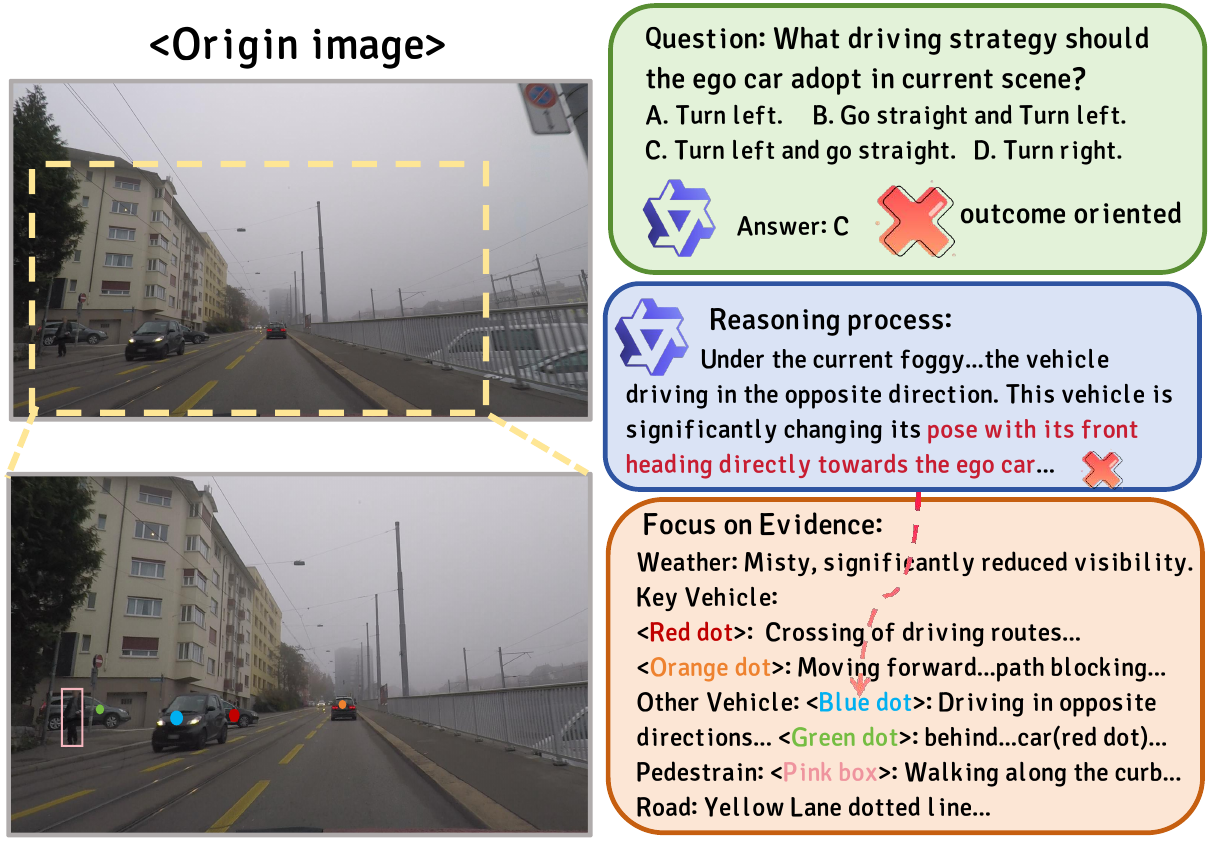}
    \vspace{-15pt}
    \caption{\textcolor{black}{Standard outcome-oriented metric (top) masks hallucinations in complex scenes. In contrast, our Evidence-Centric diagnosis (bottom) anchors reasoning to verifiable visual cues, exposing the perception-logic disconnect.}}
    \label{fig:challenge}
    \vspace{-10pt}
\end{figure}
Multimodal large language models (MLLMs) have achieved strong performance on a wide range of vision–language tasks, including captioning, visual question answering, and general cross-modal reasoning~\cite{bai2023qwenvl, chen2024internvl15, wang2024qwen2vl}. 
By coupling vision encoders with large language backbones, these models process images and text in a unified semantic space and exhibit robust zero-shot generalization in general scenarios. 
However, moving from these \emph{benign environments} to \emph{visually adverse and structurally complex real-world scenes}—such as cluttered streets under rain, fog, night illumination, or heavy occlusion—remains challenging. 
In these settings, MLLMs frequently exhibit \emph{visual hallucination} (describing non-existent objects or attributes), unstable reasoning chains, and decisions that are difficult to interpret or trust. 
Evaluating such systems solely by final-answer accuracy is therefore inadequate: a model may produce a correct answer for spurious reasons, or fail in ways that conventional metrics cannot reveal.

Diagnosing and resolving these failures is impeded by a systemic \emph{dual opacity}, as illustrated in Fig.~\ref{fig:challenge}. 
On the modeling side, standard MLLMs rely on implicit, end-to-end reasoning that lacks explicit visual evidence, leaving the reasoning process vulnerable to environmental noise. 
Simultaneously, prevailing benchmarks—ranging from general-purpose suites like MMBench~\cite{liu2024mmbench} and MME~\cite{fu2025mme} to robustness-focused sets like MME-RealWorld~\cite{zhang2024mmerw}—reinforce this challenge through \textbf{outcome-oriented evaluation} (Fig.~\ref{fig:challenge}, top right). 
This paradigm assesses models solely on final-answer correctness, effectively treating the intermediate cognition as a black box. 
Crucially, Visual Evidence serves as the missing link for both robust reasoning and transparent diagnosis. 
Without explicit evidence reasoning chain, it is intractable to disentangle whether an error stems from a failure in \emph{Perceptual Grounding} (e.g., missing key targets due to fog) or a failure in \emph{Cognitive Reasoning} (e.g., misinterpreting the vehicle's intent). 
Bridging this gap thus demands a paradigm shift: moving from implicit to chain of Evidence-Centric reasoning construction, and from outcome-based to process-oriented diagnosis.

\begin{figure}[t]
    \centering
    \includegraphics[width=1\linewidth]{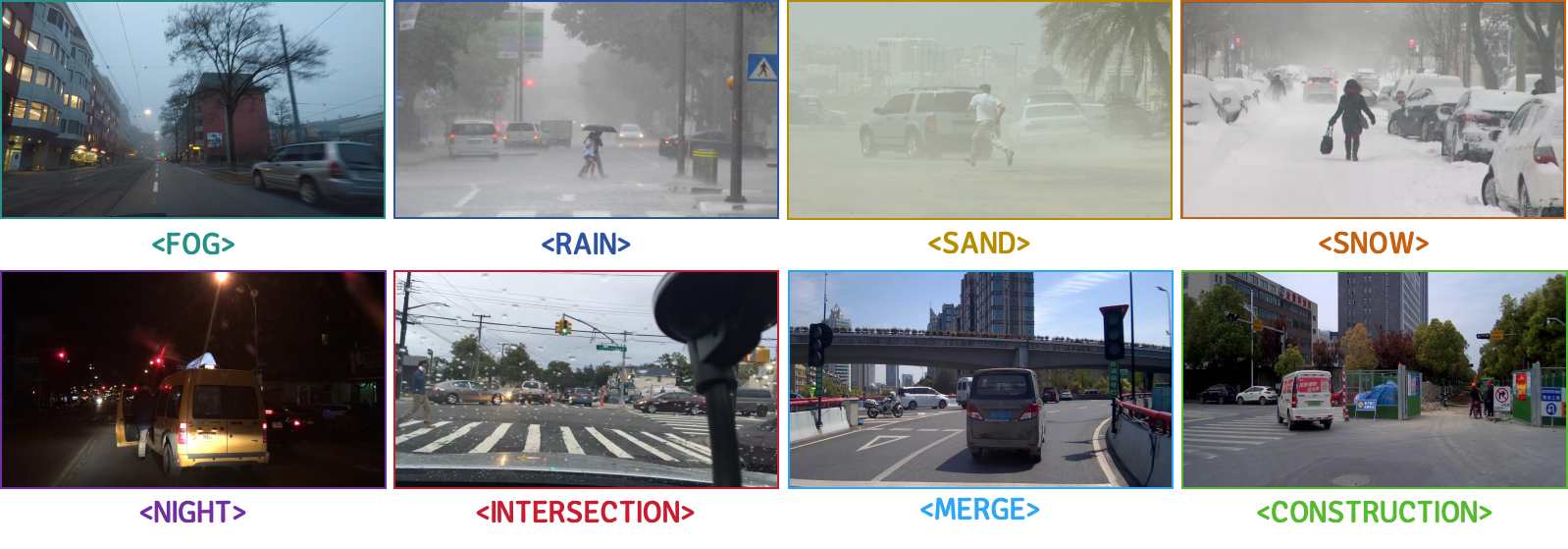}
    \vspace{-15pt}
    \caption{Adverse conditions include adverse weather (top) and complex scenarios (down).}
    \label{fig:1}
    \vspace{-5pt}
\end{figure}

To dismantle the black-box nature of current evaluations, we introduce AD$^2$-Bench, a comprehensive benchmark designed to diagnose the \emph{hierarchical cognitive capabilities} of MLLMs in visually adverse and complex scenes. 
Comprising approximately 10K images and 70K QA pairs, the benchmark captures diverse environmental degradations and intricate layouts (see Fig.~\ref{fig:1}). 
Crucially, distinct from traditional outcome-oriented formats, we establish an Atomic Annotation Protocol with multi-granularity vision prompts grounded in human expert cognition. 
We decompose complex queries into a rigorous \textbf{Chain of Evidence (CoE)}, which explicitly maps the cognitive flow from \emph{Environmental and Object Perception} to fine-grained \emph{Interaction Analysis}—resolving spatial relations and vulnerable road users—before synthesizing \emph{Holistic Decisions}. 
This structured decomposition enables a \textbf{Trustworthiness-oriented Diagnosis} beyond simple accuracy. 
By aligning the model's intermediate reasoning with human-annotated evidence chains, we propose a suite of orthogonal metrics, assessing Factual Fidelity, Logical Reliability, Self-Consistency, and Explainability to quantitatively measure the reliability of the model cognitive process.

Serving as a diagnostic prism, AD$^2$-Bench reveals a critical dependency: robust reasoning in complex scenes is predicated on precise visual grounding.
Formalizing this probabilistically, we trace the high variance in latent reasoning context to two fundamental bottlenecks:
\textbf{(1) Spatial Ambiguity:} In complex scenes with dense occlusion and clutter, the model's attention mechanism suffers from dispersion, drifting towards high-contrast distractors rather than the relevant targets.
\textbf{(2) Semantic Uncertainty:} Even when the target is localized, environmental degradation (\eg, rain streaks, low light) corrupts the feature representation. This challenge forces the model to abandon visual evidence and over-rely on language priors, leading to hallucinations.
These observations motivate an explicit evidence construction mechanism that reduces both spatial ambiguity and semantic uncertainty.

To address these challenges, we propose \textbf{Trustworthy Evidence-Grounded Visual Reasoning (EGVOR)}, which turns multimodal reasoning from implicit text generation into explicit evidence construction within an end-to-end model. We posit that robust reasoning requires Component Filtration: actively filtering out environmental noise by locking onto critical evidence. EGVOR operationalizes this idea by generating a sequence of Evidence Atoms. Each atom follows a \emph{Locate--Attend--Describe} cycle: a \emph{Spatial Anchor} restricts the visual support to reduce spatial ambiguity, while a \emph{Grounded Description} verifies semantic content to mitigate semantic uncertainty.

\begin{figure*}[t]
    \centering
    \includegraphics[width=1\linewidth]{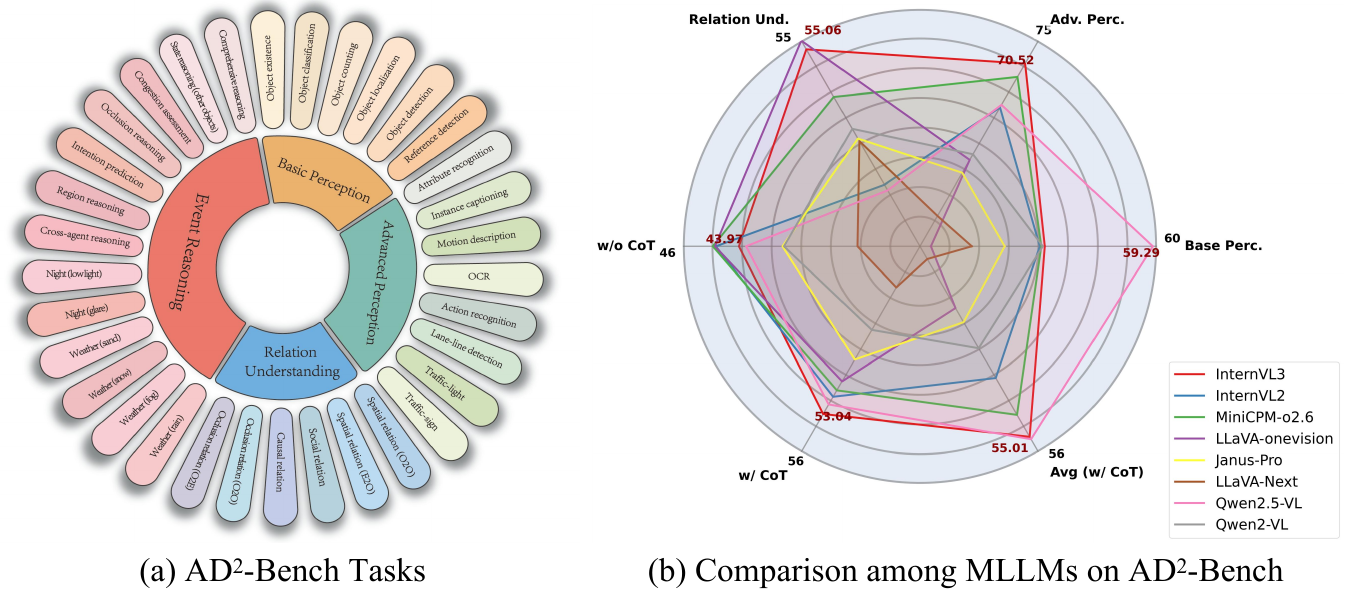}
    \vspace{-15pt}
    \caption{ (a) Task Categories. Our benchmark spans 4 key demensions and 33 subtasks highly related to real-world scenarios, including 10k high-resolution images and 70K annotations.
(b) Model Performance. Average accuracies of advanced MLLMs are shown on the dataset.}
    \label{fig:3}
    \vspace{-5pt}
\end{figure*}

We train EGVOR via a hierarchical curriculum designed to align the model with this cognitive paradigm. A \emph{Reflective Supervised Fine-Tuning} stage first establishes the syntactic structure of evidence chains and endows the model with self-correction capabilities. Subsequently, a \emph{Cognitive Alignment} stage utilizing Group Relative Policy Optimization optimizes the evidence-chain trajectory. By employing dense spatial rewards and semantic alignment rewards, we incentivize the policy to construct valid, grounded, and logically sound evidence chains, effectively closing the loop between perception and reasoning.
In summary, this paper makes the following contributions:

\begin{itemize}
    \item We present \textbf{AD$^2$-Bench}, a large-scale benchmark for evaluating MLLMs in visually complex scenes. It features a unique Hierarchical Visual Diagnosis that decomposes reasoning into Chains of Evidence, enabling a fine-grained, trustworthiness-oriented evaluation framework.

    \item We theoretically diagnose the reasoning failures in adverse scenarios as Spatial Ambiguity and Semantic Uncertainty, and propose EGVOR to address them. EGVOR reformulates multimodal reasoning as the sequential construction of Evidence Atoms, utilizing Component Filtration to reduce latent variance.

    \item Through extensive experiments on AD$^2$-Bench and external benchmarks, we demonstrate that EGVOR significantly outperforms state-of-the-art MLLMs in both grounded reasoning accuracy and trustworthiness metrics, offering a robust solution for interpretable decision-making in complex real-world environments.
\end{itemize}



\section{Related Work}
\label{sec:related}
\textbf{Multimodal Large Language Models.}
Multimodal large language models (MLLMs) couple vision encoders with large language backbones to enable unified visual–textual reasoning~\cite{ hong2024cogagent, stone2023open, yang2025octopus, dong2024insight, xu2024vlm, dong2025fusionmamba, zhang2019holistic}.  Early systems such as Flamingo~\cite{alayrac2022flamingo} and BLIP-2~\cite{li2023blip2} inject visual features into frozen LLMs via cross-attention or lightweight querying transformers, allowing image-grounded QA and instruction following.  Later models show that simply projecting CLIP-style image embeddings~\cite{radford2021learning} into the LLM token space, followed by instruction tuning, yields strong open-ended multimodal capabilities.  This design underlies the LLaVA family~\cite{liu2023llava, liu2024llava1.5, liu2024llavanext}, the Qwen2-VL and Qwen2.5-VL series~\cite{wang2024qwen2, bai2025qwen2, wang2024qwen2vl}, InternVL models~\cite{chen2024internvl15, zhu2025internvl3}, and related architectures~\cite{ye2024mplug3, qiao2025lipt}.

A recent focus is scaling MLLMs to high-resolution and large-field-of-view images, which is crucial for cluttered, detail-rich scenes.  LLaVA-NeXT~\cite{liu2024llavanext} and AnyRes-style approaches~\cite{chen2024dragonfly} preserve fine structures via tiling or adaptive rescaling, while Qwen2-VL and Qwen2.5-VL introduce multimodal rotary position encodings to support arbitrary resolutions and aspect ratios~\cite{wang2024qwen2vl, bai2025qwen25vl}.  Together with improvements in data quality and scale~\cite{zhu2025internvl3, wang2024world, wu2024deepseekvl2, qiao2025himamba}, these models provide strong generic visual grounding and form robust backbones for downstream tasks.  Beyond generic Web imagery, MLLMs have been connected to robotics and driving stacks~\cite{chen2024spatialvlm, tian2024drivevlm, chen2024drivinggpt, fu2024drive, cui2024drivellm}, where they describe scenes, query hazards, or explain actions in natural language.  However, visual grounding in these systems is largely implicit in attention patterns, and answers are typically produced in a single step, which can be brittle in complex, cluttered scenes.  Our work builds on these foundations while explicitly targeting fine-grained, stepwise reasoning in such environments.

\textbf{Multimodal Reasoning}
The success of chain-of-thought prompting in text-only LLMs~\cite{o1, guo2025deepseekr1} has motivated efforts to bring explicit reasoning into multimodal settings.  Early multimodal chain-of-thought (MCoT) approaches decompose tasks into intermediate steps via predefined workflows, auxiliary tools, or staged pipelines~\cite{liu2024chain, mondal2024kam}.  Many encourage the model to first localize regions of interest or refine latent visual features~\cite{wu2024v, fu2025refocus, qiao2025lipt} before generating a final answer, and some integrate external knowledge to support higher-level reasoning.
More recently, reinforcement learning (RL) has been used to strengthen multimodal reasoning, inspired by systems such as OpenAI-o1/o3 and DeepSeek-R1~\cite{o1, o3, guo2025deepseekr1}.  RL- and GRPO-based methods have been applied to multimodal math and science problems~\cite{huang2025visionr1, wei2025unsupervised, wei2025advancing, chen2025r1v}, general visual reasoning~\cite{meng2025mm, peng2025lmm}, and open-ended visual grounding~\cite{shen2025vlmr1, liu2025visualrft, jia2025multi}.  Other works emulate human-like “thinking with images” by allowing dynamic cropping, zooming, or multi-step attention over images~\cite{liu2024chainofspot, shao2024visualcot, wang2025vgr, fan2025grit, zheng2025deepeyes, cao2025groundr1, liu2025visionreasoner, su2025pixelreasoner}.  However, most of these approaches still supervise only the final answer, leaving intermediate visual focus and reasoning trajectories implicit and difficult to audit.  In this work we follow the general direction of multimodal chain-of-thought and RL-based training, but introduce an explicit, object-centric reasoning process whose details are presented in Sec.~\ref{sec:methodology}.

\textbf{Benchmarks for Multimodal Large Language Models}
To assess the capabilities of modern MLLMs, numerous benchmarks have been proposed beyond classical captioning~\cite{lin2014microsoft} and VQA~\cite{antol2015vqa}.  General-purpose suites such as MME~\cite{fu2024mme} and MMBench~\cite{liu2024mmbench} probe perception, commonsense, and language understanding with diverse question formats, while SEED-Bench~\cite{li2024seed} and MM-Vet~\cite{yu2023mm} extend evaluation to multi-image/video inputs, OCR, and mathematical reasoning.  Higher-level benchmarks including MMMU~\cite{yue2024mmmu}, HallusionBench~\cite{guan2024hallusionbench}, and MathVista~\cite{lu2023mathvista} focus on expert knowledge, hallucination, or math-centric visual reasoning.  These datasets provide a broad view of MLLM competence but mostly use relatively clean images and evaluate only final answers, without requiring models to expose intermediate reasoning.

A complementary line of work targets more challenging visual conditions or spatially complex scenes.  POPE~\cite{li2023pope} diagnoses object hallucination under controlled perturbations, and V*~\cite{wu2024vstar} evaluates attributes and pairwise spatial relations, albeit mainly on COCO-derived imagery~\cite{lin2014microsoft}.  MME-RealWorld~\cite{zhang2024mmerw} and HR-Bench~\cite{wang2025hrbench} introduce high-resolution, cluttered scenes but still treat grounding implicitly.  On top of driving and urban-perception datasets such as BDD100K~\cite{yu2020bdd100k} and nuScenes~\cite{caesar2020nuscenes}, several multimodal corpora enrich real-world scenes with language: BDD-X~\cite{kim2018textual} and GOHD~\cite{xu2020explainable} provide behavior rationales and hazard explanations; DriveRef~\cite{wu2023language} studies referring expressions; nuScenes-QA/MQA~\cite{qian2024nuscenes, inoue2024nuscenes} scale visual QA; and DriveLM~\cite{sima2023drivelm} offers graph-structured dialogues.  Other datasets emphasize commonsense~\cite{marcu2024lingoqa}, robustness and trust~\cite{xing2024autotrust}, or rare hazards~\cite{li2022coda, chen2025automated}.  These corpora underscore the importance of complex, real-world scenes, but generally provide answer- or sentence-level supervision and seldom disentangle perceptual grounding from subsequent reasoning quality.


\section{Construction of AD$^2$-Bench}
\label{sec:method_bench}

We introduce \textbf{AD$^2$-Bench}, a comprehensive benchmark designed to evaluate the \emph{hierarchical cognitive capabilities} of Multimodal Large Language Models within complex urban scenarios. Unlike existing datasets that focus primarily on benign conditions or end-to-end control commands, AD$^2$-Bench emphasizes the \emph{process of evidence-based reasoning} under visual degradation, bridging the gap between perceptual grounding and actionable decision.

\subsection{Complex Scenario Curation}
\label{subsec:data_collection}

To rigorously assess perceptual robustness, we move beyond synthetic or clear-weather datasets found in standard benchmarks. We curated a large-scale collection of real-world images capturing genuine visual degradation from diverse sources, including CODA~\cite{li2022coda}, ACDC~\cite{sakaridis2021acdc}, DAWN~\cite{kenk2020dawn}, BDD100K~\cite{yu2020bdd100k}, nuScenes~\cite{caesar2020nuscenes}, and diverse web repositories. This multi-source strategy ensures that the visual patterns cover a wide spectrum of sensor noise and environmental occlusions.

The dataset is structured into seven distinct adverse condition categories: \emph{Rainy, Snowy, Foggy, Sandstorm, Nighttime, Dawn, and Overcast}. 
While the \emph{Rainy}, \emph{Foggy}, and \emph{Snowy} classes emphasize signal degradation, the \emph{Daytime} and \emph{Dawn} scenes focus on relational complexity, featuring dense object occlusion and intricate road layouts that demand deep semantic parsing.

\textbf{Rigorous Filtering Pipeline.}
Ensuring data quality in adverse conditions is challenging due to the ambiguity of visual features. We employed a rigorous multi-stage filtering protocol. First, we utilized advanced MLLMs (Gemini 2.5-Pro) to perform automated pre-filtering, scoring image complexity to exclude samples with low reasoning value. Subsequently, domain experts conducted a final manual audit to verify that each image contains sufficient visual cues for logical inference, rejecting samples with unrecognizable semantics or insufficient resolution.


\begin{figure*}[t]
    \centering
    \includegraphics[width=1\linewidth]{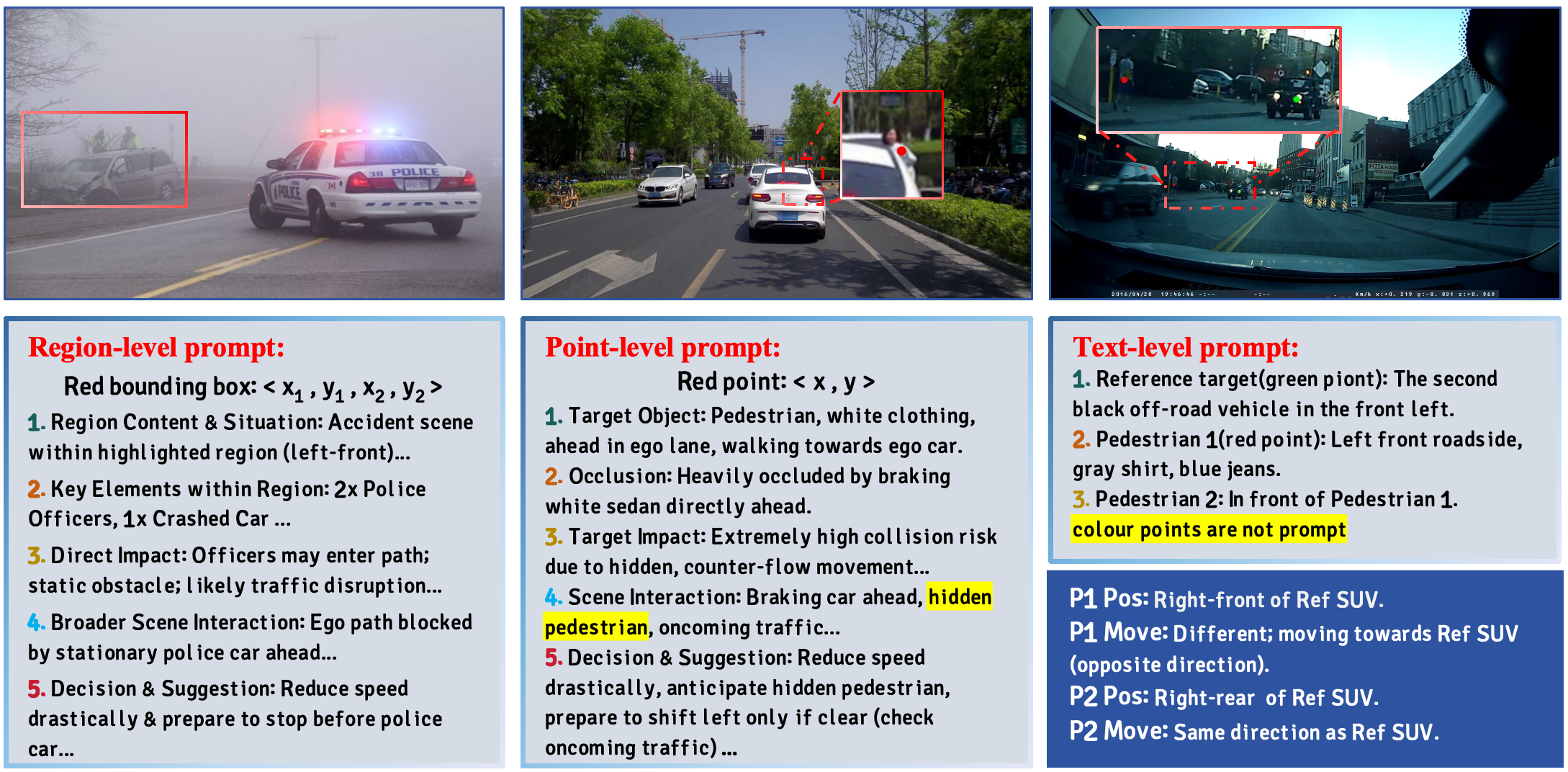}
    \vspace{-10pt}
    \caption{
    To diagnose interactive reasoning capabilities and facilitate precise evidence acquisition in adverse scenarios, we introduce a multi-granularity prompt: Region-level cues isolate contextual interactions, while Point-level cues pinpoint fragmented evidence in heavy occlusion where boxes fail. Text-level cues resolve semantic ambiguity. Addressing the limitation of MLLMs in processing coordinate text, we render these prompts as explicit visual overlays to enforce spatial attention.
    }
    \label{fig:4}
    \vspace{-10pt}
\end{figure*}

\subsection{The Atomic Annotation Protocol}
\label{subsec:atomic_annotation}

A core contribution of AD$^2$-Bench is the \textbf{Atomic Annotation} paradigm, designed to evaluate the explicit reasoning process. Unlike traditional VQA datasets that only provide a final answer, we formalize the annotation of the entire evidence chain and multi-Granularity vision prompts, enabling step-by-step diagnostic evaluation.

\textbf{Vision Prompt Strategy.}
In complex scenes, critical visual evidence is often submerged in environmental noise or fragmentation. To bridge the gap between perception and reasoning while diagnosing interactive capabilities, we employ three granular prompt levels as illustrated in Fig.~\ref{fig:4}, serving as \emph{explicit attentional prompts}.
Single image provide no cues, serving as an unguided baseline.
\emph{Region-level} prompts utilize bounding boxes to define a contextual scope, guiding the model to extract evidence regarding multi-object interactions.
Critically, we introduce \emph{Point-level} prompts to capture atomic evidence in dense occlusion scenarios. Unlike bounding boxes that suffer from significant overlap, point prompts function as spatial pins, isolating the visible fragments of targets.
Finally, \emph{Text-level} prompts provide referential grounding to resolve semantic ambiguities.
Since MLLMs struggle with raw coordinate text, we render these prompts directly onto the images to ensure  spatial attention. 
More details are in Appendix~\ref{ann_format}.

\textbf{Formal Definition.}
We define each annotation record $P_i$ as a six-tuple to capture the full context of a reasoning step:
\begin{equation}
\small
    P_i = (Q_i,\, C_i,\, I_i,\, L_i,\, B_i,\, A_i)
\end{equation}
where $Q_i$ is the question, $C_i \in [1, 33]$ denotes the sub-task index covering the full spectrum from atomic perception to holistic decision, $I_i$ is the image, $L_i$ specifies the prompt type (text/box/point), $B_i$ contains box coordinates, and $A_i$ is the atomic ground-truth answer. This formalization ensures that every reasoning step is anchored to specific visual and textual evidence.

\textbf{The Atomic Flow Strategy.}
The annotation process is executed through a rigorous \emph{Atomic Flow} collaboration strategy to ensure logical consistency across the chain of evidence. The entire process was conducted in four batches, involving domain experts overseen by independent lead annotators. 

Initially, the complex scene decision task is subjected to \emph{Atomic Decomposition}, breaking it down into a Directed Acyclic Graph (DAG) of sub-questions. This separates \emph{Perception Atoms} (identifying visual elements) from \emph{Reasoning Atoms} (interpreting their implications). Cluster analysis was employed to refine the four evaluation axes into 33 distinct sub-tasks to ensure coverage.

To maintain high fidelity, we implemented a \emph{Machine-Assisted Validation} phase. Initial atomic annotations are rigorously scored by an ensemble of advanced models (Gemini 2.5-Pro, GPT-4o, Qwen 2.5-Max) against the visual input. Discrepancies between human labels and model consensus trigger a mandatory review flag.

Finally, the \emph{Expert Cross-Pollination} mechanism ensures narrative coherence. Experts are dynamically grouped where specialists in perception verify the grounding coordinates ($B_i$), while specialists in logic verify the causal link to the answer ($A_i$). This cross-check ensures that the final \emph{Intelligence Atoms} are not only factually correct but also narratively coherent, providing a reliable ground truth for measuring the alignment between machine reasoning and human cognition.


\subsection{Analysis of Evaluation Dimensions}
\label{subsec:data_analysis}

\begin{figure*}[t]
    \centering
    \includegraphics[width=1\linewidth]{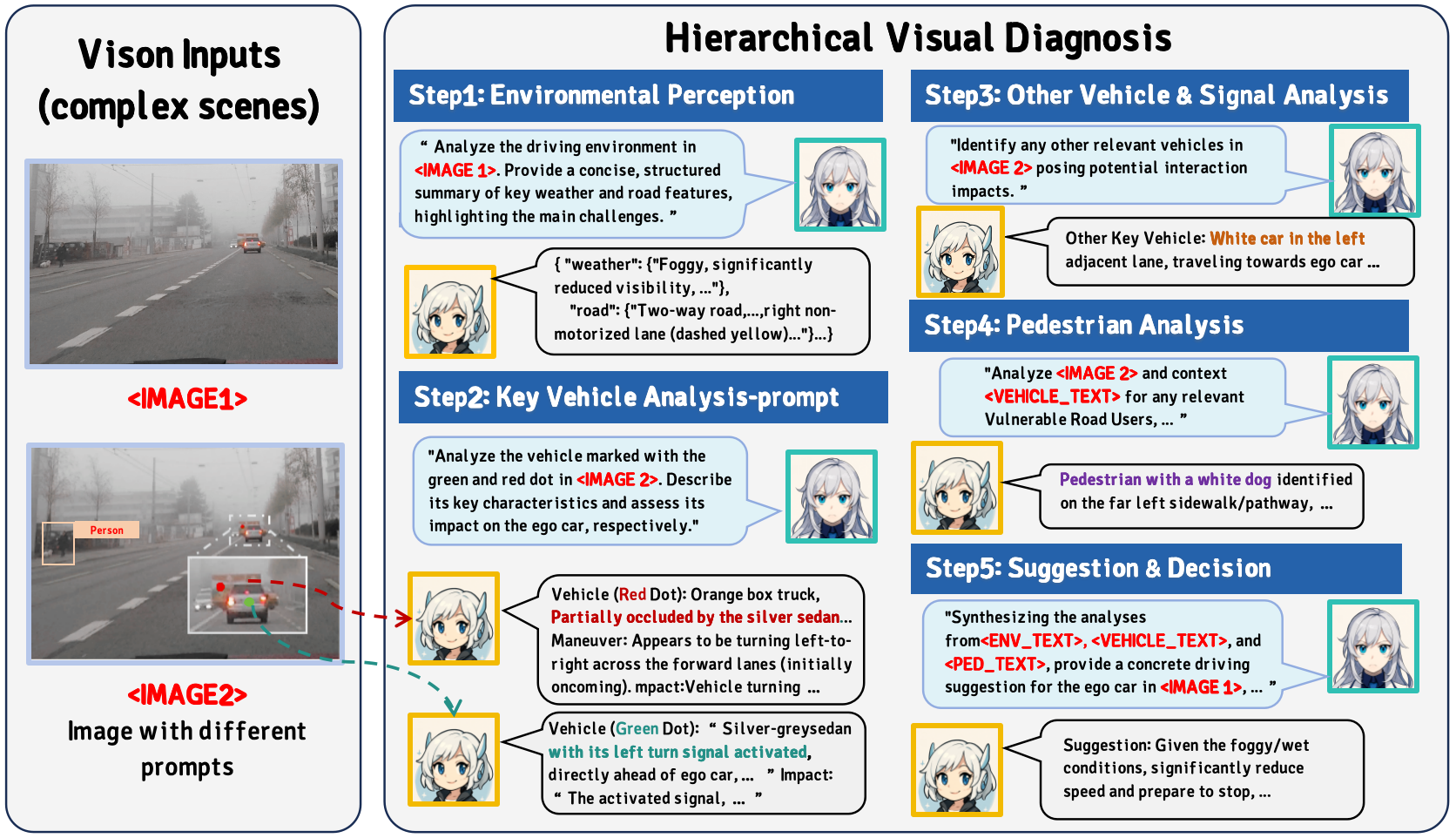}
    \vspace{-10pt}
    \caption{The \textbf{Hierarchical Diagnosis Framework} in AD$^{2}$-Bench guides MLLMs through a structured \textbf{Chain of Evidence (CoE)} process. This pipeline decomposes reasoning into adaptable stages, typically including perception (\textbf{basic \& advanced}), various analyses like key vehicle/pedestrian (\textbf{relation understanding \& reasoning}), and a final suggestion (\textbf{decision}). The vision prompt in Step 2 can be \textbf{optioned}, and image1 and image2 are the same image displayed in different steps. This structured decomposition serves two key purposes: it enhances the MLLM's ability to capture fine-grained scene details progressively, and critically, it facilitates reliable, atomic-level annotation and verification of each reasoning step on our end, ensuring high accuracy for crucial scene elements annotation.
    }
    \label{fig:2}
    \vspace{-10pt}
\end{figure*}

\textbf{Basic Perception (Level 1: Atomic Visual Grounding).}
In complex scenes, trustworthy decision-making is predicated on the precise acquisition of \textbf{visual evidence}. Before any high-level reasoning can occur, the model must filter environmental noise to establish valid physical focus for relevant entities.
Therefore, AD$^2$-Bench isolates a \emph{basic-perception} tier designed to evaluate the model's capacity to reliably extract valid foreground objects from cluttered or degraded visual backgrounds. This tier specifically evaluates: (1) \emph{Object Presence}, determining the existence of entities despite visual obstructions like rain or fog; (2) \emph{Precise Localization}, anchoring vulnerable road users (pedestrians, cyclists) to exact pixel coordinates to prevent hallucination; and (3) \emph{Category-aware Retrieval}, identifying specific targets via text-prompted queries. To ensure the reliability of GT, we mandate a rigorous cross-validation process by at least three domain experts.

\textbf{Advanced Perception (Level 2: Semantic Parsing).}
Beyond mere localization, robust reasoning requires \textbf{verifying the semantic details} of the evidence. This tier targets the deep decoding of attributes amidst visual degradation, ensuring that the captured evidence is semantically accurate rather than hallucinated.
As illustrated in Step 1 of Fig.~\ref{fig:2}, tasks include: (1) \emph{Attribute Recognition}, identifying states such as vehicle lights or pedestrian attire; (2) \emph{Dense Captioning}, generating context-rich descriptions; (3) \emph{Optical Character Recognition (OCR)}, decoding traffic signs; and (4) \emph{Lane Topology Analysis}, understanding road geometry. These problems are specifically crafted to stress-test visual encoders against real-world perturbations, such as strong glare in dawn scenarios or optical scattering in rainy night scenes. To guarantee annotation quality, twenty experts independently supply answers, followed by a two-stage expert verification to ensure semantic consistency.

\textbf{Relation Understanding (Level 3: Contextual Interaction).}
Effective evidence is not isolated; it is structurally connected. This tier bridges atomic perception and high-level logic by extracting \textbf{contextual evidence}, requiring the model to observe explicit interactions between entities to form a coherent scene graph. We define four distinct interaction types:
(1) \emph{Spatial Relations}, evaluating both ego-centric spatial awareness (target relative to ego) and allocentric relationships (layout between external objects); (2) \emph{Social \& Functional Interactions}, testing the ability to perceive group dynamics (e.g., ``parent-child'' pairs) and functional units (e.g., ``rider-bicycle'') based on fine-grained action cues; (3) \emph{Causal Relations}, inferring how dynamic scene changes (e.g., traffic light state) dictate target behaviors; and (4) \emph{Occlusion Status}, assessing \emph{Visibility and Depth Reasoning} to identify occlusion sources in dense traffic. Each relation type undergoes independent annotation and mutual cross-checks to ensure logical coherence.

\textbf{Reasoning and Decision (Level 4: Holistic Synthesis).}
This tier represents the apex of the cognitive pyramid, requiring the integration of perceptual evidence and relational context to formulate \emph{Actionable Decisions}. Leveraging the dataset's adverse condition scenarios, we curated a dedicated 5,406-image Chain of Evidence test suite. The reasoning challenges are comprehensively designed to cover: 
(1) \emph{Weather Reasoning}, inferring physical constraints like friction and visibility from visual cues; (2) \emph{Night Reasoning}, assessing risks from non-uniform lighting and glare; (3) \emph{Region-level Reasoning}, applying spatial logic within specific zones (e.g., multi-object \& accident sites); (4) \emph{Occlusion Reasoning}, using point prompts to infer the trajectory and risk of partially visible objects; (5) \emph{Cross-role Reasoning}, predicting the intent of other agents from their perspective; and (6) \emph{Holistic Reasoning}, synthesizing all evidential elements to provide a safe, interpretable action decision for the ego-object, completing the perception-to-action loop.

\subsection{Evaluation Framework: A Decoupled Cognitive Diagnosis}
\label{subsec:metrics}

To align with the hierarchical cognitive framework, we establish a comprehensive evaluation that spans from atomic perception to holistic reasoning. Unlike traditional benchmarks that rely solely on final-answer accuracy, our framework employs a multi-dimensional metric system designed to diagnose specific failures in the cognitive pipeline.

\paragraph{Metrics for General Task.}
For the foundational tiers of \textbf{Basic Perception} and \textbf{Advanced Perception}, we utilize standard quantitative metrics to assess discriminative and spatial precision. For binary classification tasks such as object existence and multiple-choice selection, we employ \emph{Binary Accuracy}, defined as:
\begin{equation}
\small
acc_i = \mathbb{I}(A_i = y_i)
\end{equation}
where $A_i$ and $y_i$ represent the predicted and ground-truth answers, respectively.

To evaluate spatial grounding capabilities, we utilize distinct metrics for points and regions. For pixel-level object localization, the accuracy is penalized by the Euclidean distance between the predicted coordinate $\mathbf{x}_i$ and the ground truth $\mathbf{x}_i^{\text{gt}}$, formulated as:
\begin{equation}
\small
    acc_i = \frac{1}{1 + \alpha_p \| \mathbf{x}_i - \mathbf{x}_i^{\text{gt}} \|_2}
\end{equation}
where $\alpha_p=0.005$ is a scaling factor. For object detection, we employ the standard Intersection over Union (IoU) metric to measure the overlap between predicted ($B_i$) and ground-truth ($B_i^{\text{gt}}$) bounding boxes:
\begin{equation}
\small
    acc_i = \text{IoU}(B_i, B_i^{\text{gt}}) = \frac{|B_i \cap B_i^{\text{gt}}|}{|B_i \cup B_i^{\text{gt}}|}
\end{equation}

Furthermore, for the semantic decoding tasks involving Optical Character Recognition (OCR), we assess transcriptional quality using two complementary metrics. The \emph{Character Error Rate (CER)} quantifies the Levenshtein distance normalized by string length:
\begin{equation}
\small
    CER = \frac{S + D + I}{N}
\end{equation}
where $S, D, I$ represent the number of substitutions, deletions, and insertions, and $N$ is the total character count. Concurrently, the \emph{Character-Level F1-score} balances precision ($P$) and recall ($R$) at the character level, computed as:
\begin{equation}
\small
    F_1 = 2 \cdot \frac{P \cdot R}{P + R}
\end{equation}

\textbf{Metrics for Relation Understanding.}
We evaluate contextual interaction via \textbf{Accuracy} on multiple-choice queries (see Tab.~\ref{tab:prompt}). By requiring the model to select correct descriptors amidst adversarial distractors, this metric rigorously verifies the establishment of valid scene graph edges (e.g., spatial or relation) as a prerequisite for higher-order reasoning.

\paragraph{Metrics for Trustworthy Reasoning.}
To assess the \textbf{Trustworthiness} of the \emph{Chain of Evidence (CoE)}, we move beyond simple accuracy to evaluate the reliability, interpretability, and calibration of the reasoning process. We propose four orthogonal metrics that operate on the generated evidence chain. 
Formally, let $Q$ be the query, $A=\{A_i\}_{i=1}^N$ be the model's answer structured as a sequential evidence chain, and $GT=\{GT_i\}_{i=1}^N$ be the ground truth. Each metric employs a scoring function $\operatorname{Eval}_{\cdot}$ mapping inputs to a scalar in $[0,1]$ ($1$: perfect; $0$: failure).

In practice, we instantiate these $\operatorname{Eval}_{\cdot}$ functions using a strong automatic evaluator (an advanced LLM judge, GPT-5.2). The evaluator is asked to rate each item on a discrete 1--10 Likert scale following a task-specific rubric, and we linearly normalize this score to $[0,1]$ before aggregation. 

\textbf{1. Hallucination Resistance Score (HRS): Visual Fidelity.} This metric measures the model's resistance to visual hallucinations. It verifies the \emph{factual correctness} of each evidence $A_i$ generated by model against the ground truth $GT_i$, focusing on whether $A_i$ only mentions entities, attributes, and relations that are actually supported by the image and the annotated CoE. The fidelity score is high when $A_i$ is fully grounded and contains no hallucinated content, and low when $A_i$ fabricates non-existent objects or attributes or contradicts $GT_i$. The overall HRS is defined as the average fidelity across the chain:
\begin{equation}
\label{eq:hrs}
\small
HRS = \frac{1}{N} \sum_{i=1}^{N} \operatorname{Eval}_{\text{fidelity}}(A_i, GT_i).
\end{equation}

\textbf{2. Logical Reliability Score (LRS): Inferential Validity.} To assess the robustness of the reasoning process, LRS evaluates the validity of each local transition $A_i \rightarrow A_{i+1}$. Crucially, this metric is \emph{truth-agnostic}: $\operatorname{Eval}_{\text{logic}}(A_i, A_{i+1}) \in [0,1]$ measures whether $A_{i+1}$ follows logically from $A_i$ \emph{assuming $A_i$ were true}, without consulting the ground-truth visual world. Scores close to $1$ correspond to logically coherent, non-contradictory inferences; scores close to $0$ indicate non sequiturs, unjustified leaps, or explicit logical conflicts. This isolates “thinking wrong” (logic failure) from “seeing wrong” (perception failure), and the chain-level LRS is given by
\begin{equation}
\label{eq:lrs}
\small
LRS = \frac{1}{N-1} \sum_{i=1}^{N-1} \operatorname{Eval}_{\text{logic}}(A_i, A_{i+1}).
\end{equation}

\textbf{3. Self-Consistency Score (SCS): Narrative Stability.} Trustworthy models must not contradict themselves over long contexts. SCS assesses the \emph{global} consistency of the entire chain $A$, detecting self-contradictions, temporal or spatial inconsistencies, and memory lapses. The function returns a high score when the sequence of evidence atoms can be interpreted as a single coherent narrative without internal conflicts (e.g., an object does not change category or location inexplicably), and a low score when multiple atoms in $A$ are mutually incompatible. We define
\begin{equation}
\label{eq:scs}
\small
SCS = \operatorname{Eval}_{\text{consistency}}(A_1, \dots, A_N).
\end{equation}
SCS therefore captures the stability of the model's internal world model during complex multi-step reasoning.

\textbf{4. Explainability Alignment Score (EAS): Decision Justification.} This metric evaluates the \emph{interpretability} of the final decision by checking whether it is causally and logically supported by the preceding evidence. Specifically, the metric measures to what extent the actionable decision $A_N$ can be derived as a justified conclusion from the accumulated evidence $(A_1, \dots, A_{N-1})$. A score of $1$ indicates that the decision is a natural, well-supported consequence of the chain, while a score near $0$ indicates that $A_N$ behaves like a black-box guess that ignores or contradicts the prior evidence.
\begin{equation}
\label{eq:eas}
\small
EAS = \operatorname{Eval}_{\text{alignment}}((A_1, \dots, A_{N-1}), A_N).
\end{equation}

By decomposing evaluation into these orthogonal dimensions, we provide a granular diagnosis of the trustworthiness visual reasoning. 

\section{Methodology}
\label{sec:methodology}

AD$^2$-Bench (Sec.~\ref{sec:method_bench}) reveals a critical dependency: robust reasoning in complex scenes is predicated on the strict acquisition of atomic evidence. 
Standard end-to-end MLLMs, however, rely on implicit inference lacking these visual anchors, leading to severe perception-logic disconnects. 
To bridge this gap, we propose \textbf{Evidence-Grounded Visual Reasoning} (EGVOR). 
Inspired by the ``Atomic Annotation'' paradigm, EGVOR reformulates reasoning as the explicit generation of verifiable evidence chains, optimized through a hierarchical curriculum of reflective supervision and cognitive reinforcement learning.

\subsection{Preliminaries: Reasoning in Complex and Adverse Environments}
\label{subsec:preliminaries}

We first formalize visual reasoning in complex scenarios from a probabilistic perspective. The goal of this analysis is not to provide an exact generative model of MLLMs, but to construct a diagnostic abstraction that identifies how adverse conditions destabilize implicit reasoning and why explicit evidence construction mitigates this instability.

\paragraph{Visual Perception under Signal Degradation.}
In AD$^2$-Bench, the model observes imperfect sensory data. Let $I_{\mathrm{clean}}$ denote the latent, ideal visual scene. The observed image $\tilde{I}$ is modeled as a corrupted transformation:
\begin{equation}
\small
    \tilde{I} = \phi(I_{\mathrm{clean}}, \xi_{\mathrm{env}}),
\end{equation}
where $\xi_{\mathrm{env}}$ represents environmental degradation variables such as rain streaks, fog, low illumination, or occlusion, and $\phi$ denotes the corresponding physical degradation process. For each visual entity $o$ in the scene, we denote by $z_o$ its latent feature representation extracted from $\tilde{I}$. Throughout this section, $z_o$ is treated as the deterministic representation associated with entity $o$ under the observed image, while residual feature uncertainty is modeled by the random representation $Z$ in the variance analysis. Under adverse degradation, $z_o$ may lose discriminative cues, making the mapping from $(\tilde{I},Q)$ to the semantic answer $A$ underdetermined.

\paragraph{Modeling Scope and Object-Centric Factorization.}
\textcolor{black}{Let $\mathcal{O}$ denote the query-dependent set of candidate visual entities in the scene. We introduce $O_f$ as the latent visual-focus random variable whose realizations are elements of $\mathcal{O}$. A lowercase $o\in\mathcal{O}$ denotes one possible realization of $O_f$. For notational simplicity, we write $P(o|\tilde{I},Q)$ as shorthand for $P(O_f=o|\tilde{I},Q)$. Therefore, the summation $\sum_{o\in\mathcal{O}}$ marginalizes over all possible realizations of the latent visual focus $O_f$.}

\textcolor{black}{For a query $Q$, we approximate the answer posterior by marginalizing over this latent visual focus:
\begin{equation}
\label{eq:total_prob}
\small
P(A|\tilde{I}, Q)
=
\sum_{o \in \mathcal{O}}
P(A|z_o, Q) \cdot P(o|\tilde{I}, Q).
\end{equation}
This factorization is a modeling approximation: $P(o|\tilde{I},Q)$ specifies where the model focuses, while $P(A|z_o,Q)$ specifies what answer is supported by the feature representation at that focus.}

\textcolor{black}{For a specific query $Q$, the candidate entity set is dynamically partitioned into the \textbf{Critical Object Set} $\mathcal{O}_{crit}$, the \textbf{Irrelevant Set} $\mathcal{O}_{irr}$, and the \textbf{Background Set} $\mathcal{O}_{bg}$. We further define $\mathcal{O}_{non}=\mathcal{O}_{irr}\cup\mathcal{O}_{bg}$ and $\mathcal{O}=\mathcal{O}_{crit}\cup\mathcal{O}_{non}$. Using this partition, Eq.~\ref{eq:total_prob} can be decomposed as
\begin{equation}
\label{eq:path_decomposition}
\footnotesize
\begin{aligned}
P(A|\tilde{I}, Q)
&=
\underbrace{
\sum_{o \in \mathcal{O}_{crit}}
P(A|z_o,Q) P(o|\tilde{I}, Q)
}_{\text{\textcircled{1} Effective Term}}
\\
&\quad+
\underbrace{
\sum_{o \in \mathcal{O}_{irr}}
P(A|z_o,Q) P(o|\tilde{I}, Q)
}_{\text{\textcircled{2} Distraction Term}}
\\
&\quad+
\underbrace{
\sum_{o \in \mathcal{O}_{bg}}
P(A|z_o,Q) P(o|\tilde{I}, Q)
}_{\text{\textcircled{3} Hallucination Term}}.
\end{aligned}
\end{equation}
The Effective Term contains query-relevant evidence that can support the correct answer. The Distraction Term corresponds to visually plausible but question-irrelevant entities. The Hallucination Term corresponds to background or environmental noise that does not provide reliable answer-specific semantics.}

\paragraph{Critical-Evidence Mass and Posterior Suppression.}
\textcolor{black}{The key effect of adverse degradation is not merely that the visual focus becomes diffuse, but that probability mass is transferred from query-critical evidence to non-critical or weakly informative regions. To formalize this, let $A_{gt}$ denote the visually grounded correct answer, and define the critical-evidence mass:
\begin{equation}
\label{eq:rho_main}
\small
\rho
=
P(O_f\in\mathcal{O}_{crit}|\tilde{I},Q)
=
\sum_{o\in\mathcal{O}_{crit}} P(o|\tilde{I},Q).
\end{equation}
Under the critical-evidence separability assumption, query-critical evidence supports $A_{gt}$ more reliably than non-critical evidence. Appendix~\ref{app:prob_proofs} shows that, for two local focus distributions with critical-evidence masses $0<\rho_2<\rho_1<1$, the posterior support for the grounded answer satisfies
\begin{equation}
\label{eq:posterior_drop_main}
\small
P_{\rho_2}(A_{gt}|\tilde{I},Q)
<
P_{\rho_1}(A_{gt}|\tilde{I},Q).
\end{equation}
Thus, the core failure mechanism is a reduction in grounded posterior support: the model assigns less probability to the correct visually supported answer when critical evidence mass is reallocated to distractors or background regions.}

\paragraph{The Failure of Implicit Aggregation: From Dilution to Prior Collapse.}
\textcolor{black}{Standard end-to-end MLLMs do not explicitly separate the Effective, Distraction, and Hallucination Terms. Instead, they compress the scene into a global context vector via soft attention:
\begin{equation}
\label{eq:aggregation_failure}
\small
z_{\mathrm{global}}
=
\sum_{o\in\mathcal{O}}
\omega_o z_o,
\quad
\omega_o=P(o|\tilde{I},Q).
\end{equation}
When $\omega_o$ assigns substantial mass to $\mathcal{O}_{irr}$ or $\mathcal{O}_{bg}$, the global feature is no longer dominated by $z_{crit}$, but becomes a mixture of critical evidence, distractors, and background noise.}

\textcolor{black}{Let $Z_{\mathrm{global}}$ denote the random variable corresponding to the global visual representation, and let $z_{\mathrm{global}}$ denote its realization. In the limiting noise-dominant regime, the global visual feature carries little answer-specific information. We express this as approximate conditional independence:
\begin{equation}
\label{eq:cond_indep}
\small
Z_{\mathrm{global}}
\mathrel{\perp\!\!\!\perp}
A
\mid Q.
\end{equation}
In the exact conditional-independence case, $P(z_{\mathrm{global}}|A,Q)=P(z_{\mathrm{global}}|Q)$. Applying Bayes' rule gives
\begin{equation}
\label{eq:bayes_collapse}
\small
\begin{aligned}
P(A|z_{\mathrm{global}},Q)
&=
\frac{
P(z_{\mathrm{global}}|A,Q)P(A|Q)
}{
P(z_{\mathrm{global}}|Q)
}
\\
&=
P(A|Q).
\end{aligned}
\end{equation}
In practice, the independence is approximate rather than exact; hence the posterior moves toward the language prior, $P(A|z_{\mathrm{global}},Q)\approx P(A|Q)$. Let $A_{\mathrm{prior}}$ denote the answer favored by $P(A|Q)$. In failure cases where $A_{\mathrm{prior}}\neq A_{gt}$, prior collapse may yield
\begin{equation}
\label{eq:inequality}
\small
P(A_{\mathrm{prior}}|Q)
>
P(A_{gt}|\tilde{I}, Q).
\end{equation}
This inequality characterizes the prior-dominant hallucination regime, where grounded visual evidence is suppressed and the model falls back to language priors.}

\paragraph{Bottleneck Analysis: Semantic Entropy and Variance Drift.}
\label{sec:variance_analysis}
\textcolor{black}{We now summarize the two statistical bottlenecks induced by the above failure mechanism.}

\textcolor{black}{\noindent\textbf{1. Posterior Suppression and Semantic Uncertainty.}
We define the entropy of the latent visual-focus distribution as
\begin{equation}
\label{eq:focus_entropy}
\small
H(O_f|\tilde{I},Q)
=
-
\sum_{o\in\mathcal{O}}
P(o|\tilde{I},Q)
\log P(o|\tilde{I},Q).
\end{equation}
The entropy is taken over the random variable $O_f$, whose realizations are candidate entities $o\in\mathcal{O}$. We do not assume an unconditional implication from visual-focus entropy to answer entropy. Instead, Appendix~\ref{app:prob_proofs} proves that a reduction in critical-evidence mass suppresses $P(A_{gt}|\tilde{I},Q)$, and further derives the answer entropy under a mixture model:
\begin{equation}
\label{eq:answer_entropy_main}
\small
H_{\rho}(A|\tilde{I},Q)
=
-
\sum_{a\in\mathcal{A}}
p_{\rho}(a)\log p_{\rho}(a),
\end{equation}
where $\mathcal{A}$ is the answer space and $p_{\rho}(a)$ is the answer distribution induced by critical-evidence mass $\rho$. In the uncertainty-dominant failure regime, non-critical regions induce diverse and mutually inconsistent answer hypotheses; in that local regime, decreasing $\rho$ increases $H_{\rho}(A|\tilde{I},Q)$. In the prior-dominant hallucination regime, however, the model may be confidently wrong, and the failure is better captured by Eq.~\ref{eq:inequality}. Thus, semantic entropy increase and prior collapse are complementary manifestations of the same underlying issue: dilution of critical visual evidence.}

\textcolor{black}{\noindent\textbf{2. Variance Decomposition via Latent Anchoring.}
To characterize the instability of the latent context, we treat $Z$ as a random representation induced by the latent focus distribution. Since $Z$ may be vector-valued, $\operatorname{Var}(Z)$ denotes either the scalar variance of a projected feature statistic or the trace of the covariance matrix. All expectations and variances over $o$ are taken with respect to
\begin{equation}
\small
O_f\sim P(\cdot|\tilde{I},Q).
\end{equation}
By the Law of Total Variance,
\begin{equation}
\label{eq:variance_decomp}
\small
\begin{aligned}
\operatorname{Var}(Z|\tilde{I},Q)
&=
\underbrace{
\mathbb{E}_{O_f\sim P(\cdot|\tilde{I},Q)}
\left[
\operatorname{Var}(Z|O_f,\tilde{I},Q)
\right]
}_{\text{Intrinsic Semantic Uncertainty}}
\\
&\quad+
\underbrace{
\operatorname{Var}_{O_f\sim P(\cdot|\tilde{I},Q)}
\left(
\mathbb{E}[Z|O_f,\tilde{I},Q]
\right)
}_{\text{Spatial Boundary Ambiguity}}.
\end{aligned}
\end{equation}
The first term captures feature-level instability caused by environmental degradation. Even if the correct entity is localized, its representation may remain noisy due to fog, rain, low illumination, or occlusion. The second term captures instability in the visual focus itself. When $O_f$ ranges over $\mathcal{O}_{crit}$, $\mathcal{O}_{irr}$, and $\mathcal{O}_{bg}$, the conditional expectation $\mathbb{E}[Z|O_f,\tilde{I},Q]$ varies across semantically unrelated regions, resulting in a high-variance mixture rather than a stable semantic anchor.}

\paragraph{Chain of Evidence (CoE) Formulation.}
\textcolor{black}{To address these bottlenecks, we propose \textbf{Chain of Evidence (CoE)}, which reformulates reasoning as a sequential process of \textbf{Spatio-Semantic Verification}. Valid evidence must satisfy a dual constraint: it should be spatially anchored to the Critical Set $\mathcal{O}_{crit}$ and semantically verified against the visual content.}

\textcolor{black}{Mathematically, CoE performs a Component Filtration operation. We define a structured evidence set
\begin{equation}
\small
E=\{e_1,\dots,e_K\}\subseteq\mathcal{O}_{crit},
\end{equation}
where each $e\in E$ denotes a critical object or region whose spatial location and semantic content are both valid for the current query. Not every critical object must be selected; $E$ contains only the evidence necessary for the reasoning task. Let $E'=\mathcal{O}\setminus E$ be the complement set containing objects that fail the evidence constraints. CoE decomposes the answer posterior as
\begin{equation}
\label{eq:full_expansion}
\footnotesize
\begin{aligned}
P(A|\tilde{I},Q)
&=
\underbrace{
\sum_{e\in E}
P(A|z_e,Q)P(e|\tilde{I},Q)
}_{\text{Signal Term}}
\\
&\quad+
\underbrace{
\sum_{e'\in E'}
P(A|z_{e'},Q)P(e'|\tilde{I},Q)
}_{\text{Interference Term}}.
\end{aligned}
\end{equation}
CoE performs Component Filtration by increasing the relative contribution of the Signal Term and suppressing the Interference Term:
\begin{equation}
\label{eq:coe_factorization}
\small
P_{\mathrm{CoE}}(A|\tilde{I},Q)
\approx
\sum_{e\in E}
P(A|z_e,Q)P(e|\tilde{I},Q).
\end{equation}
This approximation does not claim that all interference disappears. Rather, it states the intended effect of CoE: to bias reasoning toward spatially grounded and semantically verified evidence, thereby reducing the propagation of environmental noise into the reasoning process.}

\subsection{Evidence-Grounded Visual Reasoning}
\label{subsec:ce_vgr}

Inspired by the rigorous ``Atomic Annotation'' paradigm and the variance analysis in Sec.~\ref{sec:variance_analysis}, we propose \textbf{Evidence-Grounded Visual Reasoning (EGVOR)}. 
Unlike standard black-box MLLMs susceptible to high variance in the global context $z_{global}$, EGVOR reformulates visual reasoning not as a direct mapping, but as a sequential Variance Reduction Process. We introduce the concept of \emph{Implicit Evidence Atoms} to concretely instantiate the Component Filtration in Eq.~\ref{eq:coe_factorization}, explicitly bridging the gap between high-level semantic reasoning and low-level visual stochasticity.

\paragraph{Definition 1: The Atomic Evidence Triplet.}
We formalize the fundamental unit of reasoning, the evidence atom $a_t$, as a structured realization of the random variables defined in our probabilistic framework. An atom at step $t$ is a triplet:
\begin{equation}
\small
    a_t = \langle b_t, \mathbf{h}_t, c_t \rangle .
\end{equation}
\noindent where,
Explicit Spatial Anchor ($b_t$) is a deterministic realization of the stochastic visual focus $o$ introduced in Sec.~\ref{subsec:preliminaries}. Specifically, $b_t$ parameterizes an axis-aligned bounding box in the image coordinate space that is expected to contain one or more candidate objects from $\mathcal{O}$. By strictly enforcing $b_t$ to bound such a region, we truncate the probability space: conceptually, conditioning on $b_t$ induces a distribution $P(o \mid b_t, \tilde{I},Q)$ whose support is restricted to those objects lying inside $b_t$, and thus $P(o \mid b_t, \tilde{I},Q) \approx 0$ for all $o \notin b_t$. This anchors the latent focus to a localized subset of $\mathcal{O}$ instead of the entire scene.

Region-Aware Latent State ($\mathbf{h}_t$) denotes the \emph{Implicit Visual Focus}. It is not an external image crop, but a Soft Latent Aggregation of the MLLM's internal representation constrained by the spatial anchor, corresponding to the local realization of the object-centric features $\{z_o\}$ around $b_t$.
Formally, let $V \in \mathbb{R}^{H \times W \times C}$ be the global visual feature map encoded by the vision tower. We derive $\mathbf{h}_t$ via a spatial masking operation:
\begin{equation}
\small
    \mathbf{h}_t = \text{Pool}(V \odot \mathcal{M}(b_t)) .
\end{equation}
where $\mathcal{M}(b_t) \in \{0, 1\}^{H \times W}$ is the binary spatial mask generated from the coordinates $b_t$, and $\odot$ denotes the element-wise Hadamard product. This masking operation implements \emph{signal denoising}: by zeroing out features outside the critical region, it explicitly filters out the environmental background noise $\xi_{env}$ and ensures that the resulting state $\mathbf{h}_t$ aggregates only the signal energy relevant to the target.

\noindent \textbf{Grounded Description ($c_t$):} $c_t$ is the semantic decoding of $\mathbf{h}_t$, representing the conditional distribution $P(c_t \mid \mathbf{h}_t)$, which serves as the interpretable verification of the feature's semantic fidelity. In other words, $c_t$ provides a textual instantiation of the latent state $\mathbf{h}_t$ and explicitly ties the visual evidence back to the answer space, making the connection between $(o, z_o)$ and the final decision $A$ observable at the atom level.

\paragraph{Generative Process as Sequential Uncertainty Elimination.}
Within EGVOR, the inference process is modeled as a path integral over the space of evidence chains. The joint probability of generating a reasoning chain $\mathcal{A} = \{a_1, \dots, a_T\}$ is factorized auto-regressively. Each step represents a \textbf{Locate-Attend-Describe} cycle designed to progressively reduce entropy:
\begin{equation}
    \label{eq:generative_process}
    \footnotesize
    \begin{aligned}
    P(a_t \mid a_{<t}, \tilde{I},Q) = &\underbrace{P(c_t \mid \mathbf{h}_t, a_{<t})}_{\text{Semantic Verification}} \cdot \underbrace{P(\mathbf{h}_t \mid b_t, \tilde{I})}_{\text{Feature Denoising}} \\
    &\cdot \underbrace{P(b_t \mid a_{<t}, \tilde{I}, Q)}_{\text{Hypothesis Generation}}.
    \end{aligned}
\end{equation}
In our implementation, $P(\mathbf{h}_t \mid b_t, \tilde{I})$ is realized by the deterministic mapping $\mathbf{h}_t = \text{Pool}(V \odot \mathcal{M}(b_t))$, i.e., it is a degenerate distribution concentrated on the masked aggregation of the visual features. We keep the factorization in Eq.~\ref{eq:generative_process} to emphasize the distinct conceptual roles of \emph{where to look} ($b_t$), \emph{what is seen} ($\mathbf{h}_t$), and \emph{how to describe it} ($c_t$).

This factorization imposes a structural constraint on the reasoning flow:
1) \textbf{Hypothesis Generation:} The model first predicts a spatial hypothesis $b_t$ based on the query $Q$ and the previous atoms $a_{<t}$, selecting a candidate region that is encouraged (by training objectives in Sec.~\ref{subsec:training_framework}) to overlap with the Critical Set $\mathcal{O}_{crit}$.

2) \textbf{Feature Denoising:} Conditioned on $b_t$, the model deterministically extracts $\mathbf{h}_t$, effectively pruning the interference terms in Eq.~\ref{eq:full_expansion} originating from $\mathcal{O}_{irr}$ and $\mathcal{O}_{bg}$ and reducing the visual uncertainty around the current focus.

3) \textbf{Semantic Verification:} The feature $\mathbf{h}_t$ is decoded into text $c_t$, ensuring that the visual signal supports a coherent semantic claim. Maximizing $P(c_t \mid \mathbf{h}_t)$ aligns the latent state with a stable semantic manifold, which directly targets the semantic uncertainty term in Eq.~\ref{eq:variance_decomp}.

\noindent \textbf{1. Spatial Alignment via $b_t$ (Suppressing Term 2).} 
To mitigate attention drift, we treat the explicit box $b_t$ as a \emph{hard spatial constraint} $Q_{loc}(\cdot \mid b_t)$ over the object set $\mathcal{O}$. Intuitively, $Q_{loc}(o \mid b_t)$ puts most of its mass on objects whose support lies inside $b_t$ and near-zero mass elsewhere. The model, on the other hand, induces its own internal attention distribution $P_{attn}(\cdot)$ over candidate objects through its vision-language layers. Minimizing the divergence between $P_{attn}$ and this constrain encourages the model's “eye” to align with its “finger,” reducing the spatial variance $\text{Var}_{o}(\mathbb{E}[Z \mid o])$.

\noindent \textbf{2. Semantic Alignment via $c_t$ (Suppressing Term 1).} 
Spatial fixing alone is insufficient if the feature remains noisy. We therefore introduce the grounded description $c_t$ as a \emph{Semantic Control Variate}. By maximizing the likelihood $P(c_t \mid \mathbf{h}_t)$, we enforce the noisy feature $\mathbf{h}_t$ to collapse onto a distinct semantic manifold defined by $c_t$. 
Mathematically, the composite alignment objective is formulated as:
\begin{equation}
\small
\begin{aligned}
    \label{eq:dual_alignment}
    \mathcal{L}_{align} = \underbrace{D_{KL}\big(Q_{loc}(\cdot \mid b_t) \,\big\|\, P_{attn}(\cdot \mid \mathbf{h}_t)\big)}_{\text{Spatial Anchor (Term 2)}} \\ - \underbrace{\lambda \log P(c_t \mid \mathbf{h}_t)}_{\text{Semantic Anchor (Term 1)}}.
\end{aligned}
\end{equation}
This encourages the latent state $\mathbf{h}_t$ to be both spatially localized and semantically stable, thereby minimizing the intrinsic variance $\mathbb{E}_{o}[\text{Var}(Z \mid o)]$.

It is worth emphasizing that Eq.~\ref{eq:dual_alignment} describes a \emph{conceptual} alignment objective implied by our variance analysis. In practice, we do not optimize $\mathcal{L}_{align}$ explicitly. Instead, in the RL stage (Sec.~\ref{subsec:training_framework}), the spatial KL term is instantiated by the HRGW-based reward $\Psi_{spatial}$, which drives predicted boxes $b_t$ to concentrate on $\mathcal{O}_{crit}$ and away from $\mathcal{O}_{non}$, while the semantic log-likelihood term is realized via $\Xi_{align}$, which rewards agreement between $c_t$ and the ground-truth descriptions conditioned on $\mathbf{h}_t$. These two objectives act as tractable proxies to approximate the ideal spatial and semantic alignment encoded in $\mathcal{L}_{align}$.

\paragraph{Entropy Reduction via via Visual Probability Concentration.}
\textcolor{black}{We now connect the evidence atom $a_t=\langle b_t,h_t,c_t\rangle$ to the probabilistic bottlenecks identified in Sec.~\ref{subsec:preliminaries}. The spatial anchor $b_t$ acts as a support-truncation operator over the latent focus space. Let $\mathcal{O}(b_t)\subseteq\mathcal{O}$ denote the subset of objects whose spatial support lies inside the region defined by $b_t$. Conditioning on $b_t$ induces the truncated focus distribution
\begin{equation}
\label{eq:truncated_focus}
\small
P(o|b_t,\tilde{I},Q)
=
\begin{cases}
\frac{
P(o|\tilde{I},Q)
}{
\sum_{o'\in\mathcal{O}(b_t)}
P(o'|\tilde{I},Q)
},
& o\in\mathcal{O}(b_t),
\\
0,
& o\notin\mathcal{O}(b_t).
\end{cases}
\end{equation}
This equation should be interpreted as support truncation rather than a universal entropy theorem. Conditioning on an arbitrary box does not necessarily reduce the actual entropy for every possible distribution. However, when $b_t$ is a valid and compact evidence anchor, it restricts the support of $O_f$ from the full scene $\mathcal{O}$ to the local subset $\mathcal{O}(b_t)$.}

\textcolor{black}{Therefore, the focus entropy under the anchor satisfies the support-based upper bound
\begin{equation}
\label{eq:support_entropy_bound}
\small
H(O_f|b_t,\tilde{I},Q)
\le
\log|\mathcal{O}(b_t)|.
\end{equation}
Since the evidence atom $a_t$ contains the anchor $b_t$, its focus support is also contained in $\mathcal{O}(b_t)$:
\begin{equation}
\label{eq:atom_entropy_bound}
\small
H(O_f|a_t,\tilde{I},Q)
\le
\log|\mathcal{O}(b_t)|.
\end{equation}
For the unanchored case, the worst-case bound is
\begin{equation}
\small
H(O_f|\tilde{I},Q)
\le
\log|\mathcal{O}|.
\end{equation}
When $b_t$ is compact and query-critical, $|\mathcal{O}(b_t)|\ll|\mathcal{O}|$, yielding the upper-bound comparison
\begin{equation}
\small
\log|\mathcal{O}(b_t)|
\ll
\log|\mathcal{O}|.
\end{equation}
Thus, CoE reduces the worst-case support size of the latent visual focus. In the ideal case where $b_t$ tightly covers a single critical object, $P(O_f=o_{crit}|b_t,\tilde{I},Q)\approx 1$, and the anchored focus entropy approaches zero.}

\textcolor{black}{The grounded description $c_t$ further provides semantic verification. It checks whether the localized feature $h_t$ corresponds to a coherent visual concept relevant to the query. Therefore, $b_t$ reduces where-to-look uncertainty, while $c_t$ reduces what-is-seen ambiguity.}

\paragraph{Dual-Variance Suppression.}
\textcolor{black}{CoE also modifies the variance decomposition in Eq.~\ref{eq:variance_decomp} through two complementary constraints. First, the spatial anchor $b_t$ replaces the global focus distribution $P(\cdot|\tilde{I},Q)$ with a localized conditional distribution:
\begin{equation}
\small
O_f\sim P(\cdot|b_t,\tilde{I},Q).
\end{equation}
The corresponding spatial variance term becomes
\begin{equation}
\label{eq:local_spatial_variance}
\small
\operatorname{Var}_{O_f\sim P(\cdot|b_t,\tilde{I},Q)}
\left(
\mathbb{E}[Z|O_f,\tilde{I},Q]
\right).
\end{equation}
Because the support of $P(\cdot|b_t,\tilde{I},Q)$ is contained in $\mathcal{O}(b_t)$, a compact and query-critical anchor reduces the worst-case support over which the spatial variance is computed. This directly targets the Spatial Boundary Ambiguity term in Eq.~\ref{eq:variance_decomp}.}

\textcolor{black}{Second, the grounded description $c_t$ regularizes the noisy local feature $h_t$ toward a stable semantic interpretation. Maximizing $P(c_t|h_t)$ does not eliminate all feature noise, but it encourages the localized representation to align with a coherent semantic concept. This targets the Intrinsic Semantic Uncertainty term:
\begin{equation}
\small
\mathbb{E}_{O_f\sim P(\cdot|\tilde{I},Q)}
\left[
\operatorname{Var}(Z|O_f,\tilde{I},Q)
\right].
\end{equation}
Together, $b_t$ and $c_t$ address the two bottlenecks identified in Sec.~\ref{subsec:preliminaries}: $b_t$ suppresses spatial boundary ambiguity by localizing the focus distribution, while $c_t$ suppresses intrinsic semantic uncertainty by verifying the semantic content of the localized feature.}

\subsection{The EGVOR Training Framework: A Hierarchical Curriculum Learning}
\label{subsec:training_framework}

\begin{figure*}[h]
    \centering
    \includegraphics[width=1\linewidth]{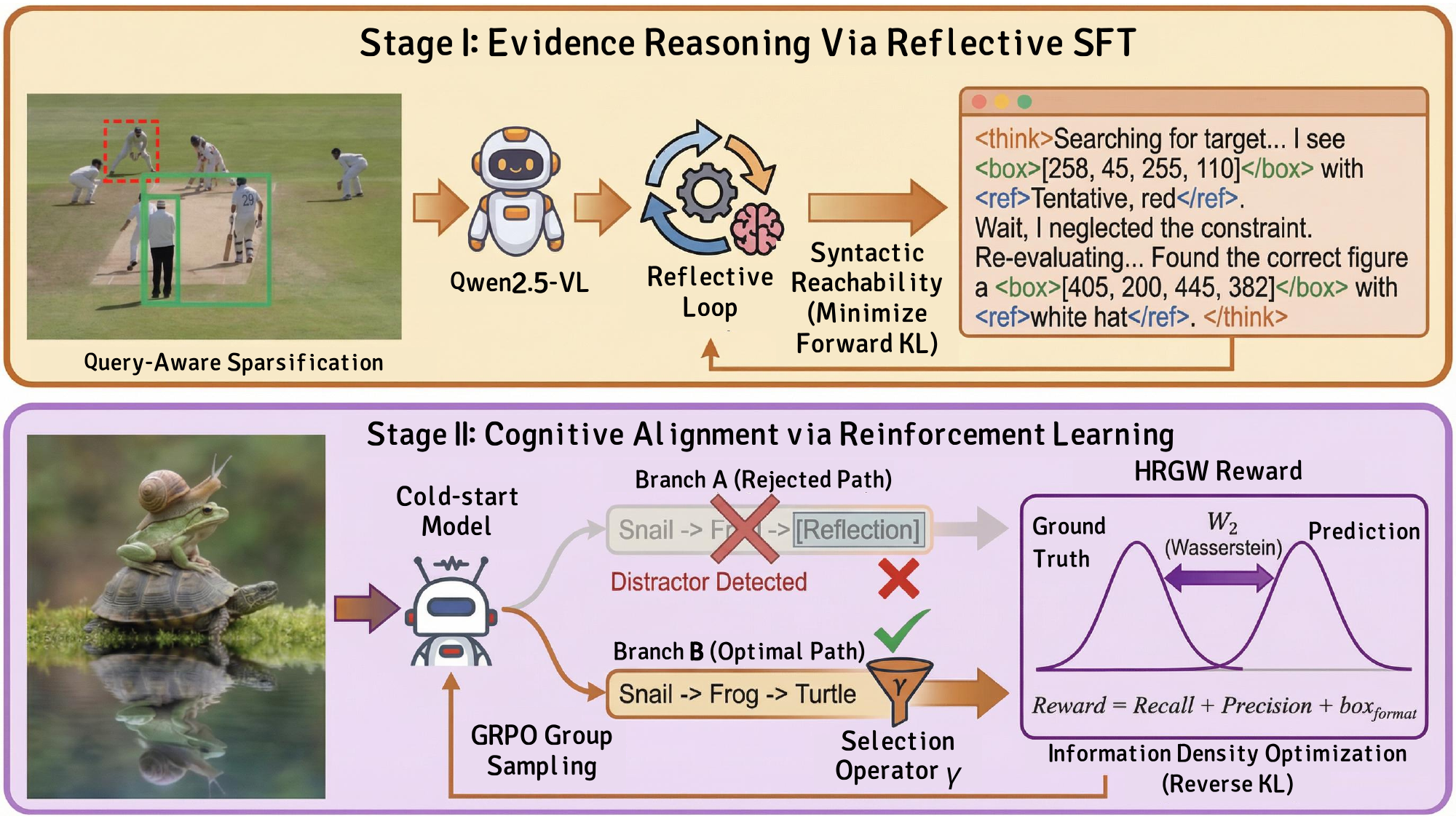} %
    \vspace{-10pt}
    \caption{The two-stage training pipeline of EGVOR. (a) The bootstrapping stage uses supervised instruction tuning to teach the model the syntax of grounded reasoning. (b) The policy refinement stage employs reinforcement learning with our novel multi-faceted reward function and spatial-semantic alignment.}
    \label{fig:pipeline}
    \vspace{-5pt}
\end{figure*}

We propose a \textbf{Hierarchical Curriculum Learning} framework to train the EGVOR model. Direct optimization of the full evidence chain is intractable due to the non-differentiable explicit spatial anchors ($b_t$) and the unobservable implicit latent states ($\mathbf{h}_t$). Therefore, we decompose training into a two-stage curriculum: first establishing the \emph{Evidence Reasoning Chain} via Supervised Fine-Tuning (SFT), and then performing \emph{Cognitive Alignment} via Reinforcement Learning (RL).

\subsubsection{Stage I: Establishing Evidence Reasoning via Reflective SFT}
\label{stage1}

The primary objective of the first stage is to initialize the policy $\pi_\theta$ to adhere to the syntactic structure of the atomic paradigm. We utilize the \textbf{SFT-Instruct} dataset to teach the model the \emph{Locate-Attend-Describe} format.

\paragraph{Evidence Chain Establishment (Syntactic Prior).}
We train the model to explicitly represent its reasoning process by maximizing the likelihood of the ground-truth evidence chain $\tau$. Let the trajectory $\tau$ be a sequence of $L$ tokens, denoted as $\tau = (\tau_1, \tau_2, \dots, \tau_L)$. The generation follows the template: $\tau = \texttt{<think>} \dots \texttt{</think>} \rightarrow \texttt{<box>} [x_1, y_1, x_2, y_2] \texttt{</box>} \rightarrow \texttt{<ref>} c_t \texttt{</ref>}$. The optimization objective is the standard autoregressive Negative Log-Likelihood (NLL) over the token sequence:
\begin{equation}
\small
    \mathcal{L}_{SFT} = -\mathbb{E}_{(I, Q, \tau) \sim \mathcal{D}_{SFT}} \sum_{l=1}^{L} \log \pi_\theta(\tau_l \mid \tau_{<l}, I, Q) .
\end{equation}
This phase establishes the foundational association between visual perception and explicit spatial output, serving as a warm-up for the policy to learn the valid output space of the evidence atoms.

\paragraph{Cognitive Recovery via Error-Induced Self-Correction.}
Standard SFT often suffers from \emph{Exposure Bias}, where the model lacks the ability to recover from cumulative errors during inference. To address the \emph{Attention Drift} discussed in Section~\ref{subsec:preliminaries}, we introduce a \emph{Reflective Decoding} mechanism. We augment the training data by injecting procedural perturbations: for a valid reasoning step, we substitute the correct bounding box with a distractor box $b_{err}$ in a subset of samples, and supervise the model to generate a correction trajectory $\tau_{corr}$ that explicitly rejects the error:
\begin{equation}
\small
\begin{aligned}
    \tau_{corr} = b_{err} \rightarrow \texttt{<think>}
    \text{Wait, the selected region is} \\
    \text{background...Refine focus.} 
    \texttt{</think>} \rightarrow b_{gt} \rightarrow c_{gt} .
\end{aligned}
\end{equation}
This mechanism teaches the model the conditional probability $P(\text{Correction} \mid b_{err}, \mathbf{h})$, equipping it with the meta-cognitive capability to detect spatial hallucinations and realign its focus.

\subsubsection{Stage II: Cognitive Alignment via Reinforcement Learning}

While SFT establishes the format, it does not guarantee the stability of the variance terms identified in Eq.~\ref{eq:variance_decomp}. In the second stage, we utilize the RL-Instruct dataset to numerically solve the \emph{Dual-Variance Suppression} problem. We formulate three functional objectives optimized via Group Relative Policy Optimization (GRPO).

\paragraph{Objective 1: Minimizing Spatial Ambiguity ($\Psi_{spatial}$).}
\emph{(Targeting Term 2: $\text{Var}_o(\mathbb{E}[Z \mid o])$)}

\textcolor{black}{Adverse weather blurs object boundaries, creating high spatial aleatoric uncertainty. Standard IoU rewards fail due to sparsity: they yield zero feedback for non-overlapping predictions. To address this, we propose the Hierarchical Robust Gaussian-Wasserstein (HRGW) objective. We model bounding boxes as 2D Gaussian distributions $\mathcal{N}(\mu, \Sigma)$ and quantify affinity via the Wasserstein-2 distance ($W_2$), wrapped in a \emph{Normalized Wasserstein Distance}:
\begin{equation}
\small
    \mathcal{A}(b_{pred}, b_{gt}) = \exp \left( - \frac{\sqrt{W_{2}^{2}(\mathcal{N}_{pred}, \mathcal{N}_{gt})}}{C} \right) .
\end{equation}
where, $C$ is a normalization constant empirically set to align with the dataset scale. This provides dense reward shaping even for disjoint predictions to guide the policy towards the Critical Set $\mathcal{O}_{crit}$.}

\textcolor{black}{To implement spatial pruning of distractors, we define the spatial gain $\Psi_{spatial}$. Let $\tau_b \subset \tau$ denote the set of all spatial anchors (bounding boxes) parsed from the generated trajectory. For each annotated object $o \in \mathcal{O}$, let $b^{gt}_o$ denote its ground-truth bounding box. We assign an importance weight $\omega_o$ where $\text{sgn}(\omega_o)=+1$ for targets ($o \in \mathcal{O}_{crit}$) and $\text{sgn}(\omega_o)=-1$ for non-critical objects ($o \in \mathcal{O}_{non}$). The reward is:
\begin{equation}
\small
\begin{aligned}
    \Psi_{spatial}(\tau) = \alpha_{rec} \cdot \frac{\sum_{o \in \mathcal{O}_{crit}} \max_{b \in \tau_b} \mathcal{A}(b, b^{gt}_o)}{|\mathcal{O}_{crit}|} \\ + \alpha_{prec} \cdot \frac{1}{|\tau_b|} \sum_{b \in \tau_b} \max_{o \in \mathcal{O}} \big(\text{sgn}(\omega_o) \cdot \mathcal{A}(b, b^{gt}_o)\big) .
\end{aligned}
\end{equation}}

\textcolor{black}{The first term improves recall by encouraging coverage of critical objects. The second term is precision-oriented: each predicted box is matched to its closest annotated object; matches to $\mathcal{O}_{crit}$ contribute positively, while matches to $\mathcal{O}_{non}$ contribute negatively, explicitly discouraging attention drift to distractors/background.}

\textcolor{black}{Additionally, we incorporate a box-format validity reward $\mathcal{R}_{box}(\tau)$ to encourage syntactically and geometrically well-formed predictions. Concretely, $\mathcal{R}_{box}(\tau)=1$ if all \texttt{<box>}...\texttt{</box>} spans in $\tau$ are present and each box is in the correct coordinate order (e.g., $x_1 < x_2$, $y_1 < y_2$); otherwise $\mathcal{R}_{box}(\tau)=0$.}

\paragraph{Objective 2: Enforcing Semantic Stability ($\Xi_{align}$).}
\emph{(Targeting Term 1: $\mathbb{E}_o[\text{Var}(Z \mid o)]$)}

Even with correct localization, feature attenuation persists. As established in Eq.~\ref{eq:dual_alignment}, robust reasoning requires minimizing the divergence between the latent state $\mathbf{h}_t$ and the semantic description $c_t$. We instantiate the log-likelihood constraint $\log P(c_t \mid \mathbf{h}_t)$ using a semantic similarity metric by GPT-5.2 (SimScore) as a reward proxy:
\begin{equation}
\small
    \Xi_{align}(\tau) = \frac{1}{T} \sum_{t=1}^T \text{SimScore}(c_t, c_{gt}^{matched} \mid b_t) .
\end{equation}
Maximizing $\Xi_{align}$ effectively forces the noisy latent features $\mathbf{h}_t$ to collapse onto the deterministic semantic manifold defined by the ground truth attributes, thereby suppressing the intrinsic semantic uncertainty.

\paragraph{Objective 3: Maximizing Cognitive Path Diversity ($\Lambda_{path}$).}
To approximate the path integral $\sum P(A \mid E)$, the policy should avoid collapsing to a single mode. We encourage diverse reasoning strategies by matching the generated chain against a set of $M$ valid trajectories (e.g., different reasoning orders):
\begin{equation}
\small
    \Lambda_{path}(\tau) = \max_{m \in \{1..M\}} \text{Sim}_{logic}(\tau_{think}, \tau_{gt}^{(m)}) + \mathcal{R}_{fmt} .
\end{equation}
where $\mathcal{R}_{fmt}$ is a structural term enforcing the validity of the reasoning format (e.g., correct usage and ordering of \texttt{<think>} tags).

\paragraph{Summary: The GRPO Optimization.}
We integrate these objectives into a unified Group Relative Policy Optimization framework. For a given query $Q$, we sample a group of trajectories $G = \{\tau^{(1)}, \dots, \tau^{(G)}\}$ and compute the total advantage $J_{total}$ for each trajectory:
\begin{equation}
\small
\begin{aligned}
    J_{total}(\tau^{(i)}) = \Psi_{spatial}(\tau^{(i)}) + \Xi_{align}(\tau^{(i)}) \\+ \Lambda_{path}(\tau^{(i)}) + \mathbb{I}(A_{pred} = A_{gt}) + \mathcal{R}_{box}(\tau^{(i)}).
\end{aligned}
\end{equation}
The policy parameters $\theta$ are updated to maximize the expected return relative to the group-wise weighting, which stabilizes training in high-variance environments:
\begin{equation}
\small
\begin{aligned}
    \nabla_\theta \mathcal{L}_{GRPO} = \mathbb{E}_{Q \sim \mathcal{D}} \\ \Bigg[
    \frac{1}{G} \sum_{i=1}^G \frac{\exp(J_{total}(\tau^{(i)}) / \beta)} {\sum_{j} \exp(J_{total}(\tau^{(j)}) / \beta)} \nabla_\theta \log \pi_\theta(\tau^{(i)} \mid Q) \Bigg] .
\end{aligned}
\end{equation}
where $\beta$ is a temperature hyper-parameter controlling the sharpness of the group-wise weighting. This formulation translates the probabilistic constraints of high-entropy reasoning into a differentiable optimization process, and empirically aligns the model’s perception and reasoning with the intended cognitive patterns.

\section{Experiments}
\subsection{Benchmark Evaluation Setup and Implementation Details}
\label{subsec:exp_setup_bench}

\begin{table*}[htbp]
\centering
\setlength{\tabcolsep}{2.5pt}
\renewcommand{\arraystretch}{1.15} 
\resizebox{1\textwidth}{!}{%
\begin{tabular}{l|c|c|cccc|ccccc|cccc|cc}
\toprule

\multirow{2}{*}{\textbf{Model}} & 
\multirow{2}{*}{\textbf{Ave.}} & 
\multirow{2}{*}{\textbf{LLM}} & 
\cellcolor{color_obs}\rotatebox{60}{\textit{Existence}} & 
\cellcolor{color_obs}\rotatebox{60}{\textit{Counting}} & 
\cellcolor{color_obs}\rotatebox{60}{\textit{Location}} & 
\cellcolor{color_obs}\rotatebox{60}{\textit{Detection}} & 
\cellcolor{color_sym}\rotatebox{60}{\textit{OCR-F1}} & 
\cellcolor{color_sym}\rotatebox{60}{\textit{OCR-CER}} & 
\cellcolor{color_sym}\rotatebox{60}{\textit{Attribute}} & 
\cellcolor{color_sym}\rotatebox{60}{\textit{Position}} & 
\cellcolor{color_sym}\rotatebox{60}{\textit{Cap/Other}} & 
\cellcolor{color_rel}\rotatebox{60}{\textit{Spatial}} & 
\cellcolor{color_rel}\rotatebox{60}{\textit{Occlusion}} & 
\cellcolor{color_rel}\rotatebox{60}{\textit{Social}} & 
\cellcolor{color_rel}\rotatebox{60}{\textit{Causal}} & 
\cellcolor{color_plan}\rotatebox{60}{\textit{w/o CoE}} & 
\cellcolor{color_plan}\rotatebox{60}{\textit{w/ CoE}} \\

& & & 
\multicolumn{4}{c|}{\cellcolor{color_obs}\textbf{Basic Perception}} & 
\multicolumn{5}{c|}{\cellcolor{color_sym}\textbf{Advanced Perception}} & 
\multicolumn{4}{c|}{\cellcolor{color_rel}\textbf{Relation Understanding}} & 
\multicolumn{2}{c}{\cellcolor{color_plan}\textbf{Event Reasoning}} \\
\midrule

\multicolumn{18}{l}{\textit{\textbf{Proprietary Models}}} \\
\hline

\multirow{2}{*}{\textcolor{black}{Gemini-3.5-Flash-0519}} & \multirow{2}{*}{61.4} & \multirow{2}{*}{Gemini} 
& 68.24 & 53.0 & 38.0 & 7.28 & 
84.80 & 39.89 & 63.03 & 54.25 & 84.0 & 
66.0 & 58.0 & 71.0 & 58.0 & 
66.8 & 69.6 \\
& & & \multicolumn{4}{c|}{41.6} & \multicolumn{5}{c|}{71.5} & \multicolumn{4}{c|}{63.3} & \multicolumn{2}{c}{69.0} \\
\hline

\multirow{2}{*}{Gemini-2.5-Flash-0520} & \multirow{2}{*}{55.5} & \multirow{2}{*}{Gemini} 
& 67.1 & 43.3 & 43.3 & 6.9 & 94.1 & 10.3 & 57.7 & 46.8 & 75.9 & 55.0 & 47.9 & 66.8 & 43.8 & 53.1 & 58.9 \\
& & & \multicolumn{4}{c|}{40.2} & \multicolumn{5}{c|}{68.6} & \multicolumn{4}{c|}{53.4} & \multicolumn{2}{c}{58.9} \\ 
\hline

\multirow{2}{*}{\textcolor{black}{GPT-4o-2024-1120}} & \multirow{2}{*}{53.1} & \multirow{2}{*}{GPT} 
& 58.48 & 53.2 & 48.84 & 10.27 & 
84.17 & 30.55 & 32.66 & 29.41 & 77.0 & 
55.5 & 46.5 & 60.5 & 55.0 & 
56.8 & 59.6 \\
& & & \multicolumn{4}{c|}{42.7} & \multicolumn{5}{c|}{55.8} & \multicolumn{4}{c|}{54.4} & \multicolumn{2}{c}{59.6} \\
\hline

\multirow{2}{*}{GPT-4.1-mini-0414} & \multirow{2}{*}{52.0} & \multirow{2}{*}{GPT} 
& 63.8 & 46.8 & 46.2 & 2.4 & 91.6 & 11.2 & 57.4 & 36.3 & 82.2 & 47.0 & 39.9 & 51.1 & 53.4 & 51.2 & 55.3 \\
& & & \multicolumn{4}{c|}{39.8} & \multicolumn{5}{c|}{66.9} & \multicolumn{4}{c|}{47.9} & \multicolumn{2}{c}{55.3} \\
\hline
\multirow{2}{*}{Claude-3.5-Sonnet-0620} & \multirow{2}{*}{47.8} & \multirow{2}{*}{Claude} 
& 60.3 & 45.2 & 32.7 & 4.8 & 71.7 & 40.5 & 52.6 & 42.9 & 65.8 & 50.0 & 43.8 & 63.6 & 34.8 & 45.7 & 51.4 \\
& & & \multicolumn{4}{c|}{35.8} & \multicolumn{5}{c|}{58.2} & \multicolumn{4}{c|}{48.0} & \multicolumn{2}{c}{51.4} \\
\hline

\multicolumn{18}{l}{\textit{\textbf{Open Source Models}}} \\
\hline

\multirow{2}{*}{LLaVA-1.5-7B} & \multirow{2}{*}{40.9} & \multirow{2}{*}{Vicuna} & 
50.8 & 50.5 & 19.1 & 0.0 & 
24.1 & 173.1 & 37.2 & 36.1 & 62.2 & 
50.1 & 51.0 & 67.4 & 50.4 & 
32.0 & 38.9 \\
& & & \multicolumn{4}{c|}{30.1} & \multicolumn{5}{c|}{39.9} & \multicolumn{4}{c|}{54.7} & \multicolumn{2}{c}{38.9} \\
\hline

\multirow{2}{*}{LLaVA-NeXT-7B} & \multirow{2}{*}{41.0} & \multirow{2}{*}{Vicuna} & 
52.8 & 43.2 & 19.1 & 0.0 & 
32.8 & 125.5 & 37.5 & 26.0 & 67.3 & 
44.5 & 50.9 & 63.2 & 45.8 & 
34.2 & 43.2 \\
& & & \multicolumn{4}{c|}{28.8} & \multicolumn{5}{c|}{40.9} & \multicolumn{4}{c|}{51.1} & \multicolumn{2}{c}{43.2} \\
\hline

\multirow{2}{*}{LLaVA-OneVision-7B} & \multirow{2}{*}{44.8} & \multirow{2}{*}{Qwen2} & 
43.7 & 22.9 & 20.7 & 0.1 & 
56.5 & 65.7 & 49.7 & 29.1 & 71.9 & 
50.5 & 42.6 & 73.7 & 53.4 & 
43.8 & 50.5 \\
& & & \multicolumn{4}{c|}{21.9} & \multicolumn{5}{c|}{51.8} & \multicolumn{4}{c|}{55.1} & \multicolumn{2}{c}{50.5} \\
\hline

\multirow{2}{*}{MiniCPM-V-2.5-8B} & \multirow{2}{*}{43.9} & \multirow{2}{*}{Llama3} & 
59.5 & 53.6 & 24.5 & 5.7 & 
67.3 & 41.0 & 50.3 & 22.5 & 71.6 & 
49.5 & 37.5 & 22.3 & 47.9 & 
40.3 & 47.5 \\
& & & \multicolumn{4}{c|}{35.8} & \multicolumn{5}{c|}{52.9} & \multicolumn{4}{c|}{39.3} & \multicolumn{2}{c}{47.5} \\
\hline

\multirow{2}{*}{MiniCPM-V-2.6-8B} & \multirow{2}{*}{49.8} & \multirow{2}{*}{Qwen2} & 
55.5 & 56.2 & 42.9 & 7.7 & 
78.5 & 27.5 & 53.6 & 40.4 & 74.2 & 
43.3 & 41.0 & 54.7 & 51.9 & 
42.1 & 49.3 \\
& & & \multicolumn{4}{c|}{40.6} & \multicolumn{5}{c|}{61.7} & \multicolumn{4}{c|}{47.7} & \multicolumn{2}{c}{49.3} \\
\hline

\multirow{2}{*}{MiniCPM-o-2.6-8B} & \multirow{2}{*}{53.1} & \multirow{2}{*}{Qwen2.5} & 
63.1 & 65.6 & 27.7 & 5.6 & 
84.2 & 27.3 & 56.8 & 51.9 & 78.9 & 
55.6 & 39.3 & 62.1 & 54.2 & 
44.0 & 51.2 \\
& & & \multicolumn{4}{c|}{40.5} & \multicolumn{5}{c|}{67.9} & \multicolumn{4}{c|}{52.8} & \multicolumn{2}{c}{51.2} \\
\hline

\multirow{2}{*}{Janus-Pro-7B} & \multirow{2}{*}{45.9} & \multirow{2}{*}{DeepSeek} & 
56.6 & 58.6 & 21.9 & 0.1 & 
22.0 & 143.6 & 54.0 & 46.0 & 75.2 & 
49.5 & 40.2 & 63.2 & 51.9 & 
39.3 & 48.8 \\
& & & \multicolumn{4}{c|}{34.3} & \multicolumn{5}{c|}{49.3} & \multicolumn{4}{c|}{51.2} & \multicolumn{2}{c}{48.8} \\
\hline

\multirow{2}{*}{Qwen2-VL-7B} & \multirow{2}{*}{47.9} & \multirow{2}{*}{Qwen2} & 
58.1 & 63.3 & 37.2 & 4.5 & 
59.0 & 49.1 & 49.3 & 30.6 & 72.7 & 
48.2 & 45.3 & 71.6 & 41.2 & 
39.2 & 46.5 \\
& & & \multicolumn{4}{c|}{40.8} & \multicolumn{5}{c|}{52.9} & \multicolumn{4}{c|}{51.6} & \multicolumn{2}{c}{46.5} \\
\hline

\multirow{2}{*}{Qwen2.5-VL-7B} & \multirow{2}{*}{55.0} & \multirow{2}{*}{Qwen2.5} & 
58.7 & 59.4 & 68.1 & 51.0 & 
86.5 & 15.4 & 54.2 & 33.7 & 75.7 & 
46.1 & 42.2 & 52.6 & 42.8 & 
41.7 & 52.3 \\
& & & \multicolumn{4}{c|}{59.3} & \multicolumn{5}{c|}{62.5} & \multicolumn{4}{c|}{45.9} & \multicolumn{2}{c}{52.3} \\
\hline

\multirow{2}{*}{Qwen2.5-VL-32B} & \multirow{2}{*}{59.3} & \multirow{2}{*}{Qwen2.5} & 
54.9 & 59.4 & 79.4 & 52.7 & 
87.7 & 13.5 & 64.1 & 44.0 & 77.5 & 
54.2 & 49.5 & 60.4 & 42.8 & 
47.2 & 55.7 \\
& & & \multicolumn{4}{c|}{61.6} & \multicolumn{5}{c|}{68.3} & \multicolumn{4}{c|}{51.7} & \multicolumn{2}{c}{55.7} \\
\hline

\multirow{2}{*}{Qwen2.5-VL-72B} & \multirow{2}{*}{60.7} & \multirow{2}{*}{Qwen2.5} & 
63.5 & 60.2 & 76.8 & 54.5 & 
89.5 & 12.6 & 60.8 & 39.1 & 79.3 & 
53.8 & 56.0 & 64.5 & 44.0 & 
50.8 & 57.2 \\
& & & \multicolumn{4}{c|}{63.7} & \multicolumn{5}{c|}{67.2} & \multicolumn{4}{c|}{54.6} & \multicolumn{2}{c}{57.2} \\
\hline

\multirow{2}{*}{InternVL2-8B} & \multirow{2}{*}{50.2} & \multirow{2}{*}{InternLM} & 
61.2 & 55.5 & 38.0 & 6.2 & 
83.3 & 35.3 & 59.1 & 38.2 & 67.5 & 
45.9 & 41.9 & 53.7 & 46.5 & 
43.8 & 51.7 \\
& & & \multicolumn{4}{c|}{40.2} & \multicolumn{5}{c|}{62.0} & \multicolumn{4}{c|}{47.0} & \multicolumn{2}{c}{51.7} \\
\hline
\multirow{2}{*}{InternVL2.5-8B} & \multirow{2}{*}{52.8} & \multirow{2}{*}{InternLM} & 
62.1 & 54.0 & 25.1 & 5.5 & 
89.7 & 15.0 & 59.9 & 45.2 & 78.4 & 
53.7 & 43.2 & 65.3 & 52.7 & 
42.9 & 52.6 \\
& & & \multicolumn{4}{c|}{36.7} & \multicolumn{5}{c|}{68.3} & \multicolumn{4}{c|}{53.7} & \multicolumn{2}{c}{52.6} \\
\hline
\multirow{2}{*}{InternVL3-8B} & \multirow{2}{*}{54.8} & \multirow{2}{*}{Qwen2.5} & 
59.6 & 57.1 & 41.4 & 6.3 & 
89.7 & 15.1 & 57.9 & 56.4 & 78.1 & 
58.8 & 45.6 & 60.0 & 54.2 & 
42.2 & 53.0 \\ 
& & & \multicolumn{4}{c|}{41.1} & \multicolumn{5}{c|}{70.5} & \multicolumn{4}{c|}{54.7} & \multicolumn{2}{c}{53.0} \\
\hline

\multirow{2}{*}{InternVL3-38B} & \multirow{2}{*}{57.4} & \multirow{2}{*}{Qwen2.5} & 
61.8 & 65.7 & 29.2 & 6.1 & 
87.3 & 16.8 & 67.7 & 53.6 & 80.4 & 
65.3 & 53.5 & 65.8 & 54.2 & 
48.5 & 56.8 \\
& & & \multicolumn{4}{c|}{40.7} & \multicolumn{5}{c|}{72.3} & \multicolumn{4}{c|}{59.7} & \multicolumn{2}{c}{56.8} \\
\hline

\multirow{2}{*}{InternVL3-78B} & \multirow{2}{*}{60.2} & \multirow{2}{*}{Qwen2.5} & 
67.3 & 65.1 & 43.1 & 6.2 & 
91.7 & 11.9 & 69.4 & 52.8 & 82.6 & 
64.5 & 54.6 & 68.1 & 55.7 & 
52.7 & 60.5 \\
& & & \multicolumn{4}{c|}{45.4} & \multicolumn{5}{c|}{74.1} & \multicolumn{4}{c|}{60.7} & \multicolumn{2}{c}{60.5} \\
\hline

\multirow{2}{*}{Deepeyes-7B} & \multirow{2}{*}{63.8} & \multirow{2}{*}{Qwen2.5} & 
65.2 & 62.1 & 82.4 & 64.6 & 
87.4 & 13.6 & 66.1 & 55.4 & 76.9 & 
54.6 & 53.7 & 58.5 & 45.1 & 
60.9 & 62.1 \\
& & & \multicolumn{4}{c|}{68.6} & \multicolumn{5}{c|}{71.5} & \multicolumn{4}{c|}{53.0} & \multicolumn{2}{c}{62.1} \\
\hline

\multirow{2}{*}{Pixel-Reasoner-7B} & \multirow{2}{*}{61.1} & \multirow{2}{*}{Qwen2.5} & 
60.5 & 61.3 & 79.6 & 57.3 & 
86.1 & 15.8 & 65.3 & 54.2 & 76.1 & 
49.1 & 50.5 & 56.2 & 43.9 & 
57.4 & 59.3 \\
& & & \multicolumn{4}{c|}{64.7} & \multicolumn{5}{c|}{70.4} & \multicolumn{4}{c|}{49.9} & \multicolumn{2}{c}{59.3} \\
\hline

\multirow{2}{*}{\textbf{EGVOR(ours)}} & \multirow{2}{*}{\textbf{67.7}} & \multirow{2}{*}{\textbf{Qwen2.5}} & 
\textbf{68.4} & \textbf{66.2} & \textbf{85.9} & \textbf{69.7} & 
\textbf{88.2} & \textbf{13.0} & \textbf{68.9} & \textbf{57.7} & \textbf{79.3} & 
\textbf{60.4} & \textbf{56.6} & \textbf{63.3} & \textbf{48.3} & 
\textbf{65.7} & \textbf{67.4} \\
& & & \multicolumn{4}{c|}{\textbf{72.6}} & \multicolumn{5}{c|}{\textbf{73.5}} & \multicolumn{4}{c|}{\textbf{57.1}} & \multicolumn{2}{c}{\textbf{67.4}} \\
\hline
\hline
\multirow{2}{*}{$\Delta$ \textit{v.s.} Qwen2.5-VL-7B} & \multirow{2}{*}{\up{12.7}} & \multirow{2}{*}{Qwen2.5} & 
\up{9.7} & \up{6.8} & \up{17.8} & \up{18.8} & 
\up{1.7} & \down{2.4} & \up{14.7} & \up{24.0} & \up{3.7} & 
\up{14.4} & \up{14.4} & \up{10.6} & \up{5.6} & 
\up{24.0} & \up{15.1} \\ 
& & & \multicolumn{4}{c|}{\up{13.3}} & \multicolumn{5}{c|}{\up{11.0}} & \multicolumn{4}{c|}{\up{11.2}} & \multicolumn{2}{c}{\up{15.1}} \\

\bottomrule
\end{tabular}
}
\vspace{3pt}
\caption{\textcolor{black}{Comprehensive Benchmark Results. Grouped by model series. $\Delta$ indicates improvement over the strong baseline Qwen2.5-VL-7B (green indicates improvement; for \textit{OCR-CER}, lower $\downarrow$ is better). Our performance is highlighted in \textbf{bold}.}}
\label{tab:main_benchmark}
\vspace{-10pt}
\end{table*}

\begin{table*}[t]
    \centering\small
    \begin{tabular}{l ccc ccc ccc}
    \toprule
    & \multicolumn{3}{c}{V* Bench} & \multicolumn{3}{c}{HR-Bench-4K} & \multicolumn{3}{c}{HR-Bench-8K} \\
    \cmidrule(lr){2-4}
    \cmidrule(lr){5-7}
    \cmidrule(lr){8-10}
    & \textbf{Overall} & Attr. & Spatial & \textbf{Overall} & Single & Cross & \textbf{Overall} & Single & Cross \\ 
    \midrule
    \multicolumn{10}{c}{\textbf{Private Models}} \\
    \midrule
    GPT-4o-2024-1120 & 66.0 & -- & -- & -- & -- & -- & -- & -- & -- \\
    o3-0416 & 95.7 & -- & -- & -- & -- & -- & -- & -- & -- \\
    \midrule
    \multicolumn{10}{c}{\textbf{Open-source General Models}} \\
    \midrule
    LLaVA-OneVision-7B & 70.7 & 73.0 & 60.5 & 64.3 & 74.8 & 53.8 & 59.8 & 65.3 & 54.3 \\
    LLaVA-OneVision-72B & 73.8 & 80.9 & 63.2 & 66.3 & 76.5 & 56.0 & 60.9 & 68.8 & 53.0 \\
    InternVL3-8B & 72.3 & 73.0 & 71.1 & 70.8 & 79.3 & 62.3 & 62.0 & 64.3 & 59.8 \\
    InternVL3-38B & 77.5 & 77.4 & 77.6 & 76.3 & 83.5 & 69.0 & 67.0 & 71.3 & 62.8 \\
    InternVL3-78B & 76.4 & 75.7 & 77.6 & 75.5 & 84.5 & 66.5 & 67.3 & 71.8 & 62.8 \\
    Qwen2.5-VL-7B & 74.3 & 77.4 & 69.7 & 72.1 & 88.8 & 55.5 & 68.8 & 83.5 & 54.0 \\
    Qwen2.5-VL-32B & 85.9 & 83.5 & 89.5 & 74.8 & 89.3 & 60.3 & 71.6 & 86.5 & 56.8 \\
    Qwen2.5-VL-72B & 84.8 & 90.8 & 80.9 & \textbf{79.4} & 88.8 & \textbf{70.0} & \textbf{76.3} & 84.3 & \textbf{68.3} \\
    \midrule
    \multicolumn{10}{c}{\textbf{Open-source Visual Grounded Reasoning Models}} \\
    \midrule
    Pixel-Reasoner-7B & 80.6 & 83.5 & 76.3 & 72.9 & 86.0 & 60.3 & 66.9 & 80.0 & 54.3 \\
    DeepEyes-7B & 90.0 & 92.1 & 86.8 & 75.1 & 91.3 & 59.0 & 72.6 & 86.8 & 58.5 \\
    \midrule
    \textbf{EGVOR (Ours)} & \textbf{91.9} & \textbf{95.4} & \textbf{88.3} & 78.6 & \textbf{91.8} & 65.4 & 75.0 & \textbf{89.5} & 60.4 \\
    $\Delta$ \textit{v.s.} Qwen2.5-VL-7B & \up{17.6} & \up{18.0} & \up{18.6} & \up{6.5} & \up{3.0} & \up{9.9} & \up{6.2} & \up{6.0} & \up{6.4} \\
    \bottomrule
    \end{tabular}
    \vspace{3pt}
    \caption{
    Comparison with state-of-the-art alternatives on V* Bench~\cite{wu2024vstar} and HRBench~\cite{wang2025hrbench}.
    The best performance is highlighted in \textbf{bold}.
    }
    \label{tab:other_table}
    \vspace{-10pt}
\end{table*}

\begin{table*}[t]
    \centering\small
    \setlength{\tabcolsep}{5pt}
    \begin{tabular}{l c ccccc cccc}
    \toprule
    & & \multicolumn{5}{c}{Perception} & \multicolumn{4}{c}{Reasoning} \\
    \cmidrule(lr){3-7}
    \cmidrule(lr){8-11}
    & \textbf{Overall} & OCR & RS & DT & MO & AD & OCR & DT & MO & AD \\ 
    \midrule
    \multicolumn{10}{c}{\textbf{Private Models}} \\
    \midrule
    GPT-4o-2024-1120 & 46.4 & 81.0 & 45.2 & 65.0 & 34.1 & 37.0 & 72.0 & 50.3 & 42.0 & 33.0 \\
    \midrule
    \multicolumn{11}{c}{\textbf{General Models}} \\
    \midrule
    Qwen2.5-VL-7B & 42.3 & 87.6 & 32.7 & 83.0 & 27.3 & 30.0 & 72.0 & 62.0 & 28.7 & 23.0 \\
    Qwen2.5-VL-32B & 45.6 & 87.2 & 40.7 & 83.0 & 29.5 & 40.7 & 74.0 & 60.0 & 27.3 & 29.5 \\
    Qwen2.5-VL-72B & 43.7 & \textbf{90.8} & 34.0 & 87.0 & 27.9 & 30.6 & 74.0 & 61.0 & 26.7 & 25.5 \\
    LLaVA-OneVision-7B & 43.7 & 80.0 & 40.0 & 56.0 & 31.7 & 39.4 & 65.0 & 33.0 & 38.0 & 32.0 \\
    LLaVA-OneVision-72B & 48.7 & 79.2 & 50.7 & 67.0 & 37.9 & 40.0 & 76.0 & 41.0 & 38.7 & 39.3 \\
    InternVL3-8B & 47.9 & 83.6 & 49.3 & 75.0 & 34.5 & 36.9 & 70.0 & 44.0 & 40.0 & 37.0 \\
    InternVL3-38B & 51.0 & 85.6 & \textbf{56.0} & 71.0 & 42.6 & 40.0 & \textbf{77.0} & 45.0 & 47.3 & 35.0 \\
    InternVL3-78B & 52.3 & 87.6 & 54.7 & 77.0 & 42.6 & 36.6 & 76.0 & 56.0 & 46.0 & 40.3 \\
    \midrule
    \multicolumn{11}{c}{\textbf{Visual Grounded Reasoning Models}} \\
    \midrule
    Pixel-Reasoner-7B & 49.7 & 89.6 & 52.0 & 86.0 & 38.9 & 30.9 & 71.0 & \textbf{72.0} & 46.0 & 32.5 \\
    DeepEyes-7B & 53.2 & 90.0 & 52.7 & \textbf{89.0} & 43.3 & 33.4 & 76.0 & 69.0 & 44.0 & 35.0 \\
    \midrule
    \textbf{EGVOR (Ours)} & \textbf{55.4} & 87.8 & 51.8 & 83.2 & \textbf{48.4} & \textbf{44.9} & 74.5 & 66.7 & \textbf{52.2} & \textbf{40.1} \\
    $\Delta$ \textit{v.s.} Qwen2.5-VL-7B & \up{13.1} & \up{0.2} & \up{19.1} & \up{0.2} & \up{21.1} & \up{14.9} & \up{2.5} & \up{4.7} & \up{23.5} & \up{17.1} \\
    \bottomrule
    \bottomrule
    \end{tabular}
    \vspace{3pt}
    \caption{
    Comparison with state-of-the-art alternatives on MME-RealWorld-Lite~\cite{zhang2024mmerw}.
    The best performance is highlighted in \textbf{bold}.
    }
    \label{tab:mmerw}
    \vspace{-10pt}
\end{table*}

\begin{table*}[t]
    \centering\small
    \setlength{\tabcolsep}{3pt}
    \resizebox{1.0\textwidth}{!}
    {\begin{tabular}{l ccccccccccccc}
    \toprule
    & & & \rotatebox{60}{Attributes} & \rotatebox{60}{Material} & \rotatebox{60}{Phy. State} & \rotatebox{60}{Obj. Retr.} & \rotatebox{60}{OCR} & \rotatebox{60}{Per. Trans.} & \rotatebox{60}{Ordering} & \rotatebox{60}{Con. \& Oc.} & \rotatebox{60}{Spa. Cont.} & \rotatebox{60}{Comparison} \\
    \cmidrule(lr){4-8}
    \cmidrule(lr){9-13}
    & \textbf{Overall} & \textbf{mIoU} & \multicolumn{5}{c}{Perception} & \multicolumn{5}{c}{Reasoning} \\
    \midrule
    \multicolumn{13}{c}{\textbf{Private Models}} \\
    \midrule
    Gemini-2.5-Flash-0520 & 45.9 & -- & 48.3 & 53.9 & 69.6 & 68.8 & 75.0 & 15.3 & 19.3 & 56.1 & 72.4 & 43.2 \\
    GPT-4o-2024-1120 & 46.9 & -- & 51.7 & 61.5 & 65.2 & 43.8 & 69.1 & 18.8 & 38.6 & 48.8 & 72.4 & 43.2 \\
    Gemini-2.5-Pro-0605 & 54.1 & -- & 51.7 & 61.5 & 56.5 & 75.0 & 83.8 & 20.0 & 36.8 & 65.9 & 86.2 & 54.6 \\
    o3-0416 & 54.8 & -- & 69.0 & 69.2 & 65.2 & 68.8 & 79.4 & 22.4 & 38.6 & 61.0 & 86.2 & 50.0 \\ 
    \midrule
    \multicolumn{13}{c}{\textbf{Open-source General Models}} \\
    \midrule
    LLaVA-OneVision-7B & 37.3 & -- & 55.2 & 53.8 & 56.5 & 50.0 & 32.4 & 21.2 & 22.8 & 41.5 & 72.4 & 36.4 \\
    LLaVA-OneVision-72B & 40.5 & -- & 62.1 & 53.8 & 65.2 & 62.3 & 36.8 & 12.9 & 28.1 & 53.7 & 65.5 & \textbf{47.7} \\
    MiniCPM-V-2.6-8B & 30.1 & -- & 51.7 & 46.2 & 60.9 & 62.5 & 42.6 & 0.0 & 3.5 & 36.6 & 62.1 & 29.5 \\
    Qwen2.5-VL-7B & 37.0 & -- & 55.2 & 53.8 & 56.5 & 62.5 & 27.9 & 20.0 & 35.1 & 39.0 & 44.8 & 43.2 \\
    Qwen2.5-VL-32B & 42.5 & -- & 51.7 & 53.8 & 69.6 & 62.5 & 54.4 & 16.5 & 33.3 & 46.3 & 62.1 & 38.6 \\
    Qwen2.5-VL-72B & 42.2 & -- & 65.5 & \textbf{69.2} & 56.5 & 56.3 & 48.5 & 11.8 & 33.3 & 51.2 & 72.4 & 38.6 \\
    InternVL3-8B & 38.8 & -- & 51.7 & \textbf{69.2} & 56.5 & 56.3 & 33.7 & 21.2 & 24.6 & 39.0 & 72.4 & 43.2 \\
    InternVL3-38B & 42.0 & -- & 51.7 & 61.5 & 52.2 & \textbf{68.8} & 51.5 & 12.9 & 33.3 & 56.1 & 65.5 & 38.6 \\
    InternVL3-78B & 46.4 & -- & 62.1 & 61.5 & 52.2 & \textbf{68.8} & 52.9 & 16.5 & 33.3 & 61.0 & \textbf{86.2} & 45.5 \\
    \midrule
    \multicolumn{13}{c}{\textbf{Open-source Visual Grounded Reasoning Models}} \\
    \midrule
    DeepEyes-7B & 37.5 & 30.0 & 62.1 & 53.8 & 65.2 & 68.8 &  51.5 & 11.8 & 24.6 & 36.6 & 51.7 & \textbf{47.7} \\
    Pixel-Reasoner-7B & 39.0 & 35.7 & 58.6 & 61.5 & 65.2 & 50.0 & 48.5 & 14.1 & 31.6 & 39.0 & 44.8 & 40.9 \\
    \midrule
    \textbf{EGVOR (Ours)} & \textbf{51.3} & \textbf{45.0} & \textbf{69.1} & 54.4 & \textbf{83.4} & \textbf{69.3} & \textbf{64.1} & \textbf{23.2} & \textbf{38.6} & 58.7 & 69.4 & 45.8 \\
    $\Delta$ \textit{v.s.} Qwen2.5-VL-7B & \up{14.3} & -- & \up{13.9} & \up{0.6} & \up{26.9} & \up{6.8} & \up{36.2} & \up{3.2} & \up{3.5} & \up{19.7} & \up{24.6} & \up{2.6} \\
    \bottomrule
    \end{tabular}}
    \vspace{3pt}
    \caption{
    Selected results of different models on Treebench~\cite{wang2025traceable}.
    Evaluations of open-source general models are implemented using VLMEvalKit~\cite{duan2024vlmevalkit}, while evaluations of visual grounded reasoning models are conducted by us.
    Reasoning pathways of o3~\cite{o3} are unavailable, and thus traceable evaluations are \textit{not} valid.
    Best performances for open-source models are highlighted in \textbf{bold}. Our \textbf{EGVOR} achieves comparable performance with InternVL3-78B~\cite{zhu2025internvl3}. }
    \label{tab:bench_table}
    \vspace{-10pt}
\end{table*}

We evaluate a diverse array of MLLMs to benchmark complex urban cognition comprehensively. Departing from prior studies limited to edge models, our assessment spans from efficient 7B-parameter baselines to large-scale server-grade architectures and proprietary APIs.

\textit{1) Model Selection and Taxonomy.}
We evaluated a total of 18 distinctive MLLMs to analyze the scaling laws and architectural impacts on complex scene reasoning.
For the open-source category, we selected representative series including InternVL (InternVL2, 2.5, 3)~\cite{zhu2025internvl3,chen2024expanding,chen2024internvl}, Qwen-VL (Qwen2-VL~\cite{wang2024qwen2}, Qwen2.5-VL~\cite{bai2025qwen2}), MiniCPM (V2.5, V2.6, o2.6)~\cite{yao2024minicpm}, LLaVA (1.5, Next, OneVision)~\cite{liu2024llava1.5, li2024llava, liu2024llavanext} and Janus-Pro~\cite{chen2025janus}. To probe the capabilities of reasoning on larger-scale models, we extended the evaluation beyond the standard 7B/8B limit to include high-parameter variants such as Qwen2.5-VL-32B, Qwen2.5-VL-72B, InternVL3-34B, and InternVL3-78B. Furthermore, we incorporated specialized reasoning models including DeepEyes~\cite{zheng2025deepeyes} and PixelReasoner~\cite{su2025pixelreasoner} to assess domain-specific adaptations. In addition, we also benchmarked leading proprietary closed-source models to establish upper-bound performance references. This set includes Gemini-2.5-Flash~\cite{gemini-2.5-flash}, GPT-4.1-mini~\cite{gpt4o} and Claude-3.5-Sonnet. 

\textit{2) Evaluation and Training Environment.}
 
The evaluation pipeline for all open-source models was similarly standardized using MS-SWIFT~\cite{zhao2025swift} to ensure fair comparison. 
To guarantee deterministic reproducibility, we rigorously fixed the random seed and set the decoding temperature to 0 for all generation tasks. 
For the evaluation of open-ended reasoning, we adopted the ``LLM-as-a-Judge'' paradigm, utilizing gpt-5.2-2025-12-11 to quantify the semantic similarity between model predictions and ground truth annotations.
All experiments were conducted on a high-performance cluster equipped with 8$\times$NVIDIA H20 GPUs. 
For training, we implemented the SFT stage using the MS-SWIFT~\cite{zhao2025swift} framework (learning rate $5\mathrm{e}{-6}$, batch size 32) and the RL stage via EasyR1~\cite{zheng2025easyr1} (learning rate $1\mathrm{e}{-6}$).


\subsection{Dataset Construction for Training}
\label{subsec:training_data}

To effectively train EGVOR via the proposed two-stage curriculum, we constructed two bespoke datasets: SFT-Instruct for establishing the reasoning format, and RL-Instruct for optimizing cognitive alignment.

\textbf{SFT-Instruct Construction.}
We curated a high-fidelity dataset of 40K reasoning chains derived from VGR-158K~\cite{wang2025vgr}. To align with our atomic requirements, we implemented a rigorous restructuring pipeline:
(1) Coordinate Adaptation: We converted all normalized coordinates to absolute pixel values to fully leverage the high-resolution grounding capabilities of the Qwen2.5-VL backbone.
(2) Semantic Densification: To support robust evidence generation, we filter for multi-hop reasoning chains and explicitly augment sparse descriptions. This ensures that every spatial anchor $b_t$ is grounded by a detailed caption $c_t$, providing indispensable supervision for the latent alignment objective.
(3) Reflective Augmentation: To instill self-correction capabilities, we constructed a ``Reflective Subset'' (5K samples) via synthetic error injection. For valid reasoning steps, we randomly inserted distractor boxes followed by corrective thought tokens, training the model to dynamically detect and rectify grounding errors.

\textbf{RL-Instruct Construction.}
For the second stage, we constructed a high-quality dataset of 40K samples designed to provide dense supervision for the composite objective $J_{total}$. We employed a Dual-Hybridization Strategy to balance general reasoning with domain-specific robustness:
(1) General Hard Samples (30K): Sourced from V*-training set~\cite{wu2024vstar}, we utilized a teacher model (Gemini-2.5-Pro~\cite{gemini-2.5-pro}) to mine ``hard samples'' characterized by high visual entropy or complex spatial dependencies, ensuring the retention of broad multimodal capabilities.
(2) Complex Urban Samples (10K): To enforce entropy reduction in high-noise environments, we integrated samples from VisDrone~\cite{zhu2021visdrone} (UAV-view) and nuScenes~\cite{caesar2020nuscenes} (Ego-view). This hybrid composition prevents overfitting while forcing the policy to handle extreme visual confusion.
To support our reward functions, we enriched these samples with hierarchical metadata. We assigned adaptive importance weights to objects to facilitate the spatial reward $\Psi_{spatial}$, and generated diverse reasoning trajectories (e.g., bottom-up perceptual vs. top-down semantic paths) using advanced LLMs to support the path diversity objective $\Lambda_{path}$. Further details on the filtering protocols and annotation generation are provided in the Appendix~\ref{subsec:training_data}.

\subsection{Main Results on AD$^2$-Bench}
Table~\ref{tab:main_benchmark} presents the performance landscape of current MLLMs, spanning proprietary APIs, large-scale open-source architectures (up to 72B), and specialized reasoners. 

\textbf{Metric Setup.} To ensure a rigorous evaluation of cognitive reliability, the reported Ave. score denotes the arithmetic mean of accuracy-based metrics, excluding \textit{OCR-CER} (where lower is better) and the \textit{w/o CoE} setting. 
Crucially, for the \textit{Event Reasoning (w/ CoE)} metric, we employ a hybrid alignment protocol: the final score is a weighted sum of the \emph{Process Verification Score} ($\lambda=0.4$, assessing the quality of intermediate evidence) and the \emph{Decision Accuracy} ($\lambda=0.6$). This strict criterion penalizes models that arrive at correct answers via hallucinated reasoning paths.

\textbf{Baseline Analysis.}
Under this rigorous protocol, a discernible scaling law emerges: large-parameter models like {Qwen2.5-VL-72B} and {InternVL3-78B} exhibit robust generalization, achieving average scores of 60.67\% and 60.20\% respectively. Proprietary models (e.g., GPT, Gemini) similarly demonstrate strong capabilities in high-level reasoning.
However, a systemic ``Perception-Reasoning Disconnect" persists across all baselines. Even the leading open-source model, {Qwen2.5-VL-7B}, suffers drastic performance drops in \textit{Basic Perception} (e.g., 50.95\% in Detection), significantly lagging behind its logic capabilities. 
Furthermore, specialized models like {Pixel-Reasoner}\cite{su2025pixelreasoner}, while designed for grounding, fail to fully mitigate \textit{Spatial Ambiguity} in our adverse scenarios, underscoring the necessity for explicit evidence construction.

\textbf{Effectiveness of EGVOR.}
By explicitly constructing evidence chains, our {EGVOR} (7B) achieves a state-of-the-art average score of {67.66\%}, remarkably surpassing its 10x larger counterpart (Qwen2.5-VL-72B). 
In the \textit{Basic Perception} tier, EGVOR secures a {+13.3\%} gain over the baseline, with substantial improvements in \textit{Location} (+17.8\%) and \textit{Detection} (+18.8\%), validating that our ``Atomic Evidence" generation effectively anchors attention to valid targets amidst clutter.
In \textit{Advanced Perception}, the {+11.0\%} improvement, particularly in \textit{Attribute Recognition} (+14.7\%), confirms that verifying evidence details enhances resilience against feature degradation.
Crucially, in the final \textit{Event Reasoning} tier, we employ a hybrid alignment protocol for the \textit{w/ CoE} metric ($\lambda=0.4$ for process quality, $\lambda=0.6$ for decision accuracy) to penalize right-answers-for-wrong-reasons. On reasoning task, EGVOR outperforms the baseline by {+15.1\%} and specialized architectures like Pixel-Reasoner (+8.1\%), demonstrating that the explicit Chain of Evidence paradigm establishes a more trustworthy and explainable path for complex multimodal decision-making.

\subsection{Generalization to Diverse Benchmarks}
To verify that EGVOR has acquired robust cognitive capabilities rather than merely overfitting to the AD$^2$-Bench domain, we conducted extensive evaluations on three external benchmarks covering high-resolution perception, real-world complexity, and traceable reasoning.

\paragraph{Fine-Grained Perception in High-Resolution Scenarios.}
Standard MLLMs often suffer from \emph{attention dispersion} when processing high-resolution images or identifying minute details. As shown in Tab. \ref{tab:other_table}, EGVOR achieves state-of-the-art performance across all metrics, validating the efficacy of our explicit spatial anchoring mechanism.
Specifically, on V*-Bench, which evaluates the ability to locate and recognize small, hard-to-find objects, EGVOR outperforms the Qwen2.5-VL-7B by a remarkable margin of +17.6\%. This confirms that our \emph{Locate-Attend} paradigm effectively acts as a ``cognitive magnifying glass,'' enabling the model to actively search for and focus on subtle visual cues that are typically lost in the global pooling of standard encoders.
Furthermore, in the 4K and 8K scenarios of HR-Bench, where valid targets are sparse within vast backgrounds, our model maintains a significant lead (e.g., +9.9\% on 4K Cross). Notably, the improvement on the Cross subset—which involves multi-image or complex relational reasoning—is consistently higher than on the Single subset. This suggests that our \emph{Chain of Evidence} not only sharpens local perception but also enhances the capability to maintain context across large-scale visual fields, preventing the model from being overwhelmed by irrelevant background noise.

\paragraph{Robustness in Real-World Domains.}
We further evaluate the model on MME-RealWorld in Tab. \ref{tab:mmerw}, which spans diverse domains including OCR, Remote Sensing (RS), Diagram/Table (DT), Monitoring (MO), and Autonomous Driving (AD).
As expected, EGVOR exhibits dominant performance in Autonomous Driving (AD) and Monitoring (MO), with perception gains of +14.9\% and +21.1\%, respectively. These domains share the characteristics of complex dynamics and tiny objects with our training data, proving that the robustness learned from this data effectively transfers to general surveillance and driving tasks.
Surprisingly, we also observe a massive improvement in Remote Sensing (RS) perception (+19.1\%). Since RS imagery typically involves top-down views of tiny objects (similar to UAV views), this indicates that our model has learned a generalized feature representation for tiny object discovery, independent of the specific camera perspective (ego-centric vs. top-down).
Crucially, the gains in the Reasoning tier for MO (+23.5\%) and AD (+17.1\%) significantly outpace the perception gains. This improvement underscores that our method does not just ``see better'' but leverages this clearer vision to ``think deeper,'' effectively resolving complex causal chains in dynamic environments.

\paragraph{The Primacy of Grounding in Reasoning.}
Finally, we utilize TreeBench~\cite{wang2025traceable} to dissect the correlation between localization precision and reasoning accuracy. As illustrated in Tab. \ref{tab:bench_table}, EGVOR achieves the highest performance in Object Retrieval (+26.9\%) and Spatial Context (+24.6\%) tasks that demand rigorous spatial understanding. Consistent with our hypothesis, we observe a strong positive correlation between localization quality (mIoU) and reasoning accuracy. Unlike baseline models that may hallucinate correct answers from language priors (high Acc, low mIoU), EGVOR achieves high scores in both metrics simultaneously (51.3\% Overall Acc, 45.0\% mIoU).
The results further reveal a critical insight: while precise localization is necessary, it is not sufficient for complex reasoning. For instance, in Comparison and Ordering tasks, the performance gap is smaller compared to retrieval tasks. This suggests that while our model excels at \emph{grounding} (Stage I \& II), these higher-order logic tasks require \emph{second-order cognitive capabilities} beyond pure perception. Nevertheless, by ensuring that the foundational evidence is accurate (high mIoU), EGVOR provides a solid footing for these complex reasoning steps, significantly reducing the error propagation seen in baseline models.
\begin{figure*}[t]
    \centering
    \includegraphics[width=1\linewidth]{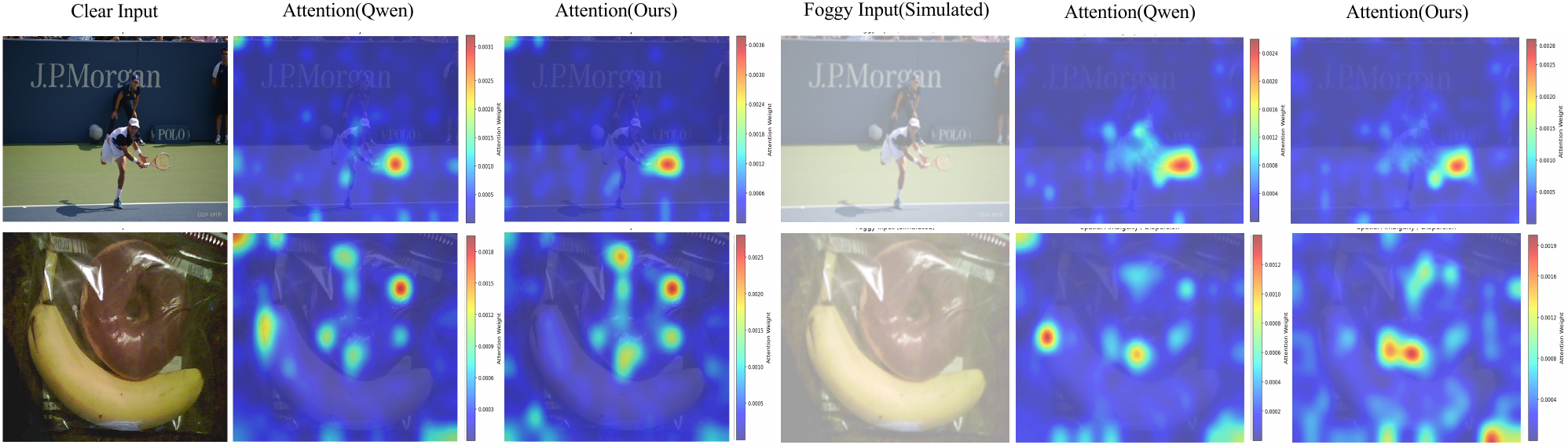}
    \vspace{-15pt}
    \caption{
\textcolor{black}{Qualitative comparison of attention maps for Qwen2.5-VL-7B and our EGVOR on clear and simulated foggy images.} 
}
    \label{fig:attention_vis}
    \vspace{-10pt}
\end{figure*}
\subsection{Holistic Multimodal Capabilities}

\begin{table*}[t]
\centering
\resizebox{\linewidth}{!}{%
\begin{tabular}{l|cccc|ccc|cc}
    \toprule
    \multirow{3}{*}{\textbf{Model}} & \multicolumn{9}{c}{\textbf{Multimodal Benchmark Performance}} \\
    \cmidrule(l){2-10}
    
    & \multicolumn{4}{c|}{\cellcolor{lightpurple}\textbf{Vision-centric QA}} 
    & \multicolumn{3}{c|}{\cellcolor{lightpink}\textbf{General VQA}} 
    & \multicolumn{2}{c}{\cellcolor{lightpurple}\textbf{Doc \& Chart}} \\
    
    \cmidrule(lr){2-5} \cmidrule(lr){6-8} \cmidrule(lr){9-10}
    & \textbf{CV-2D}~\cite{tong2024cambrian} & \textbf{CV-3D}~\cite{tong2024cambrian} & \textbf{MMVP}~\cite{tong2024eyes} & \textbf{RWQA}~\cite{xai2024grok} 
    & \textbf{MMBench}~\cite{liu2023mmbench} & \textbf{POPE}~\cite{li2023pope} & \textbf{Hallusion}~\cite{guan2024hallusionbench} 
    & \textbf{AI2D}~\cite{hiippala2021ai2d} & \textbf{ChartQA}~\cite{masry2022chartqa} \\
    \midrule
    
    Qwen2.5-VL-7B
    & 74.1 & 72.6 & 66.7 & 69.0 
    & 83.1 & 86.7 & 48.2 
    & \textbf{84.9} & 85.6 \\
    
    \rowcolor{backcolor}
    \textbf{EGVOR (Ours)}
    & \textbf{77.9} \up{3.8} & 79.1 \up{6.5} & \textbf{77.3} \up{10.6} & 72.1 \up{3.1}
    & 85.1 \up{2.0} & \textbf{87.8}\up{1.1} & 51.1\up{2.9}
    & 85.4\up{0.5} & 85.8\up{0.2} \\

    Qwen2.5-VL-72B
    & 77.7 & \textbf{87.0} & 66.7 & \textbf{75.7}
    & \textbf{88.6} & 84.9 & \textbf{55.2}
    & \textbf{88.7} & \textbf{89.5} \\
    \bottomrule
\end{tabular}%
}
\vspace{3pt}
\caption{
\textbf{Comparison with state-of-the-art alternatives.} 
We categorize the benchmarks into three groups: Vision-centric, General, and Document/Chart understanding.
RWQA: RealWorldQA. Hallusion: HallusionBench.
}
\label{tab:main_results}
\vspace{-10pt}
\end{table*}
To address concerns regarding \emph{catastrophic forgetting}, we evaluated EGVOR on a suite of 9 external benchmarks (Tab. \ref{tab:main_results}). Results demonstrate that EGVOR not only preserves general knowledge but reinforces the underlying visual cognition, yielding consistent gains across diverse domains.

\textbf{Generalizing Visual-Centric Reasoning.} 
We observe the most substantial improvements in tasks demanding rigorous perception. Notably, EGVOR achieves a +10.6\% gain on MMVP (visual puzzles) over the Qwen2.5-VL-7B baseline. This critically confirms that our learned capability—\emph{explicit evidence verification}—transcends domain boundaries, generalizing from traffic scenes to abstract visual logic. Similarly, gains on CV-3D (+6.5\%) and CV-2D (+3.8\%) validate that our spatial anchoring mechanism ($b_t$) effectively sharpens the model's fundamental geometry perception.

\textbf{Mitigating Hallucination via Grounding.} 
EGVOR consistently outperforms baselines on hallucination probes like POPE (+1.1\%) and HallusionBench (+2.9\%). These gains are directly attributed to the \emph{CoE} paradigm: by enforcing a ``Locate-then-Answer'' workflow, the model suppresses \emph{Blind Extrapolation} based on language priors. This grounding mechanism translates to more trustworthy open-ended generation, as evidenced by the +2.0\% improvement on MMBench.

\textbf{Surgical Optimization without Forgetting.} 
In text-intensive tasks (AI2D, ChartQA) relying on OCR and logic, EGVOR maintains a slight improvement (+0.5\%, +0.2\%). This stability serves as a crucial \emph{Anti-Forgetfulness} proof, indicating that our \emph{Cognitive Alignment} strategy is surgical—refining visual attention without overwriting pre-trained knowledge representations. While a capacity gap remains compared to the 72B variant due to world knowledge disparities, EGVOR (7B) narrows this gap solely through improved visual grounding, highlighting the method's efficiency.

\subsection{Ablation Study}

\begin{table*}[h]
    \centering
    \setlength{\tabcolsep}{5pt}
    \resizebox{\linewidth}{!}{
    \begin{tabular}{l c cccc | c | c c c c}
    \toprule
    & & \multicolumn{4}{c}{Optimization Objectives} & \multicolumn{1}{c}{\textbf{Target}} & \multicolumn{4}{c}{General Capabilities} \\
    \cmidrule(lr){3-6} \cmidrule(lr){7-7} \cmidrule(lr){8-11}
    
    Model & SFT & $\mathcal{R}_{\text{ans}}$ & $\Psi_{\text{spatial}}$ & $\Xi_{\text{align}}$ & $\Lambda_{\text{path}}$ & \textbf{AD$^2$-Bench} & V* & MME-RW & \multicolumn{2}{c}{TreeBench} \\
    & (Init) & (Acc) & (HRGW) & (Cap) & (Div) & Score & Acc & Acc & Acc & mIoU \\
    \midrule
    
    \grayrow Qwen2.5-VL-7B~\cite{bai2025qwen25vl} & -- & -- & -- & -- & -- & 55.0 & 71.2 & 42.3 & 37.0 & -- \\
    
    + SFT (Reflective) & \checkmark & -- & -- & -- & -- & 59.6 & 77.6 & 49.2 & 40.9 & 24.4 \\
    
    \midrule
    \multicolumn{11}{l}{\textit{Stage II: Cognitive Alignment (Incremental Components)}} \\
    
    + RL (Answer Prior) & \checkmark & \checkmark & -- & -- & -- & 61.2 & 86.4 & 51.2 & 40.2 & 27.7 \\
    
    + w/ Spatial Constraint & \checkmark & \checkmark & \checkmark & -- & -- & 65.5 & 90.8 & 53.9 & 49.9 & 43.8 \\
    
    + w/ Latent Align ($\Xi$) & \checkmark & \checkmark & \checkmark & \checkmark & -- & 66.1 & 91.1 & 54.6 & 50.8 & 44.2 \\
    
    \rowcolor{gray!15} 
    \textbf{EGVOR (Full)} & \checkmark & \checkmark & \checkmark & \checkmark & \checkmark & \textbf{67.7} & \textbf{91.9} & \textbf{55.4} & \textbf{51.3} & \textbf{45.0} \\
    
    \bottomrule
    \end{tabular}}
    \vspace{3pt}
    \caption{
    \textbf{Component-wise ablation study of EGVOR.} 
    We incrementally integrate the optimization objectives to validate their individual contributions.
    The baseline Qwen2.5-VL-7B results are aligned with standard benchmarks to ensure fair comparison.
    \textbf{$\mathcal{R}_{\text{ans}}$}: Basic answer correctness reward.
    \textbf{$\Psi_{\text{spatial}}$}: HRGW objective for robust spatial anchoring.
    \textbf{$\Xi_{\text{align}}$}: Latent alignment objective via caption consistency.
    \textbf{$\Lambda_{\text{path}}$}: Cognitive path diversity objective.
    }
    \label{tab:ablation_main}
    \vspace{-15pt}
\end{table*}

\begin{table}[h]
    \centering\small
    \setlength{\tabcolsep}{4pt}
    \resizebox{\linewidth}{!}{
    \begin{tabular}{l c ccc | c | c c}
    \toprule
    & \multicolumn{4}{c}{Spatial Reward Decomposition} & \textbf{Target} & \multicolumn{2}{c}{General (TreeBench)} \\
    \cmidrule(lr){2-5} \cmidrule(lr){6-6} \cmidrule(lr){7-8}
    
    Strategy & $\mathcal{R}_{\text{ans}}$ & $\Psi_{\text{rec}}^{\text{raw}}$ & $\Psi_{\text{rec}}^{\text{w}}$ & $\Psi_{\text{prec}}$ & \textbf{AD$^2$-Bench} & Acc & mIoU \\
    \midrule
    
    \grayrow Baseline (Ans Only) & \checkmark & -- & -- & -- & 61.2 & 40.2 & 27.7 \\
    
    \midrule
    
    + Unweighted Recall & \checkmark & \checkmark & -- & -- & 3.4 & 1.1 & 77.9 \\
    
    + Weighted Recall & \checkmark & -- & \checkmark & -- & 40.3 & 34.5 & 39.5 \\
    
    + Precision Only & \checkmark & -- & -- & \checkmark & 58.2 & 45.8 & 20.5 \\
    
    \midrule
    
    \rowcolor{gray!15} 
    \textbf{Full HRGW ($\Psi_{\text{spatial}}$)} & \checkmark & -- & \checkmark & \checkmark & \textbf{65.5} & \textbf{49.9} & 43.8 \\
    
    \bottomrule
    \end{tabular}}
    \vspace{3pt}
    \caption{
    \textbf{Ablation of the HRGW Spatial Objective.} 
    We dissect the impact of importance weighting and the trade-off between recall and precision.
    }
    \label{tab:ablation_spatial}
    \vspace{-10pt}
\end{table}
\begin{figure}[t]
    \centering
    \includegraphics[width=1\linewidth]{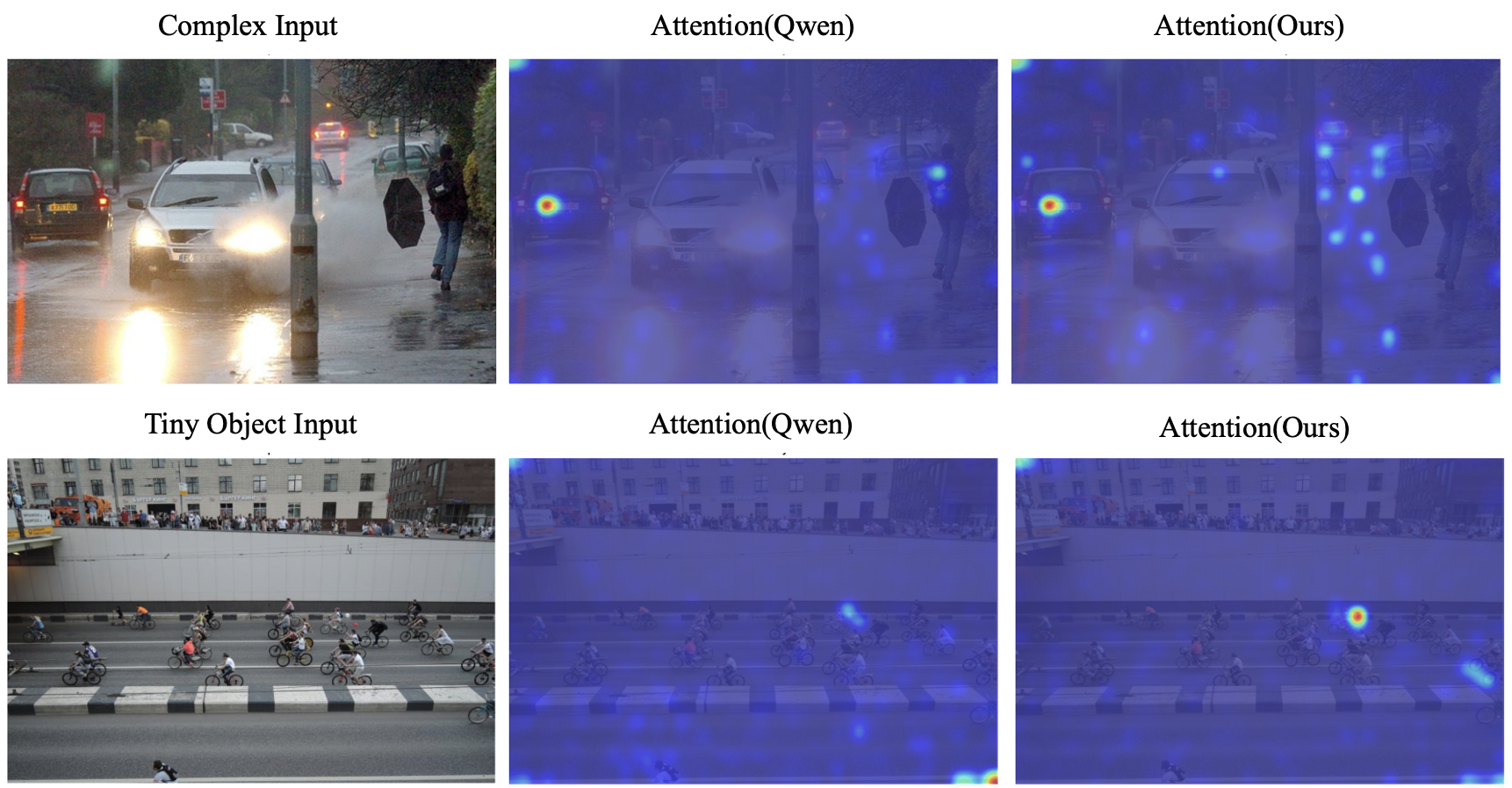}
    \vspace{-15pt}
    \caption{
Attention comparison on complex adverse and tiny-object scenes from AD$^2$-Bench (top) and V* (down).
}
    \label{fig:cross_dataset_vis}
    \vspace{-5pt}
\end{figure}
\begin{figure*}[t]
    \centering
    \includegraphics[width=1\linewidth]{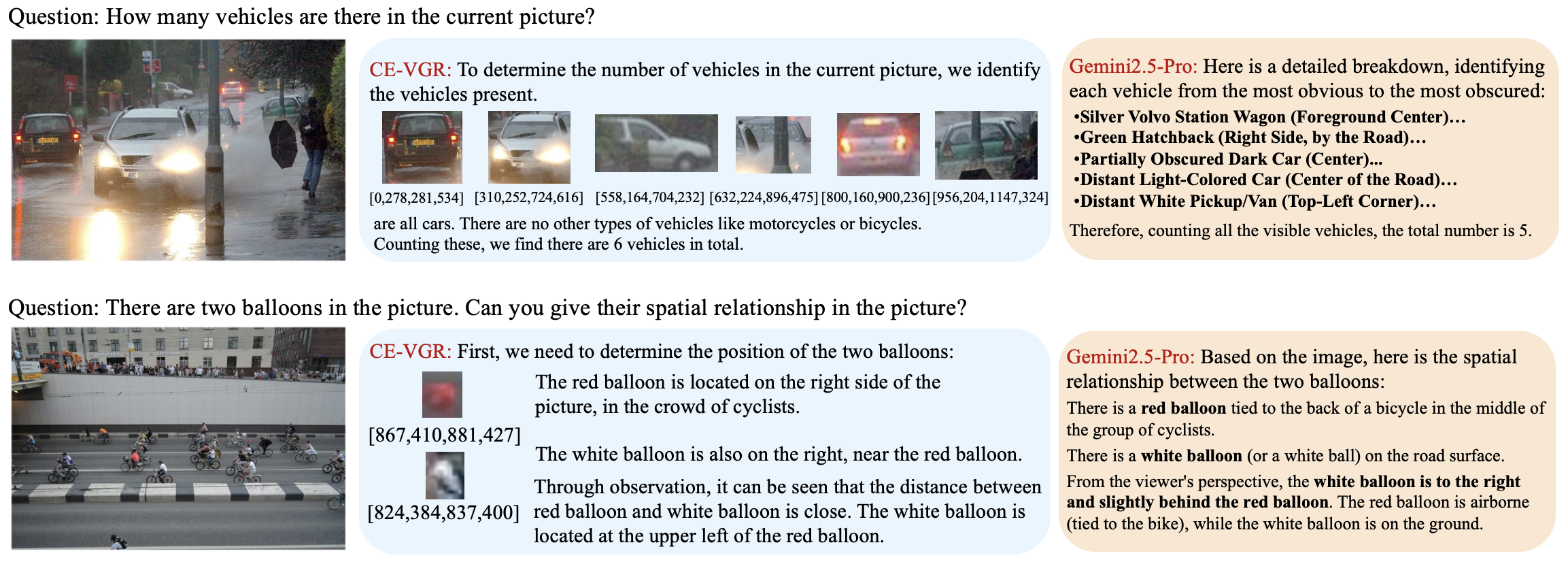}
    \vspace{-15pt}
    \caption{
Qualitative comparison of chain-of-evidence reasoning between EGVOR and Gemini2.5-Pro on an adverse occluded scene (top) and a tiny-object scene (bottom). 
}
    \label{fig:gemini_case}
    \vspace{-5pt}
\end{figure*}
We conducted a comprehensive ablation study to dissect the contribution of each component in EGVOR. The results presented in Tab.~\ref{tab:ablation_main}, \ref{tab:ablation_spatial} validate both the hierarchical training framework and the specific design of our spatial reward.

\paragraph{Component-wise Effectiveness.}
We incrementally integrated the optimization objectives to validate the hierarchical training framework.
Initially, the Reflective SFT stage serves as the foundation, initializing the reasoning format and yielding a moderate gain of +4.6\% on AD$^2$-Bench. However, the relatively low mIoU (24.4\%) on TreeBench indicates that at this stage, the model primarily mimics the output style without fully grasping precise localization.
Subsequently, introducing the basic answer reward ($\mathcal{R}_{ans}$) in the RL stage improves reasoning accuracy (+1.6\%), yet the localization capability (mIoU 27.7\%) lags behind. This discrepancy suggests that without explicit spatial constraints, the model tends to rely on language priors rather than visual evidence to answer questions.
The integration of the Spatial Constraint ($\Psi_{spatial}$) marks a pivotal turning point. It boosts mIoU significantly to 43.8\% (+16.1\%) and, crucially, drives the AD$^2$-Bench score to 65.5\%. This empirical evidence confirms our core hypothesis: enforcing precise visual grounding is the prerequisite for robust reasoning in adverse scenarios.
Finally, the addition of Alignment and Diversity objectives ($\Xi_{align}$ and $\Lambda_{path}$) further refines the policy, reducing hallucinations and encouraging diverse reasoning paths, culminating in the SOTA performance of 67.7\%.

\paragraph{Dynamics of Spatial Reward Shaping.}
Designing the spatial reward in RL is non-trivial due to the risk of the policy converging to degenerate strategies. We analyze the necessity of our HRGW objective through three distinct phases of agent behavior.
First, we identify a critical failure mode when using a naive Unweighted Recall reward. As shown in Tab. \ref{tab:ablation_spatial}, this encourages the agent to maximize coverage regardless of precision, leading to a trivial solution where the model predicts excessively large boxes (mIoU 77.9\%) to cover the entire image. While this satisfies the recall metric, it introduces catastrophic background noise, causing the reasoning accuracy to collapse to a mere 1.1\%.
To counter this, we introduce importance weighting ($\Psi_{rec}^w$) to penalize the inclusion of irrelevant background areas. This forces the model to abandon the ``canvas-covering'' strategy and focus on high-value regions. Consequently, the mIoU normalizes to 39.5\%, and reasoning accuracy recovers to 34.5\%. However, the predicted boxes remain loosely defined, adhering to a conservative policy that prioritizes coverage over tightness.
Finally, we achieve optimal performance by introducing the precision term ($\Psi_{prec}$). Acting as a spatial regularization force, it prunes the non-informative margins of the predicted boxes. Although this results in a similar mIoU (43.8\%), the reasoning accuracy on TreeBench jumps significantly to 49.9\%. This demonstrates that tighter, noise-free visual evidence significantly enhances the signal-to-noise ratio for the subsequent LLM reasoning process, validating the design of our full HRGW objective.

\paragraph{Robustness against Visual Disturbance.} To evaluate the stability of the reasoning process under environmental shifts, we analyze the Semantic Entropy Gap ($\Delta$) between clear and adverse weather conditions.
A larger gap implies that the model's confidence is heavily impacted by visual noise (e.g., rain or fog).
As shown in Tab. \ref{tab:entropy_comparison}, the standard CoT prompting exacerbates this instability ($\Delta=0.505$), as the model tends to hallucinate longer, uncertain reasoning chains when visual cues are ambiguous.
In contrast, EGVOR minimizes this gap to 0.278. This indicates that our spatial constraints effectively anchor the reasoning process to reliable visual evidence. Even in adverse conditions, the model maintains a confidence level ($H_{adverse}=1.036$) comparable to the baseline, but with significantly higher accuracy, demonstrating true cognitive robustness rather than blind confidence.

\paragraph{Efficiency and Token Economy.}
High-performance reasoning typically comes at the cost of increased inference latency (token length). We challenge this assumption by analyzing the \textbf{Marginal Utility} of generated tokens.
Tab. \ref{tab:efficiency_analysis} reveals a critical insight: standard text-based CoT is inefficient, consuming $3\times$ more tokens for a meager 2.5\% accuracy gain (Utility: 1.1).
While our SFT stage improves utility to 5.0, the RL stage achieves a breakthrough.
Surprisingly, the RL-optimized model not only achieves the highest accuracy (67.7\%) on AD$^2$-Bench, but does so with significantly fewer tokens than the SFT model ($144$ vs. $212$).
The results in an exceptional Marginal Utility of 51.7.
This phenomenon suggests that the RL objective, driven by the HRGW reward, forces the model to abandon verbose, irrelevant descriptions and focus solely on high-density visual evidence. The model learns to be \emph{concise yet precise}, making it highly suitable for latency-sensitive autonomous driving applications.

\paragraph{Visualization of Evidence-Guided Attention}

To better understand how EGVOR behaves under visual degradation, we visualize the cross-modal attention maps of Qwen2.5-VL-7B and our model on clear images and their simulated foggy counterparts (Fig.~\ref{fig:attention_vis}). On clear inputs, both models can roughly locate salient regions (Question: the racket in the tennis scene, or the donut in the grocery scene), but Qwen often spreads attention over background areas, whereas EGVOR produces more concentrated hotspots on the entities required to answer the question. After adding fog-like noise, Qwen's attention becomes scattered and drifts to irrelevant objects, consistent with the ``visual probability collapse'' in Sec.~\ref{sec:variance_analysis}, while EGVOR preserves tight, task-aligned focus on the same critical objects.
Beyond synthetic perturbations, we visualize attention on real complex scenes (AD$^2$-Bench) and tiny objects (V*), as shown in Fig.~\ref{fig:cross_dataset_vis}. 
In the AD$^2$-Bench rainy vehicle counting task, {Qwen} erroneously attends to irrelevant pedestrians, succumbing to semantic distraction. 
In contrast, {EGVOR} precisely anchors onto target vehicles, capturing even those heavily occluded by splashes or poles. 
Similarly, on V*, while Qwen disperses attention over the background, EGVOR locks onto the queried tiny cyclist. 
These visualizations confirm that explicit evidence atoms stabilize attention against distractions and occlusion, demonstrating robust generalization.

\paragraph{Case Studies}
Fig.~\ref{fig:gemini_case} compares EGVOR with Gemini2.5-Pro on two challenging examples. In the rainy street scene, EGVOR enumerates all vehicles by explicitly cropping six car instances and listing their coordinates, then concludes that there are six vehicles in total. In the tiny-object example, it first localizes the red and white balloons among dozens of cyclists, and then describes their relative positions based on the detected boxes, yielding a correct spatial relation.

Gemini2.5-Pro, in contrast, generates fluent natural-language explanations but without explicit localization. In the first case, it lists several vehicles yet still undercounts them, predicting five instead of six. In the second case, it misdescribes the relative position of the balloons. These case studies illustrate that, under adverse and small-object scenarios, ungrounded explanations can remain plausible but incorrect, while our explicit Chain-of-Evidence formulation makes the reasoning process verifiable and more reliable.
\begin{table}[t]
    \centering
\resizebox{\linewidth}{!}{
    \begin{tabular}{l c c c}
        \toprule
        \textbf{Model} & \textbf{$H_{clear}$} & \textbf{$H_{adverse}$} & \textbf{Gap ($\Delta$)} $\downarrow$ \\
        \midrule
        Qwen2.5-VL-7B & \textbf{0.5854} & 1.0270 & 0.4416 \\
        Qwen2.5-VL-7B + CoT Prompt & 1.3815 & 1.8863 & 0.5049 \\
        \rowcolor{gray!15}
        EGVOR (SFT) & 1.0348 & 1.3966 & 0.3618 \\
        \textbf{EGVOR (RL)} & 0.7580 & \textbf{1.0355} & \textbf{0.2776} \\
        \bottomrule
    \end{tabular}}
    \vspace{3pt}
    \caption{\textbf{Comparison of Semantic Entropy across different models.} $H_{clear}$ and $H_{adverse}$ denote the average token entropy under clear and adverse weather conditions, respectively. The \textbf{Entropy Gap} ($\Delta$) serves as a metric for robustness, where a lower value indicates higher stability against visual disturbances.}
    \label{tab:entropy_comparison}
    \vspace{-10pt}
\end{table}

\begin{table}[t]
    \centering
    \setlength{\tabcolsep}{5pt}
    \resizebox{\linewidth}{!}{
    \begin{tabular}{l l c c c}
        \toprule
        \textbf{Model} & \textbf{Paradigm} & \textbf{AD$^2$-Score} $\uparrow$ & \textbf{Avg. Tokens} & \textbf{Marginal Utility} $\uparrow$ \\
        & & (\%) & ($\mathcal{T}$) & ($\Delta \text{Acc} / \Delta \mathcal{T} \times 100$) \\
        \midrule
        
        Qwen2.5-VL-7B & Zero-Shot & 55.0 & \textbf{120} & -- \\
        
        + CoT Prompt & Text-Only CoT & 57.5 & 345 & 1.1 \\
        
        \midrule
        
        EGVOR (SFT) & Visual Evidence & 59.6 & 212 & 5.0 \\
        
        \rowcolor{gray!15}
        \textbf{EGVOR (RL)} & \textbf{Visual Evidence} & \textbf{67.7} & 144 & \textbf{51.7} \\
        \bottomrule
    \end{tabular}}
    \vspace{3pt}
    \caption{
        \textbf{Performance-to-Cost Efficiency Analysis.} 
        We compare the trade-off between reasoning performance and computational overhead.
    }
    \label{tab:efficiency_analysis}
    \vspace{-10pt}
\end{table}



\section{Conclusion and Future Work}
This work addresses the systemic ``dual opacity''—stemming from implicit black-box reasoning and outcome-oriented evaluation—by introducing AD$^2$-Bench and EGVOR.
Serving as a diagnostic prism, our benchmark identifies \emph{Spatial Ambiguity} and \emph{Semantic Uncertainty} as the primary bottlenecks of reasoning collapse. 
In response, EGVOR shifts the paradigm from implicit inference to the explicit construction of Evidence Atoms. 
By enforcing strict spatial-semantic alignment, our framework demonstrates that robust multimodal cognition is predicated on the verifiable acquisition of visual evidence, paving the way for trustworthy AI in complex real-world scenarios.

\textbf{Future Work.}
While EGVOR establishes a solid foundation for grounded reasoning, two promising directions remain for exploration:
(1) Spatiotemporal Evidence Chains: Currently, EGVOR focuses on frame-level reasoning. Future work will extend ``Evidence Atoms" into ``Evidence Tubes" to capture temporal dynamics. By tracking evidence consistency across video frames, we aim to resolve high-entropy scenarios involving motion ambiguity (e.g., predicting the trajectory of an occluded cyclist).
(2) Closed-loop Policy Integration: We plan to integrate the explainable \emph{Chain of Evidence} directly into end-to-end autonomous driving systems. Instead of merely outputting textual decisions, the explicit evidence representations ($z_{evidence}$) can serve as interpretable intermediate features to modulate control signals, enhancing the safety and transparency of unmanned systems.

\textbf{Data Availability Statements.}
We use the eleven publicly available datasets in our paper: V*~\cite{wu2024vstar}, HRBench~\cite{wang2025hrbench}, MME-RealWorld~\cite{zhang2024mmerw}, TreeBench~\cite{wang2025traceable}, our benchmark: AD$^2$-Bench, CV-Bench~\cite{tong2024cambrian}, MMVP~\cite{tong2024eyes}, RealworldQA~\cite{xai2024grok}, MMBench~\cite{liu2023mmbench}, POPE~\cite{li2023pope}, HallusionBench~\cite{guan2024hallusionbench},AI2D~\cite{hiippala2021ai2d}, ChartQA~\cite{masry2022chartqa}.
All benchmarks can be found on huggingface: \url{https://huggingface.co/datasets/{lmms-lab/vstar-bench, DreamMr/HR-Bench, yifanzhang114/MME-RealWorld-Lite, HaochenWang/TreeBench, ZhaoyangWei/AD2-Bench, ZzzHelloWorld/CV-Bench, MMVP/MMVP, lmms-lab/RealWorldQA, lmms-lab/MMBench, lmms-lab/POPE, lmms-lab/HallusionBench, lmms-lab/ai2d, lmms-lab/ChartQA}} 

\bibliographystyle{spbasic}
\bibliography{sn-bibliography}
\clearpage
\begin{appendices}

\section{Appendix}

\subsection{Rigorous Derivations for Probabilistic Bottleneck Analysis}
\label{app:prob_proofs}

\begin{table*}[t]
\centering
\small
\renewcommand{\arraystretch}{1.2} 
\caption{\textcolor{black}{Core notation used in Sec.~4. The table summarizes the physical meaning of the main probabilistic variables to improve readability.}}
\label{tab:notation}
\begin{tabularx}{\linewidth}{@{}>{\raggedright\arraybackslash}p{0.22\linewidth}X@{}} 
\toprule
\textbf{Symbol} & \textbf{Meaning} \\
\midrule
$\tilde{I}$ 
& Observed image under adverse degradation. \\

$Q$, $A$, $A_{\mathrm{gt}}$ 
& Query, predicted answer, and visually grounded correct answer. \\

$\mathcal{O}$ 
& Query-dependent set of candidate visual entities in the scene. \\

$O_f$, $o \in \mathcal{O}$ 
& Latent visual-focus random variable, and one realization of this focus. \\

$\mathcal{O}_{\mathrm{crit}}$, $\mathcal{O}_{\mathrm{irr}}$, $\mathcal{O}_{\mathrm{bg}}$ 
& Critical objects, irrelevant distractors, and background regions. \\

$z_o$, $Z$ 
& Deterministic feature of entity $o$, and random latent representation used in variance analysis. \\

$z_{\mathrm{global}}$, $Z_{\mathrm{global}}$ 
& Global context feature and its corresponding random variable. \\

$E$, $e$ 
& Structured evidence set and one evidence element. \\

$a_t = \langle b_t, h_t, c_t \rangle$ 
& Evidence atom: spatial anchor, region-aware latent state, and grounded description. \\

$\Psi_{\mathrm{spatial}}$, $\Xi_{\mathrm{align}}$, $\Lambda_{\mathrm{path}}$ 
& Spatial reward, semantic alignment reward, and path-diversity reward. \\
\bottomrule
\end{tabularx}
\end{table*}

\textcolor{black}{This section provides the detailed derivations supporting the probabilistic analysis in Sec.~\ref{subsec:preliminaries}. The goal is not to claim an exact generative model of MLLMs, but to make explicit the assumptions under which the diagnostic statements in the main text hold.}

\subsubsection{Latent Focus Variable and Object-Centric Mixture}

\textcolor{black}{Let $\mathcal{O}$ be the query-dependent set of candidate visual entities. We define $O_f$ as a discrete latent visual-focus random variable with support $\mathcal{O}$. A lowercase $o\in\mathcal{O}$ denotes a realization of $O_f$. We write
\begin{equation}
\small
P(o|\tilde{I},Q)
\equiv
P(O_f=o|\tilde{I},Q)
\end{equation}
for notational simplicity.}

\textcolor{black}{Assuming that $z_o$ is the deterministic latent representation associated with entity $o$ under the observed image $\tilde I$, the object-centric posterior approximation is
\begin{equation}
\small
P(A|\tilde{I},Q)
=
\sum_{o\in\mathcal{O}}
P(A|z_o,Q)P(o|\tilde{I},Q).
\end{equation}
This approximation abstracts the model's implicit visual attention as a distribution over candidate entities.}

\subsubsection{Posterior Suppression under Critical-Evidence Mass Shift}

\textcolor{black}{Let $A_{gt}$ denote the visually grounded correct answer. We partition the candidate entity set into critical and non-critical subsets:
\begin{equation}
\small
\mathcal{O}
=
\mathcal{O}_{crit}
\cup
\mathcal{O}_{non},
\quad
\mathcal{O}_{non}
=
\mathcal{O}_{irr}
\cup
\mathcal{O}_{bg}.
\end{equation}
The critical-evidence mass is defined as
\begin{equation}
\small
\rho
=
P(O_f\in\mathcal{O}_{crit}|\tilde{I},Q).
\end{equation}}

\textcolor{black}{For a local redistribution analysis, we decompose the focus distribution into two within-set distributions:
\begin{equation}
\small
q_{crit}(o)
=
P(O_f=o|\tilde{I},Q,O_f\in\mathcal{O}_{crit}),
\quad
o\in\mathcal{O}_{crit},
\end{equation}
and
\begin{equation}
\small
q_{non}(o)
=
P(O_f=o|\tilde{I},Q,O_f\in\mathcal{O}_{non}),
\quad
o\in\mathcal{O}_{non}.
\end{equation}
For $0<\rho<1$, define a family of focus distributions:
\begin{equation}
\label{eq:rho_family_app}
\small
P_{\rho}(O_f=o|\tilde{I},Q)
=
\begin{cases}
\rho q_{crit}(o),
& o\in\mathcal{O}_{crit},
\\
(1-\rho) q_{non}(o),
& o\in\mathcal{O}_{non}.
\end{cases}
\end{equation}
This family preserves the relative distribution within the critical and non-critical subsets, while varying the total probability mass assigned to each subset.}

\textcolor{black}{Define the average support for $A_{gt}$ from critical and non-critical regions as
\begin{equation}
\small
\bar p_{crit}
=
\sum_{o\in\mathcal{O}_{crit}}
P(A_{gt}|z_o,Q)q_{crit}(o),
\end{equation}
and
\begin{equation}
\small
\bar p_{non}
=
\sum_{o\in\mathcal{O}_{non}}
P(A_{gt}|z_o,Q)q_{non}(o).
\end{equation}
The critical-evidence separability assumption is
\begin{equation}
\label{eq:critical_sep_app}
\small
\bar p_{crit}
>
\bar p_{non}.
\end{equation}
This assumption states that query-critical evidence supports the visually grounded answer more strongly than non-critical evidence.}

\textcolor{black}{Under Eq.~\ref{eq:rho_family_app}, the posterior support for $A_{gt}$ becomes
\begin{equation}
\label{eq:posterior_rho_app}
\small
P_{\rho}(A_{gt}|\tilde{I},Q)
=
\rho\bar p_{crit}
+
(1-\rho)\bar p_{non}.
\end{equation}
For two critical-evidence masses $0<\rho_2<\rho_1<1$, subtracting Eq.~\ref{eq:posterior_rho_app} gives
\begin{equation}
\small
\begin{aligned}
&
P_{\rho_2}(A_{gt}|\tilde{I},Q)
-
P_{\rho_1}(A_{gt}|\tilde{I},Q)
\\
&=
(\rho_2-\rho_1)
(\bar p_{crit}-\bar p_{non}).
\end{aligned}
\end{equation}
Since $\rho_2-\rho_1<0$ and $\bar p_{crit}-\bar p_{non}>0$, we obtain
\begin{equation}
\small
P_{\rho_2}(A_{gt}|\tilde{I},Q)
<
P_{\rho_1}(A_{gt}|\tilde{I},Q).
\end{equation}
Thus, reducing the critical-evidence mass strictly suppresses the posterior probability of the visually grounded answer under the separability assumption.}

\subsubsection{Semantic Entropy under Critical-Evidence Mass Shift}

\textcolor{black}{Let $\mathcal{A}$ denote the answer space. Define the answer distributions induced by critical and non-critical evidence as
\begin{equation}
\small
p_{crit}(a)
=
P(A=a|O_f\in\mathcal{O}_{crit},\tilde{I},Q),
\end{equation}
and
\begin{equation}
\small
p_{non}(a)
=
P(A=a|O_f\in\mathcal{O}_{non},\tilde{I},Q).
\end{equation}
The answer distribution induced by critical-evidence mass $\rho$ is
\begin{equation}
\label{eq:answer_mixture_app}
\small
p_{\rho}(a)
=
\rho p_{crit}(a)
+
(1-\rho)p_{non}(a).
\end{equation}
We define
\begin{equation}
\label{eq:H_rho_app}
\small
H_{\rho}(A|\tilde{I},Q)
\equiv
-
\sum_{a\in\mathcal{A}}
p_{\rho}(a)\log p_{\rho}(a).
\end{equation}
Assuming $p_{\rho}(a)>0$ on the considered local interval, differentiating Eq.~\ref{eq:H_rho_app} with respect to $\rho$ gives
\begin{equation}
\small
\begin{aligned}
\frac{d H_{\rho}(A|\tilde{I},Q)}{d\rho}
&=
-
\sum_{a\in\mathcal{A}}
\frac{d p_{\rho}(a)}{d\rho}
\left(1+\log p_{\rho}(a)\right)
\\
&=
-
\sum_{a\in\mathcal{A}}
\left(p_{crit}(a)-p_{non}(a)\right)
\left(1+\log p_{\rho}(a)\right).
\end{aligned}
\end{equation}
Because both $p_{crit}$ and $p_{non}$ are probability distributions,
\begin{equation}
\small
\sum_{a\in\mathcal{A}}
\left(p_{crit}(a)-p_{non}(a)\right)
=
1-1
=
0.
\end{equation}
Therefore, the constant term cancels, yielding
\begin{equation}
\label{eq:entropy_derivative_app}
\small
\frac{d H_{\rho}(A|\tilde{I},Q)}{d\rho}
=
-
\sum_{a\in\mathcal{A}}
\left(p_{crit}(a)-p_{non}(a)\right)
\log p_{\rho}(a).
\end{equation}}

\textcolor{black}{The uncertainty-dominant failure regime is defined as a local interval $[\rho_2,\rho_1]$ where
\begin{equation}
\label{eq:uncertainty_regime_app}
\small
\frac{d H_{\rho}(A|\tilde{I},Q)}{d\rho}
<
0,
\quad
\forall \rho\in[\rho_2,\rho_1].
\end{equation}
For $0<\rho_2<\rho_1<1$, the fundamental theorem of calculus gives
\begin{equation}
\small
\begin{aligned}
H_{\rho_2}(A|\tilde{I},Q)
-
H_{\rho_1}(A|\tilde{I},Q)
&=
-
\int_{\rho_2}^{\rho_1}
\frac{d H_u(A|\tilde{I},Q)}{du}
\,du.
\end{aligned}
\end{equation}
Under Eq.~\ref{eq:uncertainty_regime_app}, the integral is negative, and hence
\begin{equation}
\small
H_{\rho_2}(A|\tilde{I},Q)
>
H_{\rho_1}(A|\tilde{I},Q).
\end{equation}
This establishes a conditional statement for semantic entropy growth: when non-critical regions induce diverse inconsistent answer hypotheses, reducing the critical-evidence mass increases the answer entropy.}

\textcolor{black}{This result is intentionally local and conditional. In the prior-dominant hallucination regime, the model may instead become confidently wrong, in which case the entropy need not increase. That regime is captured by the prior-collapse inequality in Sec.~\ref{subsec:preliminaries}.}

\subsubsection{Conditional-Independence Form of Prior Collapse}

\textcolor{black}{Let $Z_{\mathrm{global}}$ denote the random global visual representation and $z_{\mathrm{global}}$ one of its realizations. In the exact noise-dominant limiting case, assume
\begin{equation}
\small
Z_{\mathrm{global}}
\mathrel{\perp\!\!\!\perp}
A
\mid Q.
\end{equation}
Then, for all $z_{\mathrm{global}}$ with positive density,
\begin{equation}
\small
P(z_{\mathrm{global}}|A,Q)
=
P(z_{\mathrm{global}}|Q).
\end{equation}
Applying Bayes' rule,
\begin{equation}
\small
\begin{aligned}
P(A|z_{\mathrm{global}},Q)
&=
\frac{
P(z_{\mathrm{global}}|A,Q)P(A|Q)
}{
P(z_{\mathrm{global}}|Q)
}
\\
&=
P(A|Q).
\end{aligned}
\end{equation}
Thus, in the exact conditional-independence limit, the posterior collapses to the language prior. In practice, the condition is approximate, so the equality is replaced by the approximation
\begin{equation}
\small
P(A|z_{\mathrm{global}},Q)
\approx
P(A|Q).
\end{equation}
When the prior-favored answer $A_{\mathrm{prior}}$ differs from the visually grounded answer $A_{gt}$, this prior collapse can produce
\begin{equation}
\small
P(A_{\mathrm{prior}}|Q)
>
P(A_{gt}|\tilde{I},Q).
\end{equation}}

\subsubsection{Support-Truncation Bound of Spatial Anchors}

\textcolor{black}{For an evidence atom $a_t=\langle b_t,h_t,c_t\rangle$, let $\mathcal{O}(b_t)\subseteq\mathcal{O}$ denote the subset of entities whose spatial support lies inside the region defined by $b_t$. Conditioning on $b_t$ induces the truncated distribution
\begin{equation}
\small
P(o|b_t,\tilde{I},Q)
=
\begin{cases}
\frac{
P(o|\tilde{I},Q)
}{
\sum_{o'\in\mathcal{O}(b_t)}
P(o'|\tilde{I},Q)
},
& o\in\mathcal{O}(b_t),
\\
0,
& o\notin\mathcal{O}(b_t).
\end{cases}
\end{equation}
The support of this distribution is contained in $\mathcal{O}(b_t)$. For any discrete distribution supported on a finite set $\mathcal{S}$, its entropy is upper-bounded by $\log|\mathcal{S}|$. Therefore,
\begin{equation}
\small
H(O_f|b_t,\tilde{I},Q)
\le
\log|\mathcal{O}(b_t)|.
\end{equation}
Since the full evidence atom $a_t$ includes the anchor $b_t$, its focus support is also contained in $\mathcal{O}(b_t)$, yielding
\begin{equation}
\small
H(O_f|a_t,\tilde{I},Q)
\le
\log|\mathcal{O}(b_t)|.
\end{equation}
For the unanchored global distribution,
\begin{equation}
\small
H(O_f|\tilde{I},Q)
\le
\log|\mathcal{O}|.
\end{equation}
Thus, when $b_t$ is compact and valid, $|\mathcal{O}(b_t)|\ll|\mathcal{O}|$, and the worst-case entropy bound under the anchor is much smaller than the global worst-case bound:
\begin{equation}
\small
\log|\mathcal{O}(b_t)|
\ll
\log|\mathcal{O}|.
\end{equation}}

\textcolor{black}{This is a support-based bound, not a claim that conditioning on any arbitrary box always reduces the actual entropy. The conclusion holds for valid compact anchors, which is precisely what the spatial grounding objective in EGVOR is designed to encourage.}

\subsubsection{Explicit Distribution in the Law of Total Variance}

\textcolor{black}{Let $Z$ be the latent representation induced by the visual focus. For scalar $Z$, the standard Law of Total Variance gives
\begin{equation}
\begin{aligned}
\small
\operatorname{Var}(Z|\tilde{I},Q)
=
\mathbb{E}_{O_f\sim P(\cdot|\tilde{I},Q)}
\left[
\operatorname{Var}(Z|O_f,\tilde{I},Q)
\right]  \\
+
\operatorname{Var}_{O_f\sim P(\cdot|\tilde{I},Q)}
\left(
\mathbb{E}[Z|O_f,\tilde{I},Q]
\right).
\end{aligned}
\end{equation}
For vector-valued $Z$, the same decomposition applies to covariance matrices; in the main text, $\operatorname{Var}(Z)$ can be read as the trace of the covariance matrix or the variance of a projected feature statistic.}

\textcolor{black}{Under an anchor $b_t$, the focus distribution becomes $P(\cdot|b_t,\tilde{I},Q)$, whose support is contained in $\mathcal{O}(b_t)$. Define
\begin{equation}
\small
m(o)
=
\mathbb{E}[Z|O_f=o,\tilde{I},Q].
\end{equation}
The localized spatial variance term is
\begin{equation}
\small
V_{\mathrm{sp}}(b_t)
=
\operatorname{Var}_{O_f\sim P(\cdot|b_t,\tilde{I},Q)}
\left(
m(O_f)
\right).
\end{equation}
For vector-valued $m(O_f)$, using the trace covariance identity
\begin{equation}
\small
\operatorname{tr}\operatorname{Cov}(X)
=
\frac{1}{2}\mathbb{E}\|X-X'\|_2^2,
\end{equation}
where $X'$ is an independent copy of $X$, we obtain the support-diameter bound
\begin{equation}
\small
V_{\mathrm{sp}}(b_t)
\le
\frac{1}{2}
\operatorname{diam}
\left(
\{m(o):o\in\mathcal{O}(b_t)\}
\right)^2.
\end{equation}
Thus, a compact and query-critical anchor reduces the worst-case spatial-variance support. This formalizes the role of $b_t$ in suppressing Spatial Boundary Ambiguity. The semantic description $c_t$ complements this by encouraging the local representation $h_t$ to align with a stable semantic concept, thereby targeting the Intrinsic Semantic Uncertainty term.}

\subsection{Why Wasserstein-2 Distance Provides Dense Spatial Reward}
\label{app:iou_wasserstein}

\textcolor{black}{This section provides additional explanation for the HRGW spatial reward in Sec.~4.3.2. Intersection-over-Union (IoU) is widely used as a localization metric, but it is less suitable as a reinforcement-learning reward in the early exploration stage. For two bounding boxes $b_{\mathrm{pred}}$ and $b_{\mathrm{gt}}$, IoU is defined as
\begin{equation}
\small
\operatorname{IoU}(b_{\mathrm{pred}},b_{\mathrm{gt}})
=
\frac{
|b_{\mathrm{pred}}\cap b_{\mathrm{gt}}|
}{
|b_{\mathrm{pred}}\cup b_{\mathrm{gt}}|
}.
\end{equation}
When the two boxes are disjoint,
\begin{equation}
\small
b_{\mathrm{pred}}\cap b_{\mathrm{gt}}=\emptyset
\quad\Longrightarrow\quad
\operatorname{IoU}(b_{\mathrm{pred}},b_{\mathrm{gt}})=0.
\end{equation}
Therefore, IoU assigns the same reward to all non-overlapping predictions, regardless of whether the predicted box is close to the target or far away from it. This creates a sparse reward landscape: a near-miss prediction and a completely irrelevant prediction are indistinguishable as long as neither overlaps with the ground truth. Such sparsity is undesirable for policy optimization, because the model receives no directional signal for how to move the predicted anchor toward the target.}

\textcolor{black}{To obtain dense geometric feedback, HRGW represents each bounding box as a 2D Gaussian distribution. For a box
\begin{equation}
\small
b=(x_c,y_c,w,h),
\end{equation}
where $(x_c,y_c)$ is the box center and $(w,h)$ are its width and height, we map it to
\begin{equation}
\small
b
\mapsto
\mathcal N(\mu,\Sigma),
\quad
\mu=
\begin{bmatrix}
x_c\\
y_c
\end{bmatrix},
\quad
\Sigma=
\begin{bmatrix}
w^2/4 & 0\\
0 & h^2/4
\end{bmatrix}.
\end{equation}
The exact scale of $\Sigma$ only rescales the distance and is absorbed by the normalization constant $C$ in the normalized affinity.}

\textcolor{black}{For two Gaussian boxes
\begin{equation}
\small
\mathcal N_{\mathrm{pred}}
=
\mathcal N(\mu_p,\Sigma_p),
\quad
\mathcal N_{\mathrm{gt}}
=
\mathcal N(\mu_g,\Sigma_g),
\end{equation}
the squared Wasserstein-2 distance has the closed form
\begin{equation}
\begin{aligned}
\small
W_2^2(\mathcal N_{\mathrm{pred}},\mathcal N_{\mathrm{gt}})
=
\|\mu_p-\mu_g\|_2^2 \\
+
\operatorname{Tr}
\left(
\Sigma_p+\Sigma_g
-
2(\Sigma_g^{1/2}\Sigma_p\Sigma_g^{1/2})^{1/2}
\right).
\end{aligned}
\end{equation}
where $\operatorname{Tr}(\cdot)$ denotes the trace of a square matrix, i.e., the sum of its diagonal elements. The first term measures the displacement between box centers, while the second term measures the mismatch between box scale and aspect ratio. Thus, unlike IoU, $W_2$ remains informative even when two boxes do not overlap.}

\textcolor{black}{In the axis-aligned case where the covariance matrices are diagonal, the Wasserstein-2 distance can be further simplified. Let
\begin{equation}
\small
\Sigma_p=
\operatorname{diag}(\sigma_{x,p}^2,\sigma_{y,p}^2),
\quad
\Sigma_g=
\operatorname{diag}(\sigma_{x,g}^2,\sigma_{y,g}^2).
\end{equation}
Then
\begin{equation}
\begin{aligned}
\small
W_2^2(\mathcal N_{\mathrm{pred}},\mathcal N_{\mathrm{gt}})
=
\|\mu_p-\mu_g\|_2^2
+
(\sigma_{x,p}-\sigma_{x,g})^2 \\
+
(\sigma_{y,p}-\sigma_{y,g})^2.
\end{aligned}
\end{equation}
This expression shows that the distance provides a continuous penalty for both center displacement and size mismatch. Consequently, a disjoint but nearby prediction receives a smaller distance than a far-away prediction, which gives the policy a meaningful optimization direction.}

\textcolor{black}{We convert this distance into a normalized affinity:
\begin{equation}
\small
A(b_{\mathrm{pred}},b_{\mathrm{gt}})
=
\exp
\left(
-
\frac{
\sqrt{
W_2^2(\mathcal N_{\mathrm{pred}},\mathcal N_{\mathrm{gt}})
}
}{C}
\right).
\end{equation}
This affinity satisfies
\begin{equation}
\small
0<A(b_{\mathrm{pred}},b_{\mathrm{gt}})\le 1,
\end{equation}
where the score approaches $1$ when the predicted and ground-truth boxes are geometrically aligned, and decreases smoothly as their center or shape discrepancy increases. Therefore, HRGW provides dense reward shaping for spatial anchoring, especially in adverse scenes where early predictions are frequently disjoint from the target. This dense signal helps the policy gradually move predicted anchors toward $\mathcal O_{\mathrm{crit}}$, while avoiding the zero-feedback issue of IoU-based rewards.}

\subsection{Generalization on the Latest MLLMs}
\label{sec:qwen3.5_generalization}



\textcolor{black}{To examine the generalization of EGVOR on stronger MLLMs, we apply it to \textbf{Qwen3.5-9B}. As shown in Table~\ref{tab:qwen3.5_generalization}, although the base model benefits from stronger pretrained reasoning, it still shows clear spatial weaknesses, such as only 11.8\% on Perspective Transform. SFT improves format following and boosts Object Retrieval from 68.8\% to 87.5\%, but also causes drops in OCR and Contact \& Occlusion, suggesting that format imitation alone is insufficient. With RL-based spatial and semantic rewards, EGVOR recovers these drops and further improves complex reasoning, achieving gains of +17.5\% in Ordering, +11.7\% in Perspective Transform, and 51.3\% mIoU. These results show that EGVOR remainseffective and complementary to stronger base models.}

\textcolor{black}{We further evaluate the generalization ability of EGVOR on the Qwen3.5-9B backbone. As shown in Table A3, the original Qwen3.5-9B achieves an average score of 59.0, while supervised fine-tuning already improves it to 64.6, indicating that Qwen3.5 benefits more from the SFT stage due to its stronger pretrained reasoning ability. After applying the full EGVOR pipeline, the performance further increases to 69.4, yielding an overall improvement of +10.4 over the original model. The gains are especially pronounced in Basic Perception (+17.3), mainly driven by large improvements in location (+24.3) and detection (+31.5), while Relation Understanding (+6.4) and Event Reasoning (+9.2) also improve consistently. These results suggest that EGVOR can effectively transfer to a stronger reasoning-capable backbone and generalize beyond the single baseline.}
\begin{table*}[h]
\centering
\small
\resizebox{\textwidth}{!}{
\begin{tabular}{lccccccccccccc}
\toprule
& & & \multicolumn{5}{c}{Perception} & & \multicolumn{5}{c}{Reasoning} \\
\cmidrule{4-8} \cmidrule{10-14}
\textbf{Model} & \textbf{Overall} & \textbf{mIoU} & 
\rotatebox{45}{Attributes} & \rotatebox{45}{Material} & \rotatebox{45}{Phy. State} & \rotatebox{45}{Obj. Retr.} & \rotatebox{45}{OCR} & &
\rotatebox{45}{Per. Trans.} & \rotatebox{45}{Ordering} & \rotatebox{45}{Con. \& Oc.} & \rotatebox{45}{Spa. Cont.} & \rotatebox{45}{Comparison} \\
\midrule
Qwen3.5-9B (Base) & 44.0 & -- & 58.6 & 46.2 & 65.2 & 68.8 & \textbf{75.0} & & 11.8 & 21.1 & 53.7 & 72.4 & 36.4 \\
+ SFT (Reflective) & 46.5 & 21.1 & 62.1 & 61.5 & 69.6 & 87.5 & 66.2 & & 11.8 & 22.8 & 41.5 & 62.1 & 50.0 \\
\textbf{+ RL (EGVOR Full)} & \textbf{54.1} & \textbf{51.3} & \textbf{69.0} & \textbf{53.8} & \textbf{78.3} & \textbf{81.2} & 73.5 & & \textbf{23.5} & \textbf{38.6} & \textbf{61.0} & \textbf{79.3} & \textbf{47.7} \\
\midrule
$\Delta$ (EGVOR \textit{v.s.} Base) & \textcolor{green!70!black}{$\uparrow 10.1$} & \textcolor{green!70!black}{$\uparrow 51.3$} & \textcolor{green!70!black}{$\uparrow 10.4$} & \textcolor{green!70!black}{$\uparrow 7.6$} & \textcolor{green!70!black}{$\uparrow 13.1$} & \textcolor{green!70!black}{$\uparrow 12.4$} & \textcolor{gray}{$\downarrow 1.5$} & & \textcolor{green!70!black}{$\uparrow 11.7$} & \textcolor{green!70!black}{$\uparrow 17.5$} & \textcolor{green!70!black}{$\uparrow 7.3$} & \textcolor{green!70!black}{$\uparrow 6.9$} & \textcolor{green!70!black}{$\uparrow 11.3$} \\
\bottomrule
\end{tabular}
}
\caption{\textcolor{black}{Generalization of EGVOR on the Qwen3.5-9B architecture (TreeBench).}}
\label{tab:qwen3.5_generalization}
\vspace{-10pt}
\end{table*}

\begin{table*}[htbp]
\centering
\setlength{\tabcolsep}{2.5pt}
\renewcommand{\arraystretch}{1.15}
\resizebox{1\textwidth}{!}{%
\begin{tabular}{l|c|c|cccc|ccccc|cccc|cc}
\toprule

\multirow{2}{*}{\textbf{Model}} & 
\multirow{2}{*}{\textbf{Ave.}} & 
\multirow{2}{*}{\textbf{LLM}} & 
\cellcolor{color_obs}\rotatebox{60}{\textit{Existence}} & 
\cellcolor{color_obs}\rotatebox{60}{\textit{Counting}} & 
\cellcolor{color_obs}\rotatebox{60}{\textit{Location}} & 
\cellcolor{color_obs}\rotatebox{60}{\textit{Detection}} & 
\cellcolor{color_sym}\rotatebox{60}{\textit{OCR-F1}} & 
\cellcolor{color_sym}\rotatebox{60}{\textit{OCR-CER}} & 
\cellcolor{color_sym}\rotatebox{60}{\textit{Attribute}} & 
\cellcolor{color_sym}\rotatebox{60}{\textit{Position}} & 
\cellcolor{color_sym}\rotatebox{60}{\textit{Cap/Other}} & 
\cellcolor{color_rel}\rotatebox{60}{\textit{Spatial}} & 
\cellcolor{color_rel}\rotatebox{60}{\textit{Occlusion}} & 
\cellcolor{color_rel}\rotatebox{60}{\textit{Social}} & 
\cellcolor{color_rel}\rotatebox{60}{\textit{Causal}} & 
\cellcolor{color_plan}\rotatebox{60}{\textit{w/o CoE}} & 
\cellcolor{color_plan}\rotatebox{60}{\textit{w/ CoE}} \\

& & & 
\multicolumn{4}{c|}{\cellcolor{color_obs}\textbf{Basic Perception}} & 
\multicolumn{5}{c|}{\cellcolor{color_sym}\textbf{Advanced Perception}} & 
\multicolumn{4}{c|}{\cellcolor{color_rel}\textbf{Relation Understanding}} & 
\multicolumn{2}{c}{\cellcolor{color_plan}\textbf{Event Reasoning}} \\
\midrule

\multicolumn{18}{l}{\textit{\textbf{Qwen3.5 Series}}} \\
\hline

\multirow{2}{*}{Qwen3.5-9B} & \multirow{2}{*}{59.0} & \multirow{2}{*}{Qwen3.5} & 
61.8 & 60.4 & 58.9 & 37.0 & 
87.4 & 14.8 & 58.0 & 41.0 & 79.0 & 
59.5 & 48.5 & 62.0 & 55.5 & 
56.7 & 58.8 \\
& & & \multicolumn{4}{c|}{54.5} & \multicolumn{5}{c|}{66.3} & \multicolumn{4}{c|}{56.4} & \multicolumn{2}{c}{58.8} \\
\hline

\multirow{2}{*}{Qwen3.5-9B-SFT} & \multirow{2}{*}{64.6} & \multirow{2}{*}{Qwen3.5} & 
65.5 & 63.2 & 74.8 & 56.5 & 
89.0 & 12.9 & 65.5 & 52.3 & 80.4 & 
63.0 & 52.3 & 67.0 & 60.5 & 
60.0 & 60.9 \\
& & & \multicolumn{4}{c|}{65.0} & \multicolumn{5}{c|}{71.8} & \multicolumn{4}{c|}{60.7} & \multicolumn{2}{c}{60.9} \\
\hline

\multirow{2}{*}{\textbf{EGVOR(Qwen3.5-9B)}} & \multirow{2}{*}{\textbf{69.4}} & \multirow{2}{*}{\textbf{Qwen3.5}} & 
\textbf{68.9} & \textbf{66.6} & \textbf{83.2} & \textbf{68.5} & 
\textbf{89.7} & \textbf{12.5} & \textbf{71.5} & \textbf{57.8} & \textbf{80.7} & 
\textbf{65.5} & \textbf{55.8} & \textbf{68.5} & \textbf{61.4} & 
\textbf{66.2} & \textbf{68.0} \\
& & & \multicolumn{4}{c|}{\textbf{71.8}} & \multicolumn{5}{c|}{\textbf{75.0}} & \multicolumn{4}{c|}{\textbf{62.8}} & \multicolumn{2}{c}{\textbf{68.0}} \\
\hline
\hline

\multirow{2}{*}{$\Delta$ \textit{v.s.} Qwen3.5-9B} & \multirow{2}{*}{\up{10.4}} & \multirow{2}{*}{Qwen3.5} & 
\up{7.1} & \up{6.2} & \up{24.3} & \up{31.5} & 
\up{2.3} & \down{2.3} & \up{13.5} & \up{16.8} & \up{1.7} & 
\up{6.0} & \up{7.3} & \up{6.5} & \up{5.9} & 
\up{9.5} & \up{9.2} \\
& & & \multicolumn{4}{c|}{\up{17.3}} & \multicolumn{5}{c|}{\up{8.7}} & \multicolumn{4}{c|}{\up{6.4}} & \multicolumn{2}{c}{\up{9.2}} \\

\bottomrule
\end{tabular}
}
\vspace{3pt}
\caption{\textcolor{black}{Generalization of EGVOR on the Qwen3.5-9B architecture (AD$^2$-Bench). $\Delta$ indicates the improvement of EGVOR(Qwen3.5-9B) over the original Qwen3.5-9B. For \textit{OCR-CER}, lower is better.}}
\label{tab:qwen35_series}
\vspace{-10pt}
\end{table*}

\begin{figure*}[t]
    \centering
    \includegraphics[width=1.\linewidth]{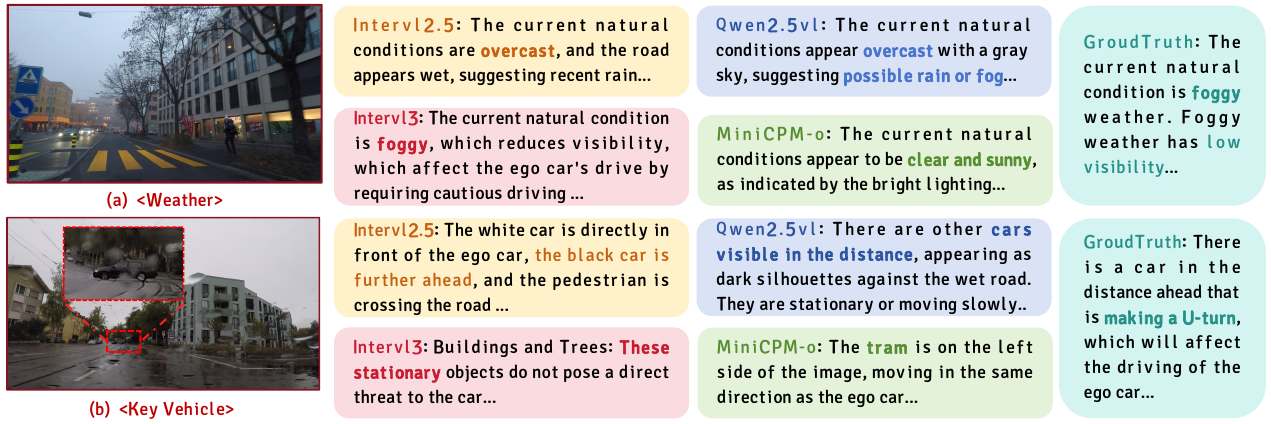}
    \vspace{-5pt}
    \caption{Bad case: (a) In light fog conditions, the ambiguity caused by the non-dense fog leads many MLLMs to make erroneous judgments.
(b) When key vehicles are distant and occluded by rain, most MLLMs fail to identify them accurately, sometimes even exhibiting object hallucinations.}
    \label{fig:bad_case}
    \vspace{-5pt}
\end{figure*}

\subsection{Bad Case Analysis and Findings}
 \vspace{-5pt}
\textbf{Qualitative Error Analysis.}
Qualitative analysis in Fig.~\ref{fig:bad_case} reveals significant MLLM failure modes. Models sometimes misinterpret subtle weather conditions (\eg, light fog, Fig.~\ref{fig:bad_case}(a)) easily perceived by humans. More critically, under adverse conditions like rain (Fig.~\ref{fig:bad_case}(b)), models frequently fail to detect crucial safety hazards, such as an unexpected U-turning vehicle directly ahead, while sometimes exhibiting dangerous object hallucinations (\eg, MiniCPM-o-2.6 perceiving a non-existent 'tram'). These errors highlight the urgent need for improved fine-grained perception and robust grounding in visual input, strictly avoiding hallucinations, for safe autonomous driving deployment.

\textbf{Instruction Following Ability.}
We observed limitations in models' adherence to specific formatting instructions. Several MLLMs generated extraneous analysis beyond the required single-letter answers for multiple-choice questions, produced excessively verbose outputs often exceeding token limits (particularly for reasoning VQA), and failed to consistently apply the required structural tags (\eg, $<startN>...<end>$) for CoE step evaluation. This indicates substantial room for improvement in the instruction following capabilities of current MLLMs, especially regarding structured output formats.
\subsection{Detailed Data Composition and Scene Partition} \label{data_coll}
While Section~\ref{subsec:data_collection} outlines the primary data sources, this section details the specific criteria used to partition the complex scenarios into intensity levels. This fine-grained categorization is crucial for analyzing model robustness under varying degrees of signal degradation.

\paragraph{Stratified Expert Annotation based on Intensity.}
To ensure consistent labeling across diverse data sources, we established quantitative visual criteria for the three intensity levels for different expert. Specifically, the \emph{Light} intensity is characterized by visual noise (e.g., drizzle or thin fog) that does not compromise the distinct boundaries of objects at distances greater than 50 meters. The \emph{Moderate} level involves scenarios where object boundaries become blurred at a medium range (20--50 meters), and textures of road surfaces, such as lane lines, are partially obscured. Finally, \emph{Severe} conditions represent a high signal-to-noise ratio environment where visibility is restricted to less than 20 meters (e.g., dense fog or heavy snowstorms). In these scenarios, semantic inference relies heavily on contextual reasoning rather than direct visual recognition.

\paragraph{Hybrid Filtering Protocol.}
For datasets lacking explicit weather labels, such as parts of ACDC, our semi-automated filtering pipeline employed a specific complexity threshold. We prompted Gemini 2.5-Pro to score ``Scene Complexity'' on a scale of 0 to 10 based on object density and visual clarity. Only samples with a complexity score exceeding 6.0 were forwarded to the manual verification stage, ensuring that the benchmark strictly targets \emph{complex} urban scenarios rather than empty or benign streets.

\subsection{Annotation Taxonomy and Prompt Implementation} \label{data_pre}
Complementing the \emph{Atomic Flow} strategy described in Section~\ref{subsec:atomic_annotation}, this section details the operational aspects of the annotation campaign and the technical implementation of vision prompts.

\paragraph{Taxonomy of 33 Sub-tasks.}
The cluster analysis of scene complexity yielded 33 distinct sub-tasks, hierarchically organized to support the cognitive pyramid. The Perception Tier encompasses 12 tasks, including fine-grained existence checks, dense and sparse counting, and the absolute localization of 8 distinct urban object classes. Moving up the hierarchy, the Relation Tier consists of 10 tasks specifically targeting spatial relations (e.g., relative positioning), social interactions (e.g., pedestrian-vehicle conflicts), and occlusion status reasoning. The highest level, the Reasoning \& Decision Tier, comprises 11 tasks focused on causal inference (e.g., explaining vehicle stops), future prediction, and the generation of actionable safety advice.

\paragraph{Vision Prompt Rendering Details.}
To ensure that vision prompts effectively guide attention without obscuring critical features, we applied strict rendering standards. Point Prompts are rendered as circular markers with a radius of 3 pixels (scaled to image resolution) using high-contrast colors (Red for target, Green for reference). Region Prompts utilize bounding boxes drawn with a 2-pixel width. For cases of ``Severe'' occlusion, the protocol prioritizes Point prompts to avoid the visual clutter typically associated with overlapping box borders.

\paragraph{Response Normalization and Instruction Adherence.}
We observed varying degrees of instruction-following capabilities across different model families, necessitating a robust output normalization protocol. Systematic formatting issues were particularly prevalent in object localization and Optical Character Recognition (OCR) tasks. In localization scenarios where models were explicitly prompted to output point coordinates formatted as $<x, y>$, certain architectures—most notably the MiniCPM series—exhibited a tendency to output bounding box coordinates (e.g., $<x_1, y_1, x_2, y_2>$) instead. To address this, we implemented an automatic geometric parsing layer that converts such bounding box outputs into their geometric center points for consistent metric calculation. Similarly, in OCR tasks involving traffic signs or license plates, models such as the LLaVA series frequently generated excessive hallucinations or verbose descriptions beyond the requested text content. To mitigate this, we enforced a strict length penalty; responses exceeding a threshold of 50 words were automatically classified as incorrect, with their accuracy scores set to zero. These normalization strategies ensure that the benchmark evaluates genuine perceptual capability rather than formatting compliance.

\subsection{Construction of SFT-Instruct.}
To facilitate the reasoning format establishment in Stage I, we constructed a bespoke dataset, SFT-Instruct, comprising 40K high-fidelity reasoning chains. While we utilize the VGR-158K~\cite{wang2025vgr} corpus as a foundational source due to its rich pseudo-chain-of-thought annotations, we introduce a rigorous \emph{Restructuring and Enrichment Pipeline} to align the data with the atomic requirements of EGVOR.

\textit{1) Coordinate System Adaptation for Spatial Fidelity.}
Standard datasets often employ normalized coordinates ($r \in [0, 1]$). However, to fully leverage the high-resolution grounding capabilities of our base model (Qwen2.5-VL series~\cite{bai2025qwen25vl}), we perform a precision-preserving conversion to \emph{absolute coordinates}. For every bounding box in the reasoning trace, we transform normalized vectors $[r_{x_1}, r_{y_1}, r_{x_2}, r_{y_2}]$ into absolute pixel values $[Wr_{x_1}, Hr_{y_1}, Wr_{x_2}, Hr_{y_2}]$, where $H \times W$ denotes the native image resolution. This adaptation ensures that the spatial anchors $b_t$ retain maximum fidelity, mitigating quantization errors during the ``Locate'' phase.

\textit{2) Semantic Densification and Filtering.}
To ensure the training signal supports complex cognitive paths, we impose a strict complexity filter, retaining only samples involving multi-hop spatial reasoning (i.e., trajectories with multiple distinct bounding boxes). This yields a core set of 40K samples. Crucially, to support the ``Describe'' phase of our evidence atom $a_t = \langle b_t, \mathbf{h}_t, c_t \rangle$, we perform \emph{Semantic Densification}. We rigorously verify that each spatial anchor $b_t$ is paired with a descriptive caption $c_t$ that explicitly details the visual attributes of the region. Where original descriptions were sparse, we augmented the annotations to ensure that $c_t$ provides sufficient semantic supervision for the latent alignment objective ($\Xi_{align}$).

\textit{3) Cognitive Augmentation via Synthetic Error Injection.}
To instantiate the \emph{Reflective Decoding} mechanism described in Section~\ref{stage1}, we constructed a ``Reflective Subset'' comprising 5K samples. We employ a controlled perturbation strategy: for a valid reasoning step, we inject a synthetic hallucination by inserting a random distractor box $b_{err}$ followed immediately by a corrective thought token sequence (e.g., ``\emph{Wait, this box seems to be wrong...}''). This design explicitly transforms the dataset from a static collection of correct answers into a dynamic curriculum that trains the model to detect and correct erroneous visual grounding—a critical capability for deployment in the complex environments of AD$^2$-Bench.

\subsection{Construction of RL-Instruct.}
To support the optimization of the composite objective functional $J_{total}$ defined in Stage II, we constructed RL-Instruct, a high-quality dataset comprising 40K samples. Unlike standard VQA datasets, RL-Instruct is meticulously curated to provide the necessary supervision signals for spatial uncertainty minimization ($\Psi_{spatial}$), latent alignment ($\Xi_{align}$), and cognitive path diversity ($\Lambda_{path}$). Our construction follows a \emph{Dual-Hybridization Strategy} to balance general reasoning capabilities with domain-specific robustness in complex urban environments.

\textit{1) Hard Sample Mining in General scenario.}
To ensure the policy learns from challenging scenarios rather than trivial patterns, we utilized the V* training dataset~\cite{wu2024vstar} as the foundational source. We implemented a \emph{Teacher-Guided Complex Scene Filtering} approach using Gemini-2.5-Pro. Specifically, we performed inference on the original 191K training set and retained only the ``Hard Samples'' where the teacher model identified high visual entropy or complex spatial dependencies. This resulted in a core set of 30K General Reasoning Samples. This selection ensures the model retains broad multimodal understanding capabilities, preventing catastrophic forgetting of general knowledge while learning domain-specific preferences.

\textit{2) Multi-Perspective Urban Augmentation.}
To strictly enforce the entropy reduction mechanism in high-noise environments, we augmented the dataset with 10K Complex Urban Samples. This subset is stratified into two distinct perspectives to maximize geometric invariance:
\begin{itemize}
    \item \textbf{UAV-View (5K):} Sampled from the VisDrone dataset~\cite{zhu2021visdrone}, focusing on high-density, small-object counting and tracking from a top-down perspective.
    \item \textbf{Ego-View (5K):} Sampled from the nuScenes dataset~\cite{caesar2020nuscenes}, focusing on occlusion reasoning and causal prediction from an ego perspective.
\end{itemize}
This hybrid composition prevents the policy from overfitting to a specific visual domain while forcing it to handle the extreme visual confusion characteristic of complex urban scenes.

\textit{3) Cognitive Annotation Enrichment.}
Standard annotations lack the granularity required for our reward functions. We employed a synthetic enrichment pipeline to generate hierarchical metadata. For the spatial objective $\Psi_{spatial}$, we implemented an \emph{Adaptive Weighting Protocol}. We first assigned raw importance scores to objects based on their causal relevance to the query (e.g., critical agents vs. background context). Crucially, to maintain a consistent optimization scale across diverse scenes, we enforced a Normalization Constraint per image, ensuring that the weights of all valid objects sum to unity ($\sum \omega_o = 1$). This effectively models the spatial reward as a weighted expectation over a fixed attention budget. Furthermore, to instantiate the path diversity objective $\Lambda_{path}$, we prompted Gemini-2.5-Pro to generate three distinct reasoning trajectories for each sample, formally defined as follows: (1) Perceptual-First (Bottom-Up): where reasoning strictly originates from objective, physical evidence in the image (e.g., spatial relations, attributes) before forming a conclusion; (2) Semantic-First (Top-Down): where reasoning is guided by high-level world knowledge and contextual understanding of the scene to infer object properties; and (3) Hypothetico-Deductive: where the process involves systematically formulating a hypothesis and seeking visual evidence to verify or falsify the question's central premise. 

By providing these distinct yet valid logical templates, we enable the GRPO algorithm to reward cognitive flexibility, preventing the model from collapsing into a single rigid reasoning pattern.




\subsection{Hybrid Annotation Protocols: Balancing Rigor and Depth} 
\label{ann_format}
To facilitate a comprehensive evaluation that encompasses both precise evidence verification and complex reasoning trajectories, AD$^2$-Bench employs a \textbf{Hybrid Annotation Strategy}. This dual-stream protocol assigns distinct formats to different cognitive tasks based on their output characteristics, ensuring that the evaluation metric aligns naturally with the cognitive complexity.

\paragraph{Discriminative Format (MC) for Evidence Verification.}
For atomic cognitive tasks such as \emph{Attribute Recognition} and \emph{Relation Understanding}, we adopt a rigorously designed Multiple-Choice (MC) format. This approach enables deterministic evaluation of the model's ability to distinguish fine-grained visual details against hard negatives. Crucially, to mitigate the ``blind guessing" and "Yes-bias" common in standard MCQA, we implement a robust Anti-Hallucination Mechanism directly within the option design. Unlike random sampling, distractors are explicitly drafted via an \emph{Adversarial Distractor Mining} strategy, introducing \emph{Visual Cousins} (visually similar objects, e.g., confusing a truck with a bus in fog) and \emph{Hallucination Traps} (statistically co-occurring but absent objects). Furthermore, the protocol enforces an \emph{Existence Verification} constraint by incorporating specific ``Abstain'' options (E: None of the above; F: Target absent). This design compels the model to validate the target's presence before classification, strictly penalizing unfounded confidence.

\paragraph{Generative Format (VQA) for Cognitive Trajectories.}
In contrast, for tasks requiring open-ended expressiveness—specifically \emph{Target Localization} and the construction of the Chain of Evidence (CoE)—we preserve the Hierarchical VQA format. Since \emph{Localization} requires precise coordinate regression and \emph{CoE} necessitates the articulation of a logical progression (from perception to decision), restricted options cannot adequately capture these high-dimensional outputs. The VQA format allows us to evaluate the \emph{completeness} and \emph{logic consistency} of the model's reasoning path, providing a granular view of its internal cognitive process beyond simple classification accuracy.

\begin{table*}[ht]
\centering
\begin{tabular}{p{\textwidth}} 
\toprule
\textbf{A. Bounding Box Identification (Perception Task)} \\
\texttt{[Image]} \texttt{[Question]} \ \ This task aims to identify the bounding box coordinates \texttt{[min\_x, min\_y, max\_x, max\_y]} of a specific object in an image, \\
where (\texttt{min\_x, min\_y}) is the top-left corner and (\texttt{max\_x, max\_y}) is the bottom-right \\ corner of the rectangle,
with the output format specified as \\ \texttt{<OUTPUT START>[min\_x, min\_y, max\_x, max\_y]<OUTPUT END>}. \ \ The answer is:\\
\midrule
\textbf{B. Multiple-Choice Reasoning (Cognitive Task)} \\
\texttt{[Image]} \texttt{[Question]} Here are the options below: \\
(A) \texttt{[Choice A]} \\
(B) \texttt{[Choice B]} \\
(C) \texttt{[Choice C]} \\
(D) \texttt{[Choice D]} \\
(E) \texttt{[Choice E]} \\
(F) \texttt{[Choice F]} \\
Select the best answer to the above multiple-choice question based on the image. \\
Please reply with the letter (A, B, C, D, E or F) corresponding to the correct option. \\ The best answer is: \\
\bottomrule
\end{tabular}
\caption{\textbf{Standardized Annotation Format for AD$^{2}$-Bench Evaluation}}\label{tab:prompt}
\end{table*}

\subsection{Hierarchical Atomic Decomposition}
\label{subsec:atomic_decomposition}

To simulate the cognitive process of an intelligent agent navigating complex visual environments, we enforce a strict \emph{Cognitive-Process-Oriented} annotation sequence. Experts are mandated to deconstruct the scene in a logical order, establishing a causal chain from global context to local details.

\textbf{1. Environmental Condition \& Visual Constraint Analysis.} The process begins with a global assessment, where experts analyze visual degradations (e.g., visibility range, glare intensity) and physical surface conditions. This step establishes the \emph{environmental priors} necessary for robust reasoning in complex scenes.

\textbf{2. Key Risk Entity Identification.} Instead of exhaustively listing all background objects, annotators prioritize \emph{salient entities} that influence the agent's decision-making. This mimics the selective attention mechanism of human perception, filtering out noise to focus on critical elements.

\textbf{3. Fine-grained State \& Attribute Inference.} Experts then perform deep visual decoding on the identified entities. Crucially, this involves inferring attributes obscured by adverse weather, such as determining an object's orientation or status through fog and low light.

\textbf{4. Predictive Interaction Modeling.} Finally, the process culminates in causal reasoning. Experts explicitly annotate potential interactions, predicting future trajectories and defining the causal relationship between the entity's behavior and the observer's required response.

\subsection{Trust-Aware Annotation Pipeline}
\label{subsec:annotation_pipeline}

To guarantee both \emph{internal coherence} within a reasoning chain and \emph{inter-annotator agreement (IAA)} across the dataset, we implemented a rigorous pipeline governed by a Single-Annotator Coherence Policy augmented with Quality Checks.

\textbf{Privacy-Utility Trade-off.} Acknowledging the tension between privacy protection and task utility (specifically OCR), we adopt a differentiated privacy protocol. \emph{Biometric Anonymization} is applied to all human faces using an automated detection-blurring pipeline followed by manual verification. However, for \emph{License Plates}, which are critical for the OCR sub-task, we retain the visual information while enforcing strict data usage protocols.

\paragraph{Quality Assurance via IAA.}
To mitigate subjective variance, we established a rigorous qualification standard using Fleiss' Kappa ($\kappa$). During the training phase, candidate experts were required to annotate a challenging set of 50 pre-validated samples (e.g., sandstorm scenarios). We quantified agreement across four key dimensions. As shown in Table~\ref{tab:iaa}, initial agreement on fine-grained objects was substantial ($\kappa \approx 0.75-0.83$). However, the ``Final Decision" initially showed divergence ($\kappa=0.86$). Through iterative discussion and the establishment of a ``Conservative Reasoning Protocol"—which mandates prioritizing safety when visual cues are ambiguous—we achieved an ``Almost Perfect" agreement ($\kappa=0.92$). This ensures that the reasoning logic in AD$^2$-Bench aligns with a unified, safety-critical expert consensus.

\begin{table*}[htbp]
\centering
\resizebox{1.0\linewidth}{!}{%
\begin{tabular}{lccccc}
\toprule
\textbf{Evaluation Dimension} & \textbf{\makecell{More Detailed\\than GT}} & \textbf{\makecell{Identical\\to GT}} & \textbf{\makecell{Weaker\\than GT}} & \textbf{\makecell{Lacking\\Critical Info}} & \textbf{$\kappa$ (Kappa)} \\ \midrule
Person Perception & 2 & 39 & 6 & 3 & 0.75 \\
Vehicle Perception & 5 & 36 & 7 & 2 & 0.78 \\
Traffic Sign Perception & 3 & 42 & 4 & 1 & 0.83 \\
Final Decision (Initial) & 4 & 41 & 4 & 1 & 0.86 \\ 
\rowcolor{gray!10} \textbf{Final Decision (Calibrated)} & \textbf{3} & \textbf{44} & \textbf{2} & \textbf{1} & \textbf{0.92} \\ \bottomrule
\end{tabular}}
\caption{\textbf{Inter-Annotator Agreement (IAA) Analysis.} The table displays the distribution of annotation consistency against Ground Truth (GT) and the resulting Fleiss' Kappa score. The significant improvement in "Final Decision" demonstrates the effectiveness of our calibration protocols.}
\label{tab:iaa}
\vspace{-10pt}
\end{table*}
\begin{table*}[t]
    \centering\small
    \setlength{\tabcolsep}{4.5pt}
    \begin{tabular}{l c cccc}
    \toprule
    \textbf{Model Under Test} & \textbf{Human (Ref)} & \textbf{Prompt 1} & \textbf{Prompt 2} & \textbf{Prompt 3} & \textbf{Prompt 4} \\
    & (Expert) & (Nominal) & (Detailed) & (Few-Shot) & (Many-Shot) \\
    \midrule
    InternVL3-8B & 53.0 & 61.1 \color{gray}\scriptsize(+8.1) & 56.2 \color{gray}\scriptsize(+3.2) & 53.5 \color{blue}\scriptsize(+0.5) & 52.8 \color{teal}\scriptsize(-0.2) \\
    Qwen2.5-VL-7B & 52.3 & 60.8 \color{gray}\scriptsize(+8.5) & 55.6 \color{gray}\scriptsize(+3.3) & 52.7 \color{blue}\scriptsize(+0.4) & 52.7 \color{teal}\scriptsize(+0.4) \\
    MiniCPM-o-2.6 & 51.2 & 58.5 \color{gray}\scriptsize(+7.3) & 54.4 \color{gray}\scriptsize(+3.2) & 51.8 \color{blue}\scriptsize(+0.6) & 50.7 \color{teal}\scriptsize(-0.5) \\
    \midrule
    \textit{Avg. Gap ($\Delta$)} & -- & +7.97 & +3.23 & \textbf{+0.50} & -0.1 \\
    \bottomrule
    \end{tabular}
    \vspace{5pt}
    \caption{
    \textbf{Calibration analysis of the GPT-5.2 evaluator.} 
    We report the scores and their deviation ($\Delta$) from the expert human reference. 
    }
    \label{tab:human_alignment}
    \vspace{-10pt}
\end{table*}
\subsection{Metric Validation and Evaluator Calibration}
\label{subsec:metric_validation}

To ensure the reliability of our automated evaluation framework (utilizing \textbf{GPT-5.2}), we conducted a rigorous calibration study against human judgment. We established a ``Reference Standard'' human baseline using a team of domain experts who underwent strict training to unify evaluation criteria, resulting in a conservative and precise scoring distribution.

\paragraph{Iterative Prompt Optimization.}
We systematically refined the evaluator's alignment with human experts through four progressive prompt levels. We detail the evolution of these prompts below to ensure reproducibility:

\begin{description}
    \item[\textbf{Level 1: Nominal Definition (P1).}] 
    This prompt provides only high-level, nominal definitions of the four trustworthiness metrics. It relies entirely on the model's inherent semantic understanding of terms like ``Factual Fidelity'' or ``Logical Soundness'' without defining specific penalty thresholds. This serves as a baseline to measure the model's unconstrained bias.
    
    \item[\textbf{Level 2: Expert-Derived Specification (P2).}] 
    This prompt incorporates a comprehensive grading rubric derived from our expert annotation manual. It explicitly details nuances often overlooked by standard models, such as: (1) distinguishing between ``plausible'' and ``visually grounded'' hallucinations; (2) penalizing ``correct answers derived from wrong reasoning'' (False Positive Logic); and (3) defining the boundaries of spatial ambiguity in adverse weather.
    
    \item[\textbf{Level 3: Few-Shot Anchor Calibration (P3).}] 
    To bridge the gap between abstract definitions and concrete judgment, we introduce specific ``Anchor Examples'' into the context. This prompt includes a curated set of difficult cases (e.g., tiny objects in fog) accompanied by expert scores and detailed justifications. These anchors serve as cognitive reference points, calibrating the model's output distribution to match the conservative nature of human experts.
    
    \item[\textbf{Level 4: Many-Shot Saturation (P4).}] 
    Built upon P3, this variant doubles the number of anchor examples to fully saturate the context window, testing the upper bound of inter-rater agreement.
\end{description}

\paragraph{Analysis and Selection.}
As summarized in Table~\ref{tab:human_alignment}, simple prompts (P1) lead to severe score inflation ($\Delta \approx +8.0$), reflecting the model's tendency towards leniency. The introduction of detailed rubrics (P2) reduces this gap, but substantial alignment is only achieved with the inclusion of anchor examples in Prompt 3, where the deviation narrows to $\le 0.6$. While P4 achieves near-perfect alignment ($\Delta \approx 0$), the marginal gain does not justify the doubled computational cost. Consequently, we adopt Prompt 3 as the standard protocol for AD$^2$-Bench, striking the optimal balance between evaluation fidelity and efficiency.

\subsection{Reasoning Format}\label{format}
Fig.~\ref{fig:case1}(left) illustrates our Chain-of-Evidence (CoE) process for decomposing complex autonomous driving decision-making problems. This methodology employs a hierarchical reasoning process, structured around key elements within the traffic scene, to address intricate decision-making tasks.

The reasoning process commences with an assessment of \textbf{environmental factors}, which typically constitute the most readily perceived and predominant visual information. This foundational environmental perception then grounds the identification of \textbf{key vehicles} pertinent to the ego-vehicle's driving behavior. As other vehicles often represent the most significant immediate threats, the CoE process, upon establishing environmental awareness, prioritizes reasoning about these potentially hazardous agents.

The reasoning hierarchy further extends to \textbf{Vulnerable Road Users (VRUs)}, such as pedestrians and cyclists, whose dynamic presence critically influences the ego-vehicle's maneuvers. Although potentially posing a less immediate physical threat than other vehicles, VRUs demand dedicated perceptual analysis and reasoning. Subsequently, the fourth reasoning layer, designated herein as \textbf{Step 4}, addresses \textbf{traffic rules}. The accurate extraction and interpretation of relevant traffic regulations from the scene are paramount for ensuring sound subsequent decision-making.

Finally, by consolidating insights derived from these hierarchically processed layers of perception and understanding, the Chain-of-Thought process culminates in a rational and justifiable driving decision.
\subsection{Case Study of Reasoning Task}\label{case}
\begin{figure*}[t]
    \centering
    \includegraphics[width=1.\linewidth]{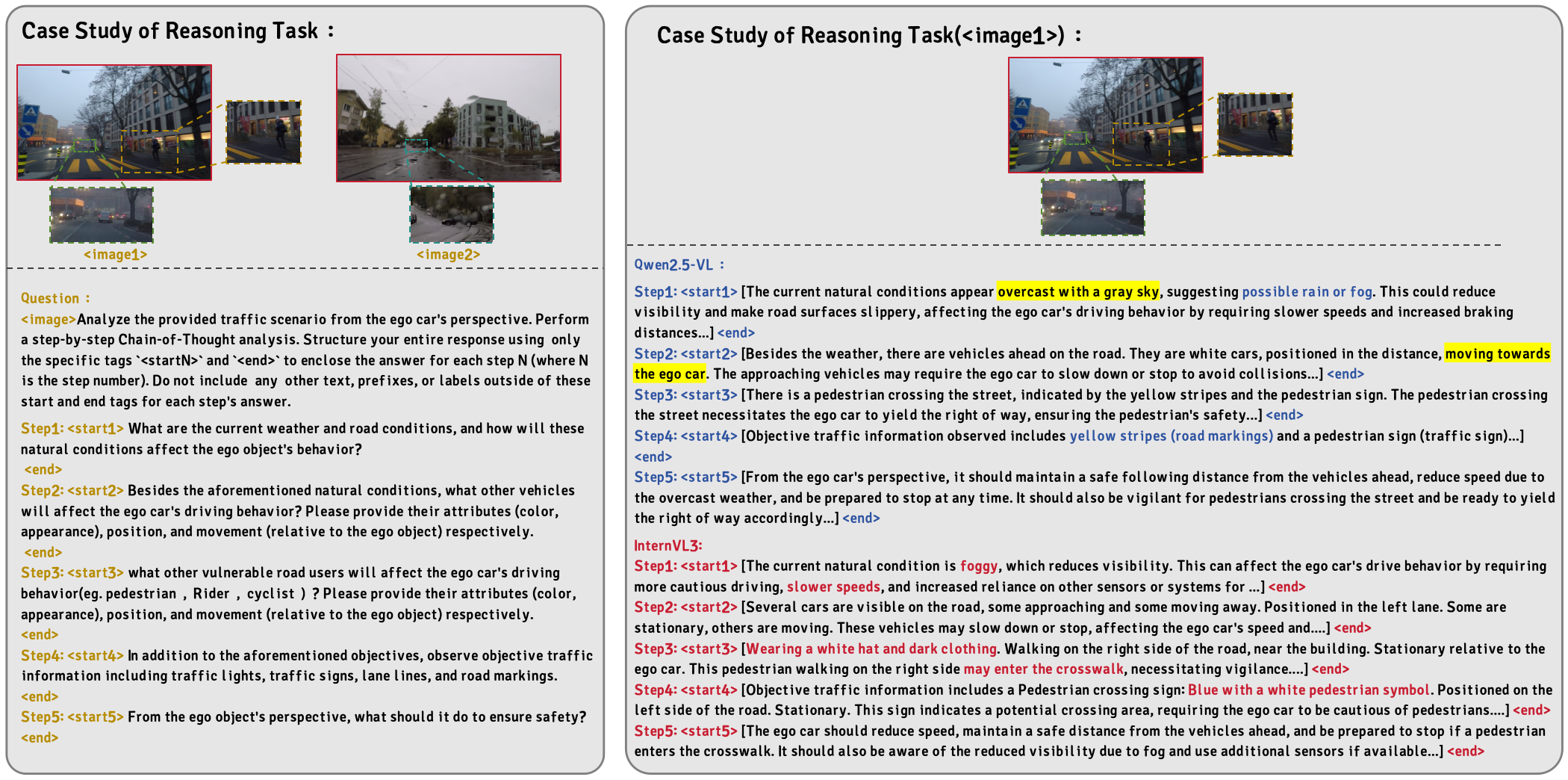}
    \caption{The visualization of Hierarchical Visual Diagnosis for MLLM.}
    \label{fig:case1}
\end{figure*}

\begin{figure*}[t]
    \centering
    \includegraphics[width=1.\linewidth]{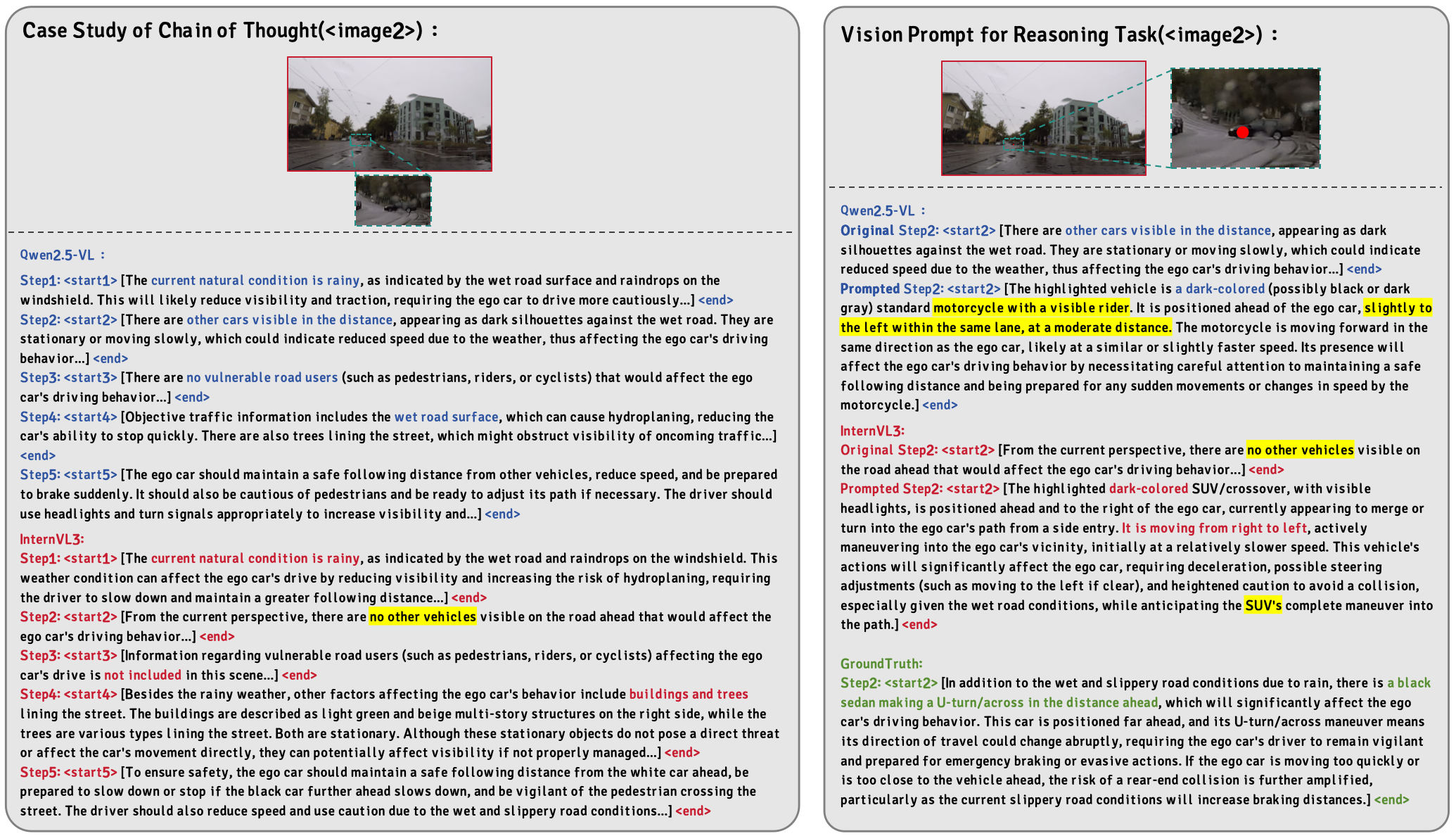}
    \caption{The visualization of Hierarchical Visual Diagnosis for MLLM. By breaking down unified scene problems into fine-grained perception, understanding, and reasoning tasks, CoE assists MLLM in achieving a hierarchical and comprehensive understanding of the current scene for making the most reasonable decisions.}
    \label{fig:case2}
\end{figure*}

\textbf{Weather Analysis.} Fig.~\ref{fig:case1}(right) and~\ref{fig:case2} present responses from two advanced models, InternVL3 and Qwen2.5-VL, to challenges posed by adverse weather conditions in complex scenarios. In these figures, areas shaded yellow indicate reasoning errors, while red or blue highlights denote correct portions of the answer. For the first image in Fig.~\ref{fig:case1}(left), we observe that Qwen2.5-VL, despite perceiving road fog, still concludes the weather is `overcast' and fails to correctly distinguish whether the current scene is rainy or foggy. Such light fog conditions, being similar in appearance to rainy scenes, can easily lead to model misjudgment. In contrast, InternVL3 accurately identifies the current weather and also provides a fine-grained description of pedestrians. However, both models exhibit distinct focuses when describing traffic signs and road markings: Qwen2.5-VL identifies a yellow pedestrian crossing, whereas InternVL3 points out a sign on the left. Our expectation is for models to comprehensively perceive and understand the entire scene; thus, their current capabilities in understanding this specific scenario still reveal certain deficiencies.

\textbf{Vision Prompt Analysis.} Regarding the second image in Fig.~\ref{fig:case2}(left), the most critical element is a black sedan ahead, obscured by rain, which is either making a U-turn or crossing the road. Qwen2.5-VL perceives several white cars parked on the roadside; however, these are irrelevant to the ego car's current driving status. The priority is key information that can impact the ego car's driving decisions. InternVL3 performs poorly in this instance. Although it segments the input into patches and resizes them to a fixed dimension for the vision encoder, this method appears ineffective here, as the model fails to detect any vehicles. This oversight is extremely detrimental to safe driving.

To further evaluate the reasoning capabilities of MLLMs and to mitigate the issue of insufficient perceptual acuity in vision encoders, we experimented with incorporating vision prompts in Fig.~\ref{fig:case2}(right). By overlaying annotated points onto the image and providing this composite input to the MLLM, we found that Qwen2.5-VL, while misidentifying the vehicle type, correctly recognized the vehicle's color and direction of movement (from right to left). This indicates that our vision prompts are largely effective, at least enabling the model to perceive the presence and movement of the vehicle ahead. Similarly, InternVL3 demonstrated even better performance: its vehicle type identification was more accurate than that of Qwen2.5-VL, and its overall responses were largely correct. This showcases the potential of vision prompts for conducting in-depth assessments of MLLMs' advanced reasoning capabilities.
\subsection{Case Study of Other Tasks}\label{case_per}
\begin{figure*}[t]
    \centering
    \includegraphics[width=1.\linewidth]{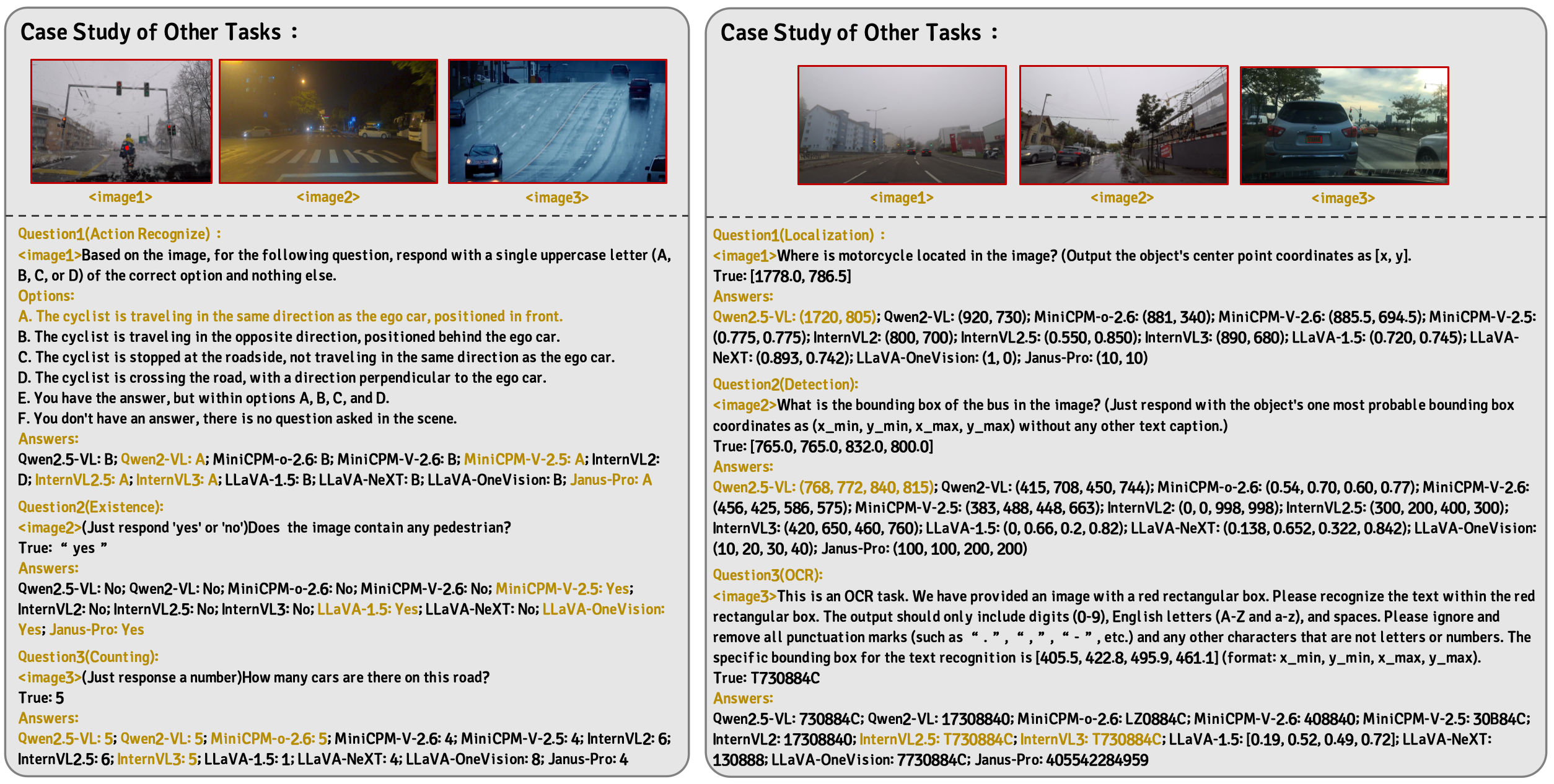}
    \caption{The visualization of Basic and Advanced Perception Tasks of Multimodal large language Models under Adverse Weather and Complex Scene Conditions, with yellow highlights representing correct answers.}
    \label{fig:case_other}
\end{figure*}

Fig.~\ref{fig:case_other} presents further case studies across various tasks. From Fig.~\ref{fig:case_other}(left), it is evident that adverse weather and complex scenarios significantly impair the perceptual capabilities of MLLMs. This is particularly apparent in a specific image from this figure (referred to as image2), concerning the detection of pedestrian presence. In this nighttime intersection scene, although several pedestrians are situated on the sidewalk across from the ego-vehicle's front-right, most models fail to detect them. This failure is attributed to occlusion by a white car turning from the front-right, compounded by the low-light conditions. Such perceptual misses are highly detrimental to subsequent understanding and reasoning processes.

Fig.~\ref{fig:case_other}(right) showcases MLLM performance on localization, detection, and Optical Character Recognition (OCR) tasks. For localization and detection, Qwen2.5-VL notably outperforms other models. Many models tend to output coordinates in a proportional (normalized) format; however, due to image patching and sampling operations during input processing, relying on these proportions often fails to accurately reconstruct true bounding boxes. In contrast, approaches similar to those used by Qwen2.5-VL and Qwen2-VL, which involves passing the entire image to subsequent modules via a dynamic resolution scheme—achieve higher scores on these tasks. This improved performance stems from the fact that the final bounding box and localization point coordinates are established within the entire image's coordinate system, rendering full-image input superior to patched or sampled inputs in terms of overall perceptual integrity and output format consistency.

Regarding OCR tasks, especially in low-light conditions, most models manage to output text in the expected format. LLaVA-1.5, however, only outputs coordinates, indicating poor instruction-following capabilities. Our findings show that InternVL series models achieve significantly better OCR recognition results compared to others. Conversely, MiniCPM series models, which also utilize patched image inputs, deliver underwhelming performance. This suggests that challenges in OCR perception are not solely attributable to the input format and may necessitate greater consideration of training data. We observe that MiniCPM models, designed for edge-device deployment, possess training datasets with substantially less OCR-specific content than more general-purpose MLLMs like the InternVL series, a factor that critically contributes to their diminished OCR efficacy.

\subsection{Per-condition Robustness Analysis}
\label{sec:per_condition_analysis}

\begin{figure*}[t]
    \centering
    \includegraphics[width=0.98\textwidth]{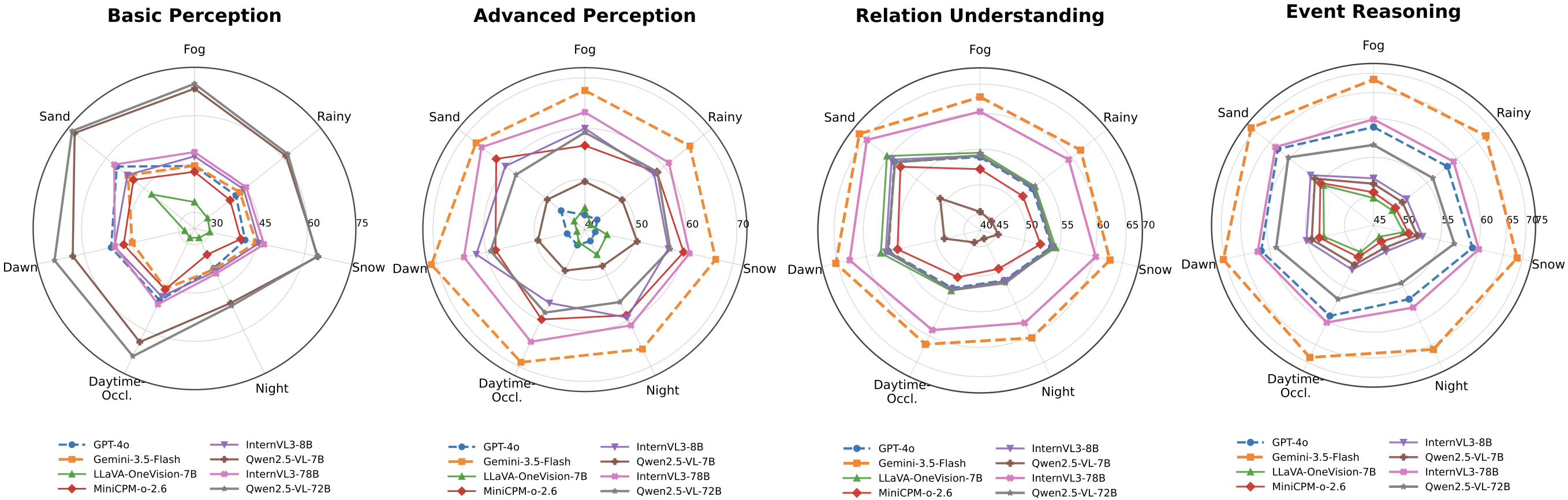}
    \caption{
    \textcolor{black}{Per-condition robustness analysis on AD$^2$-Bench. The four radar plots report representative MLLMs across seven adverse conditions from Basic Perception, Advanced Perception, Relation Understanding, and Event Reasoning. }}
    \label{fig:per_condition_radar}
    \vspace{-4mm}
\end{figure*}

\textcolor{black}{To further examine robustness under different adverse conditions, we provide a per-condition breakdown in Fig.~\ref{fig:per_condition_radar}. Instead of reporting only an overall average, we evaluate representative proprietary and open-source MLLMs across seven conditions and summarize their behavior from four perspectives: Basic Perception, Advanced Perception, Relation Understanding, and Event Reasoning. This visualization reveals that adverse conditions affect different stages of the reasoning pipeline in different ways. For instance, night and daytime-occlusion introduce more visible degradation in perception-related and relation-understanding scores, suggesting that low-level visual uncertainty can propagate to higher-level reasoning.}

\textcolor{black}{The results also show that stronger overall models are not uniformly robust across all adverse conditions. Some models maintain competitive relation or event reasoning performance, but still exhibit weakened visual grounding under challenging conditions, indicating that reasoning ability alone cannot fully compensate for unreliable perceptual evidence. Conversely, models with stronger visual grounding tend to be more stable in Basic Perception, but may not always dominate in higher-level reasoning categories. These findings support our motivation that robust adverse-condition understanding requires both reliable evidence extraction and effective reasoning over the constructed evidence.}




\end{appendices}

\end{document}